\documentclass[times,twocolumn,final,authoryear]{elsarticle}

\usepackage{ycviu}
\usepackage{framed,multirow}

\usepackage{amssymb}
\usepackage{latexsym}

\usepackage{url}
\usepackage{xcolor}
\definecolor{newcolor}{rgb}{.8,.349,.1}

\journal{Computer Vision and Image Understanding}

\begin{document}

\clearpage

\ifpreprint
  \setcounter{page}{1}
\else
  \setcounter{page}{1}
\fi

\begin{frontmatter}

\title{Context Understanding in Computer Vision: A Survey}

\author[1]{Xuan \snm{Wang}\corref{cor1}}
\ead{xwang4@gradcenter.cuny.edu}
\author[1,2]{Zhigang \snm{Zhu}} 
% \ead{zzhu@ccny.cuny.edu}
\cortext[cor1]{Corresponding author}

\address[1]{The Graduate Center, The City University of New York, 365 5th Avenue, New York, NY 10016, USA}
\address[2]{The City College, The City University of New York, 160 Convent Avenue, New York, NY 10031, USA}

\received{1 May 2013}
\finalform{10 May 2013}
\accepted{13 May 2013}
\availableonline{15 May 2013}
\communicated{S. Sarkar}

\begin{abstract}
Contextual information plays an important role in many computer vision tasks, such as object detection, video action detection, image classification, etc. Recognizing a single object or action out of context could be sometimes very challenging, and context information may help improve the understanding of a scene or an event greatly. Appearance context information, e.g., colors or shapes of the background of an object can improve the recognition accuracy of the object in the scene. Semantic context (e.g. a keyboard on an empty desk vs. a keyboard next to a desktop computer ) will improve accuracy and exclude unrelated events. Context information that are not in the image itself, such as the time or location of an images captured, can also help to decide whether certain event or action should occur. Other types of context (e.g. 3D structure of a building) will also provide additional information to improve the accuracy. In this survey, different context information  that has been used in computer vision tasks is reviewed. We categorize context into different types and different levels. We also review available machine learning models and image/video datasets that can employ context information. Furthermore, we compare context based integration and context-free integration in mainly two classes of tasks: image-based and video-based.  Finally, this survey is concluded by a set of promising future directions in context learning and utilization.
\end{abstract}

\begin{keyword}
\MSC 41A05\sep 41A10\sep 65D05\sep 65D17
\KWD Context\sep Computer Vision\sep Context Integration

\end{keyword}

\end{frontmatter}

%\linenumbers

\section{Introduction}

\begin{figure}[ht]
  \centering
  \includegraphics[width=1\linewidth]{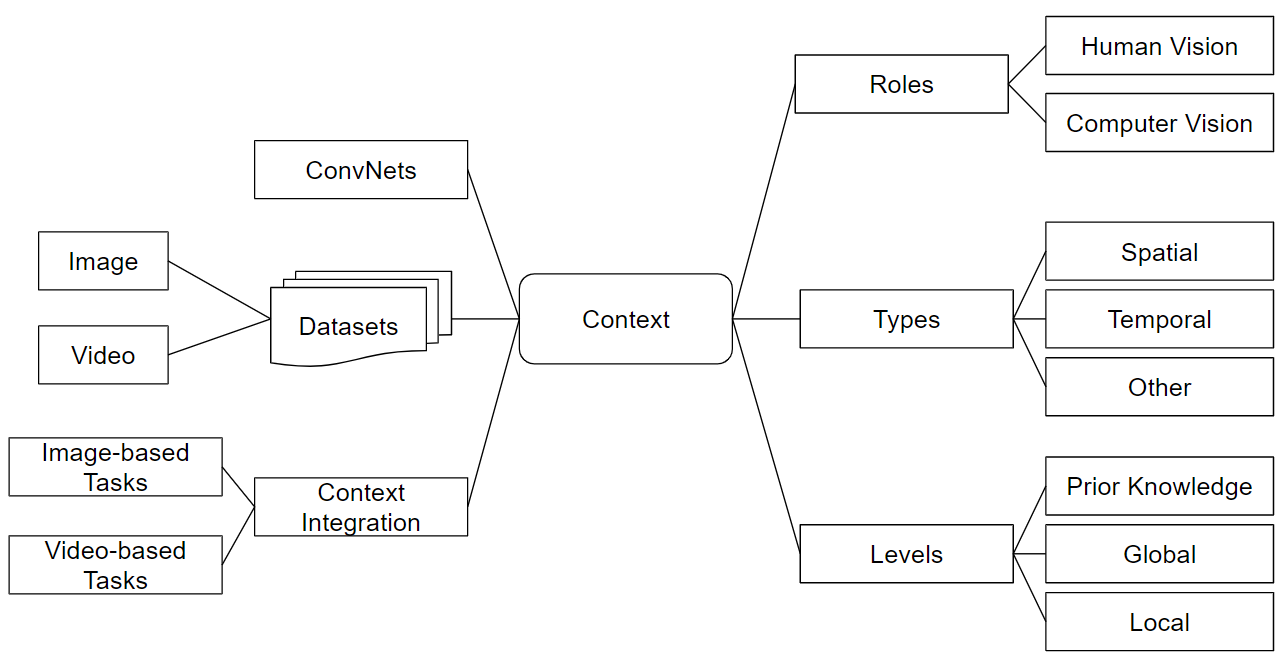}
  \caption{Overview of the survey.}
  \label{fig:introduction_diagram}
\end{figure}

Context refers to any information not only relating to the appearance of a target object or event itself, but also including other objects or events in the scene, visual or non-visual. Contextual information plays an important role in many computer vision tasks, such as object detection, video action detection, image classification, etc. In these tasks, context information may provide important clues for recognition and understanding. Recognizing a single object or action out of context may be challenging sometimes, but context information would be able to help improve the understanding of a scene or an event by providing additional information. Appearance context information, e.g., colors or shapes of the background of an object in the background can improve the recognition accuracy of the object. Semantic context (e.g. a keyboard on an empty desk vs. a keyboard next to a desktop computer ) will improve accuracy and exclude unrelated events.  Other contextual information, such as the time or location of an images captured, that are not in the image itself can also help to decide whether certain event or action should occur. Context information with a different sensory measurement, e.g. 3D structure of a building, will also provide additional information to improve the accuracy. This survey provides an overview on context understanding in various computer vision tasks and how context has been employed in different approaches. Figure \ref{fig:introduction_diagram} shows an overview of the survey. We first categorize context into several different types: spatial context, temporal context, and other context. Then we look into the use of context at different levels: prior knowledge level, global feature level, and local feature level. 

Humans and machines may treat context differently. Humans have the ability of detecting and recognizing objects, and performing other visual tasks in various environments at ease. These environments include occlusions, illumination changes, different viewpoints, which are still challenging for modelling, reasoning by computer vision solutions. Context is used by humans effectively and effortlessly to perceive the real world. Even the object is blurred, and can not be recognized in isolation, humans can recognize the target immediately with the help of the context (Fig.\ref{fig:hvexample}). Compare to humans, context modelling is still challenging for computer vision. We will discuss the role of context in human vision and computer vision in Section \ref{contextrole}. 

Context has been be used in various ways in computer vision tasks. Co-occurrence of the objects can influence the presence of a target object or event. Spatial relation between objects (e.g., painting is on the wall) provides cues to the location of the target. Temporal information such as nearby frames, previous clips can help to predict what will happen in the future. Semantic context from the scene potentially indicates how likely of an object or event to be found in certain scenes but not others, and it can be spatial semantic or temporal semantic. Other context like non-visual information in the meta-data of image collection (such as dates, environments, locations), can also be used as context information. In Section \ref{contexttypes}, we review context based approaches and discuss context in three major types: spatial context, temporal context and \textcolor{black}{other} context.

Intrinsic information like camera parameters, non-visual information like description of a target or an event, and metadata from the image or video, can all serve as prior knowledge context. These prior knowledge can indicate a target (e.g., TV in the living room) or an event (e.g. parking a car in a parking lot) we should expect in the scene and, what should not appear in the scene (e.g., a watermelon in a football court). Global context from the whole scene can be used for image recognition, and serves as the prior of object recognition as mention before. Local context from a target or an event itself has obvious features and can be used for detecting and recognizing the object or the event. It can also be used with global context and prior knowledge for small targets, because of lack of the local context representation or inconspicuous features. We further discuss context in different levels and review how these context have been employed in context based approaches in Section \ref{contextlevel}.

Many context based approaches use deep learning methods. Different kinds of network architectures have been used as backbones in context based integration. Various convolutional network architectures have been proposed to train the deep convolutional neural networks (ConvNets). Among reviewed literatures, ResNet and VGGNet are mostly used architectures in context based approaches. Many researches use either the existing ResNet and VGGNet in employing context information, or a modified version to better incorporate with context related to the tasks. Other architecture like Graph Convolutional Network (GCN), is used for modelling the spatial relation between target and others, and semantic relation between different object categories because of its unique graph structure. In Section \ref{secconvnets}, we review deep convolution network architectures that have been used in context based approaches.

Many datasets have been proposed and widely used in computer vision tasks, such as object detection, image classification and video event recognition etc. Although many state-of-art methods have good performance on large-scale datasets, these methods are lack of using rich context information provided by these datasets. In section \ref{datasets}, we separate these datasets into image datasets and video datasets, and review the datasets by further providing details on what context information is included.

Furthermore, we review various context based integration in two categories: image-based context integration and video-based context integration. Spatial context and semantic context are mostly used in image-based context integration, which can provide information like locations, environments, weathers, etc., and lead to potential performance improvement. Video-based context integration incorporates spatial context and semantic context through temporal context. Context along the temporal dimension serves as prior knowledge of target object or event, which could improve the performance over context-free approaches. Global and local context are used for extracting features of the scene and the target object or event. In Section \ref{contextintegration}, we provide details of some represented context based integration in different computer vision tasks, and we compare the performance of image-based context integration (Section \ref{secimagebasedperformance}) and video-based context integration (Section \ref{secvideobasedperformance}). \textcolor{black}{The merits of the reviewed works are also summarized in terms of (1) human likeness, (2) accuracy, and (3) efficiency for data and time (Section \ref{secmeritsofcontextintegration}). }

In summary, this survey paper is organized as the follows. Section \ref{contextrole} provides an overview of how important of context information in human vision (\ref{subseccontextinhv}) and how context has been used in computer vision (\ref{subseccontextincv}). Section \ref{contexttypes} discusses context in three major types: spatial context (\ref{subsecspatial}), temporal context (\ref{subsectemporal}) and \textcolor{black}{other} context (\ref{subsecother}). Section \ref{contextlevel} presents context in three levels: prior knowledge level (\ref{priorlevel}),  global context level (\ref{globalcontextlevel}) and local context level (\ref{localcontextlevel}). We review some popular deep convolutional network architectures which have been used in context based approaches in Section \ref{secconvnets}.  Frequently used datasets that can employ context information are discussed in Section \ref{datasets}. Section \ref{contextintegration} presents how context has been integrated in image-based tasks and video-based tasks. Performance comparison of context based approaches is discussed in sub-section \ref{secimagebasedperformance} for image-based context approaches and sub-section \ref{secvideobasedperformance} for video-based context approaches. The merits of the reviewed works are also summarized in Section \ref{secmeritsofcontextintegration}. Finally, we conclude our survey and provide a set of promising future directions in Section \ref{secfuturedirection}.

\section{Roles of Context} \label{contextrole}

Human and machine treat context differently. Our brain not only processes the signal that our eyes send, but is also influenced by the rich context from the seeing. Human can localize and recognize the objects or events even in considerable amount of occlusions, illumination changes and various viewpoints, etc., which are still big challenges for computer vision. This gap can be caused by the differences of training and testing data. Machine are trained on images or video of certain objects or events, with certain context, but the models might be used for images or videos in a totally different context. Whereas human vision systems are very experienced with large variances of scenes (with or without objects or events, environment changes, appearance changes, etc.). Nevertheless, machine vision models and algorithms have been explored in decades in understanding context in a systematic way hopefully like humans, in various forms in computer vision tasks.

In this section, we review what is the role of context in human vision and in computer vision. We discuss the differences of context understanding between human vision and computer vision, and explain why context reasoning is still challenging but critical for computer vision. Examples from previous works are reviewed, to explain why context is important for both human and machine.

\subsection{Context in Human Vision} \label{subseccontextinhv}
\begin{figure}[ht]
  \centering
  \includegraphics[width=1\linewidth]{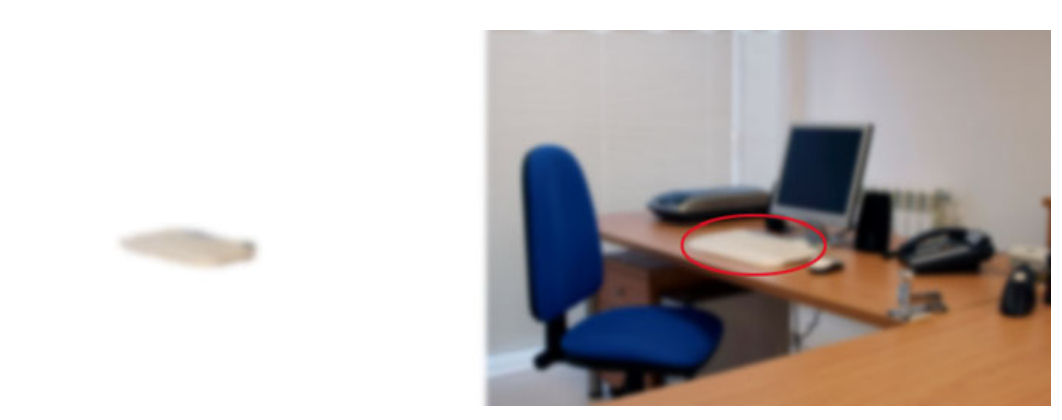}
  \caption{An example from \citep{marques2011context}. Left: An isolated object. Right: The object in its relevant scene.}
  \label{fig:hvexample}
\end{figure}

Humans use visual context effortlessly to perceive the real world. What we see is not only based on the signals that our eyes send to our brain, but is influenced strongly by the context. The visual stimulus is presented in, on our previous knowledge, and expectations. Intrinsic features (shapes, colors, texture, etc.) of an object against a background of the scene in the retinal images of our eyes provides enough information to determine what the object is. Human can also easily recognize the object under normal conditions. However, when an object appears in isolation from its surrounding scene, recognizing the object becomes unreliable. \textcolor{black}{A recent review \citep{vo2021meaning} discussed how human memorized objects in scenes. Our past experiences become  the key when interacting with real-world environments. Our mixed memories provide a scene grammar, which is the generic knowledge about what objects tend to be where, as well as more specific memories about that particular scene stored in episodic memory. Knowledge such as probable scene regions where objects tend to be found provides strong contextual cues during real-world search.} Fig \ref{fig:hvexample} shows an example of an object in isolation and the same object in context. When the keyboard is taking out from the office environment, human can barely recognize it. But within the office scene, we can identify the object in front of the monitor is a keyboard, even the surrounding area is blurry. Context provides critical information to help us visually find and recognize objects faster and more accurately.

Context encapsulates rich information not only on how natural scenes and objects are related to each other, but also the relative positions of objects with respect to a scene or co-occurrence of objects within a scene. Besides the visual form of context, non-visual information can also provide important cues. For example, without looking at an image, and if we know that there is a ship in the image, we can easily guess there is a river or sea in the image. Human can even draw a picture with only description of an object or an event. Human can benefit from either the visual information between objects and scenes, or the relations between semantically related objects.
 
\begin{figure}[ht]
  \centering
  \includegraphics[width=0.85\linewidth]{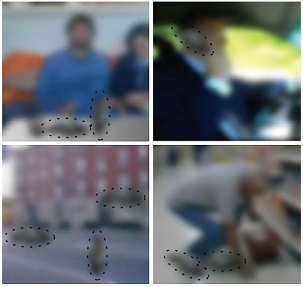}
  \caption{"Multiple personalities of a blob" from \citep{torralba2010using}. Object category is strongly influenced by its context.}
  \label{fig:hvblobexample}
\end{figure}

 What if an object present in irrelevant scenes? Behavior studies \citep{bar2003cortical, goh2004cortical} show that objects appearing in a familiar background can be detected more accurately and faster than objects in an unusual scene, where the objects are clearly recognizable to human observers in isolation. Context under such conditions could be misleading to the recognition, thus the context cannot provide useful information. %since it is indirect. 
 Antonio Torralba \citep{torralba2010using} presents "multiple personalities of a blob", which demonstrates how the same blob can be interpreted as different objects in different scenes. Fig. \ref{fig:hvblobexample} shows that the same blob can be interpreted as a car, a pedestrian, a phone, a bottle and even shoes under different context.

Nevertheless, contextual information in natural scenes provides critical information to help us visually find and recognize objects faster and more accurately. An object that cannot be recognized in isolation can be identified when it appears in relevant context scene. Context besides an object itself can serve as a supplement available for the object. Human can also infer information about the scene that will be useful for interpreting other parts of the scene. We can easily build a hierarchical relations between objects using not only visual context but also non-visual context.

So far it seems that humans can always perform better than machines. One potential reason for this performance gap is that humans and machines have qualitatively different learning mechanism. Machines are typically learned on images containing objects in a certain context with limited amount of data, whereas humans view objects in different context in real world daily. Another reason is that computer vision is trying to mimic human vision, but with the help of our brain, human vision is more advanced. We not only can learn context and object separately, but also easily build up connections and relations between objects and their context. In contrast, computer vision is still challenging to model the relation between objects and context. When the object has a weak correlation with its surrounding context, the context can be difficult to learn in the presence of more informative object features. On the other hand, if the object has a strong correlation with its context (e.g., living room usually contains a TV), the object can be learned effectively along with the context. These variations make machines hard to learn context systematically and independently of the object. Context also has different presentations between human vision and computer vision. 

\subsection{Context in Computer Vision} \label{subseccontextincv}
\begin{figure}[ht]
  \centering
  \includegraphics[width=1\linewidth]{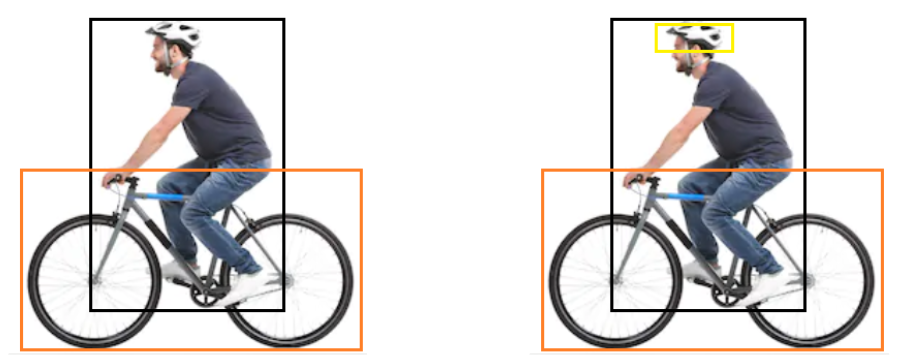}
  \caption{Example of information captured by machine. Machine can easily capture single object, but lack of relations between them. Images are from internet source.}
  \label{fig:cvexample}
\end{figure}

Although we can use context effortlessly with our human vision system, context reasoning about objects and relations is still challenging and critical to computer vision.  Fig. \ref{fig:cvexample} shows that a machine algorithm can recognize the object clearly: a rider (Black box), a bike (Orange box) and a helmet (Yellow box). With only capturing these kind information, parallel relations (aA man rides a bike) and hierarchical relations (helmet affiliated to head) are missing. 

Given an image or a video of the real world, the final goal of a computer vision system is to determine what visual elements and structures are presented, how these elements are related to each other, and to have a complete understanding of what is happening in the visual input. Visual understanding is difficult to define and evaluate, therefore researchers have concentrated on solving more focused, specialized, low-level problems like object detection or image classification. Object recognition does not occur as an isolated process since, it can be influenced by the presence of other objects as well as by the overall context of the scene. Global context provides a rich source of information that can help to improve the performance of the recognition task.

In order to build up the relation and understand the scene in the image, different context information has been employed in computer vision tasks. Many tasks, such as object detection \citep{du2012context, fang2017object, sun2017seeing, zhu2016could, zhu2021semantic}, video event recognition \citep{wang2015video, wang2016hierarchical}, video action detection \citep{yang2019step, zhu2013context}, scene graph generation \citep{xu2017scene, zellers2018neural}, data augmentation \citep{dvornik2018modeling}, image classification \citep{mac2019presence}, and image inpainting \citep{pathak2016context}. \citep{xu2017scene, zellers2018neural} use global context and semantic relations to generate scene graphs. Context from an object itself and neighborhood has been employed in object detection \citep{du2012context}, data augmentation \citep{dvornik2018modeling} and image inpainting \citep{pathak2016context}. Wang et al. \citep{wang2015video, wang2016hierarchical} introduce a hierarchical context model to recognize events in videos. Although contextual information has been used in different ways and gains more successes over context-free approaches, context could be misleading if an object present in irrelevant scenes (Fig.\ref{fig:hvblobexample}). Sun et al. \citep{sun2017seeing} use co-occurrence of curb ramps at street crossing to detect missing curb ramps at city street regions. Different context has been used widely in various computer vision tasks.

Context reasoning about objects and relations is critical to computer vision. Context reasoning consists of integrating object appearances and the spatial relations, semantic relations or prior knowledge to solve visual tasks. Several neural network architectures incorporating contextual information have been successfully applied in aforementioned computer vision tasks, demonstrating improved performance over context-free approaches. Some of the approaches have combined graphical models with neural network for structural inference, by leveraging semantic relations and spatial relations among scenes \citep{xu2017scene, zellers2018neural}, objects \citep{chen2019multi, zhu2021semantic} and their attributes \citep{yang2015facial, liu2015faceattributes, li2016human}. Although machine vision cannot understand context in a highly systematic way like humans, context has been effectively used in various forms and with different integrations in computer vision tasks.

\section{Major Types of Context} \label{contexttypes}

\begin{figure}[ht]
  \centering
  \includegraphics[width=1\linewidth]{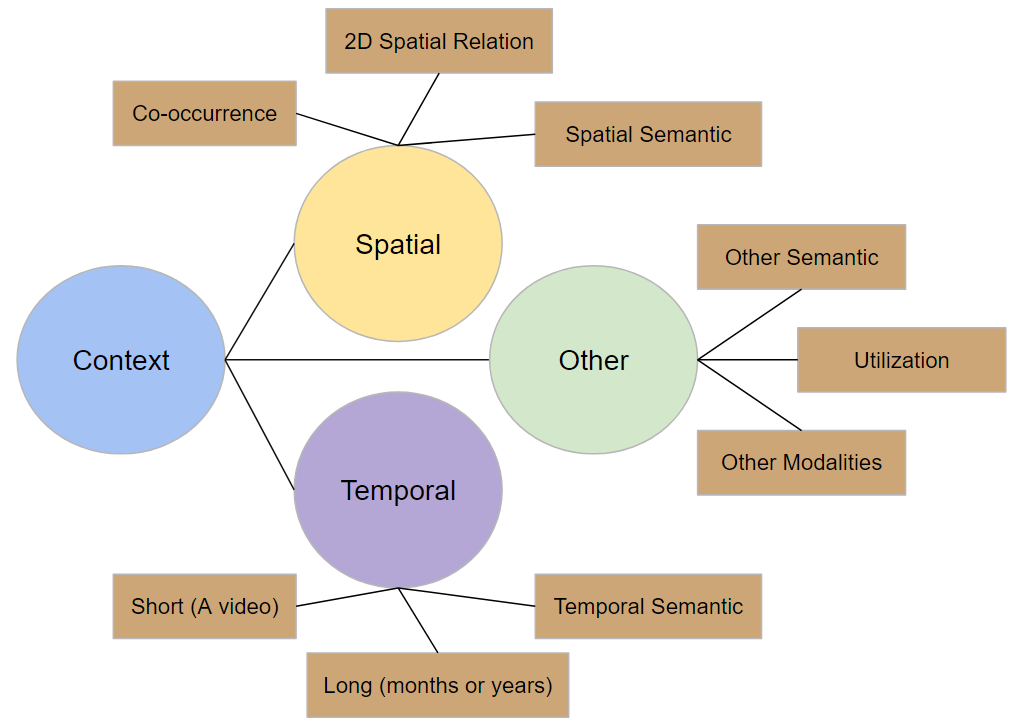}
  \caption{Three major context types, their relations and their sub-types.}
  \label{fig:majortyperepresentation}
\end{figure}

Context information can be from the appearance of the object or the event in consideration, such as shape, color, texture, etc., and can be also from any other information or data not directly related to the appearance of the object or the event, such as environment (inside or outside), location (classroom, restaurant, gym, etc.) and description (drinking coffee, riding bike, etc.), etc. We separate context into three major types: spatial context, temporal context and \textcolor{black}{other} context, as shown in Fig. \ref{fig:majortyperepresentation}. Spatial context represents the spatial relation between objects and events, such as \textcolor{black}{co-occurrence, 2D spatial relations and spatial semantic constraints}. Temporal context refers to temporally proximal information, either from nearby frames of a video in a short period, or similar scenes captured in months or years\textcolor{black}{, or temporal semantic constraints}. Semantic context can indicates an object or an event should be found in some scenes but not others. Semantic context can easily used to describe spatial context or temporal context in a language model \textcolor{black}{and we categorize them under spatial and temporal context types.} \textcolor{black}{Other context includes other semantic context that are neither spatial or temporal, and context clues from other modalities such as audio, thermal and weather, etc, and context information stemmed from utilization and purpose.} Three types of context are sometimes used in combination in various computer vision tasks.

In this section, we present a summary of three major context types. We describe each context type in detail and how they represent in different computer vision tasks.

\subsection{Spatial Context} \label{subsecspatial}
Spatial context can be defined as the likelihood of finding an object in some positions and not others with respect to other objects in the scene. A car is on the road, not in the sea. If a piece of glass is not on the wall, then it is not a window. An object in an image is supposed to fit into reasonable relationships with other objects in the image. The spatial context can provide information about these spatial knowledge. One of the simplest way to introduce relationships between objects in a scene is co-occurrence. Spatial knowledge such as "A bird is flying in the sky", can be translated directly into spatial relations between objects in the scene. As common sense, certain objects (e.g. Chopping boards, TVs) should occur more frequently in certain places (e.g., kitchens and living rooms, respectively). Spatial context usually refers to:
\begin {enumerate}
\item Environment of the object at present time and location.
\item  Related context around target object.
\item  Path/direction to destination.
\item  Events happen around the object. 
\end {enumerate}
How can we arrange the relationship between these context and the target objects? In the following we will discuss two major spatial context representations effectively used in context modeling and contextual reasoning: co-occurrence and 2-D spatial representation. \textcolor{black}{In addition to these two representations, we will also discuss spatial semantic context where semantic constraints can restrict spatial relations between objects.}  %The relations between spatial locations of objects or events will need extra information rather than the co-occurrence relations.

\subsubsection{Co-occurrence Representations}

\begin{figure}[ht]
  \centering
  \includegraphics[width=1\linewidth]{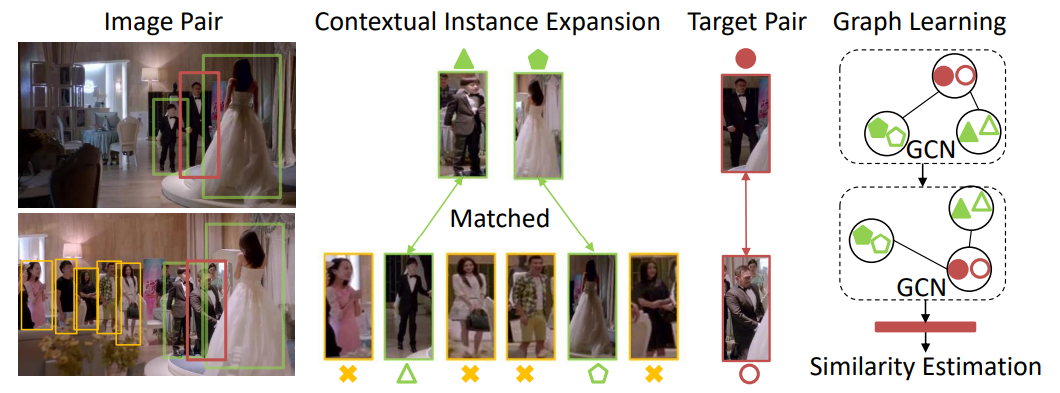}
  \caption{ An example of using co-occurrence of persons around the target. The figures was presented in \citep{yan2019learning}.}
  \label{fig:spatialexample_3}
\end{figure}

Co-occurrence is one of the simplest way to introduce relationships between objects in a visual scene. Contextual interactions such as “cars appear on roads” can be translated directly in contextual relations between object labels. In this case, the presence of a certain object class in an image (e.g., a road) statistically influences the presence of a target object (e.g., cars). It is straightforward to build context matrices to count co-occurrence of labels given a dataset where many objects are labeled. Such co-occurrence matrices can easily be translated. It is also well known that certain objects (e.g., computer monitors, beds) occur more frequently in some places (e.g., offices and bedrooms, respectively). Starting from these learned co-occurrence statistics, Rabinovich et al. \citep{rabinovich2007objects} devised interaction potentials for Condition Random Field (CRF) in order to measure contextual agreement between detected objects. It is interesting to notice that the terms “semantic context” and “co-occurence” are sometimes used interchangeably. The statistical model proposed by Carbonetto, Freitas and Barnard \citep{carbonetto2004statistical} also learns co-occurrence between concepts (e.g., image caption words). However in their model, Markov Random Field (MRF) interaction potentials are estimated only between neighboring image segments (e.g., object blobs). 

More than co-occurrence, such potentials also describe the ‘next to’ relationship between object labels. Wang et al. \citep{wang2007shape} give a formal definition of co-occurence and occurrence functions. These functions provide a probability distribution of labels over different regions centered around a given labeled pixel. Perhaps the most interesting idea in this scheme is to relate two independent sets of labels by using occurrence (and not only co-occurence). This integrates both shape and appearance labels into a shape and appearance context descriptor. An example is illustrated in Figure \ref{fig:spatialexample_3}. The objective is to identify whether the men in red bounding boxes belong to a same identity. However, the results are usually not confident as the appearance of the person suffer from great variation across different scenes. In this case, the model observes that the same persons in green bounding boxes appear in both scenes, thus a more confident judgment can be made that the men in red bounding boxes do belong to the same identity. Therefore, the persons in green bounding boxes play a positive role, while other persons in the scene are noise contexts. 

\begin{figure}[ht]
  \centering
  \includegraphics[width=1\linewidth]{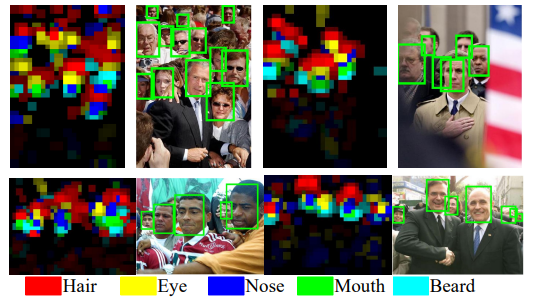}
  \caption{ The part-level response maps (or ‘partness’ maps) generated by a deep network as presented in \citep{yang2015facial}, given a full image without prior face detection by using spatial relation of different facial parts.}
  \label{fig:spatialexample_2}
\end{figure}

The occurrence analysis can be also made between object parts (e.g., detecting the nose and the mouth as part of a face). In this case again, the relative position between the parts is crucial. Fink and Perona \citep{fink2003mutual} detect faces by using both the individual detections of facial parts (left eye, right eye, mouth, nose, entire face) and their spatial arrangements. Hence, they treat M entities at each boosting iteration and compute M maps that give the likelihood of each entity appearing in different positions in the image. Combining face parts is made possible by using all the likelihood maps as additional input channels for subsequent boosting iterations. Consequently, the likelihood map for eye detection can be used to further detect mouth, and likelihood maps for face detection can be helpful to detect multiple faces, since faces tend to be horizontally aligned in the considered dataset. A large contextual window is used to analyze such co-occurrence and spatial relations. 

Another similar approach is brought by Perko and Leonardis \citep{perko2010framework}, in which the horizontal alignments between pedestrians in street scenes are learned by estimating a 2D probability distribution of other pedestrian locations given a pedestrian in the center of the image. A recent work by Yang et al. \citep{yang2015facial}, proposed Faceness-Net for face detection. Instead of using the whole face, the Faceness-Net considers using spatial structure and co-occurrence of face parts as a context cue to detect faces. As shown in Fig. \ref{fig:spatialexample_2}, a face considers co-occurrence of eyes, nose, mouth and hair. Each facial parts was generated by using the spatial relation of different facial parts. For instance, the hair should appear above the eyes, and the mouth should only appear below the nose, etc.

\subsubsection{2D Spatial Representations} \label{subsubsecspatial2d}

Marques et al. \citep{marques2011context} distinguish three classes to describe 2D relations: 
(1) Direction relations.
(2) Distance relations.
(3) Topological relations.
For better understanding how spatial context has been used in computer vision tasks, we focus on (1) Direction relations and (3) Topological relations in this survey. Direction relations express the direction of one object (the primary object) relative to another (the reference object). Such relations can be defined if a frame of reference is known. Usually the cardinal directions (E,N,S,W) and their refinements (NE, NW,SW, SE) can be used (tacitly) as an extrinsic frame of reference. One can also assume an intrinsic orientation of reference in the image space (so that a person can talk, for example, of an object being to the “right” of a building). The relative vertical positions (“above”, “below”) are frequently used and judged discriminative enough to detect object in conventional dataset like PASCAL \citep{Everingham10}, where as horizontal positions do not necessary carry much discriminative information.

\textcolor{black} {Distance relations provide the measurements on absolute pixels (e.g. the bird is 300 pixel away from the ground) or relative distance (e.g. the bird is about 1500m away from the ground). In normalized image spaces some distances can be expressed in terms of (absolute) pixels (e.g., “object A is about 200 pixels away from object B”). Heitz and Koller \citep{heitz2008learning} also cite some human knowledge e.g., “cars park 20 feet away from buildings” that highlight the limitation of 2D spatial reasoning with a single image since a 3D geometric context would be required to capture that relation. More frequently, relative measurements lead to 2D qualitative relations such as “close”, “far”, or “equidistant”. Determining the correct scale and associated thresholds for such relations is a difficult task in the general case, yet tractable in domain-specific applications.}

Topological relations describe the relationship between an object and its neighbors. Formally defined by considering interior, boundary and exterior of an object, intersection relations such as “touches”, “overlaps” , “contains” (in, inside), and “crosses” are often used in practice. Some authors also propose a slightly nuanced version (“encloses”) of the simpler relation “contains”. One should also notice that the simplest topological relation is when two regions/objects are “disjoint”. Of course, all these spatial arrangements can be further combined. For instance Singhal et al. \citep{singhal2003probabilistic} use “far above” and “far below” to mix direction and distance relations. They also mingle “left” and “right” relations and introduce a weaker “beside” relation. Heitz and Koller \citep{heitz2008learning} also combine all types of relation (eight directional relations, two different distances, and a topological “in” relation) to generate 25 candidate relationships from which they extract the most useful ones.  All these spatial relations can easily be described in language form. Many works \citep{sun2017seeing, xu2017scene, zellers2018neural, yang2019step} use semantic context to describe the spatial relation between objects and events.

\subsubsection{Spatial Semantic Context} 
\label{subsecspatialsemantic}

\textcolor{black}{In the above discussions, spatial context is encoded as the co-occurrence of or the 2D spatial relations with other objects.
%, or provides more specific information (distance, geographic information, intrinsic relation between parts of object, etc.) about the scene in which objects are usually found. 
Semantic context, which will be detailed below, can describe these spatial relations in a more general and effective way. In fact, most of the approaches with spatial context also employ semantic context in some degrees.} 
 %This will be detailed in Section \ref{subsecspatialsemantic}.

\begin{figure}[ht]
  \centering
  \includegraphics[width=0.9\linewidth]{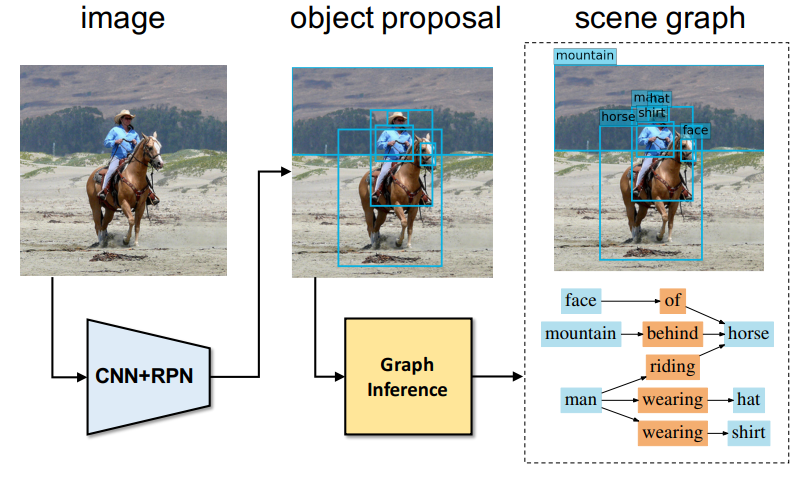}
  \caption{The model in \citep{xu2017scene} takes an image as input, and the output is a scene graph that consists of object categories, their bounding boxes, and semantic relationships between pairs.}
  \label{fig:semanticexample2}
\end{figure}

\textcolor{black}{Semantic context corresponds to the likelihood of an object to be found in some scenes but not others. In terms of spatial context, semantic context can provide constraints of spatial relations between objects in a scene. Objects have typical environments, such as a bath tub in a bathroom or a bed in a bedroom. Sources of semantic context in early works were obtained from common expert knowledge \citep{strat1991context} which constrained the recognition system to a narrow domain and allowed just a limited number of approaches to deal with uncertainty of real world scenes. On the other hand, annotated image databases \citep{wolf2006critical} and external knowledge bases \citep{rabinovich2007objects} can deal with more general cases of real world images. A similar evolution happened when learning semantic relations from those sources: pre-defined rules were replaced by approaches that learned the implicit semantic relations as pixel features \citep{wolf2006critical} and co-occurrence matrices \citep{rabinovich2007objects}. Semantic context are heavily used in scene graph generation tasks \citep{xu2017scene, zellers2018neural, johnson2018image} to describe the spatial relations between different objects (Fig. \ref{fig:semanticexample2})}. 

\begin{figure}[ht]
  \centering
  \includegraphics[width=0.9\linewidth]{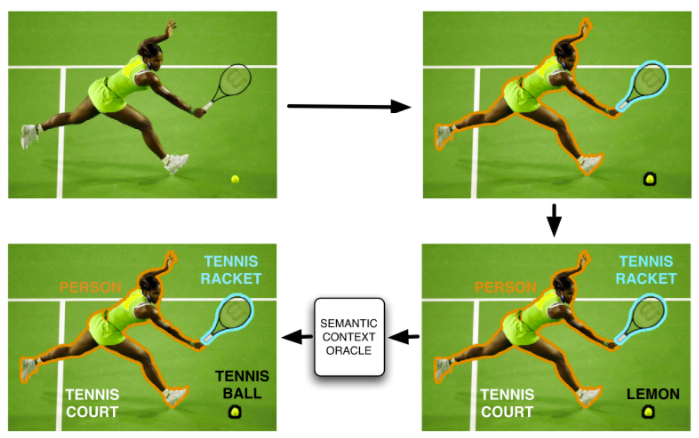}
  \caption{ Objects are perfectly segmented by using contextual relations between objects’ labels to help satisfy semantic constraints \citep{rabinovich2007objects}.}
  \label{fig:semanticexample}
\end{figure}

\textcolor{black}{Spatial semantic context can be obtained from strongly labeled training data. Several works \citep{sun2017seeing, xu2017scene, zellers2018neural, yang2019step} employed semantic context for scene graph generation tasks. They state that even though a object detector can detect all the objects appears in a scene, it still cannot understand the semantic relationship. Rich semantic context can indicate the specific spatial relations between objects, and result in a deeper understanding of visual scene. These semantic context can indicate the co-occurrence of the objects in a scene. If we know there is a horse rider in a image, we can easily build up the spatial relations, such as co-occurrence (there is a rider and a horse in the image) and 2D spatial relations (The rider is on the house; There might be a helmet on the rider's head.). In Fig. \ref{fig:semanticexample2}, the model proposed in \citep{xu2017scene} takes an image as input, and the output is a scene graph that consists of object categories, their bounding boxes, and semantic relationships between pairs. Instead of inferring each object in isolation during training, the model passed rich semantic context information to refine its predictions. Another work \citep{chen2019multi} performs an multi-label image classification task by using label co-occurrence in training data as a semantic context prior, to model the relationship between each object category.}

\textcolor{black}{Spatial semantic context in non-visual forms of scenes, events or the presence of other objects can also help in predicting the presence of an object. %As mention in 2D Spatial Representations (Section \ref{subsubsecspatial2d}), semantic context can describe the spatial relations between different object or event more clearly. Spatial semantic context plays an important role of reducing ambiguity in objects' visual appearance. 
As shown in Fig. \ref{fig:semanticexample}, Rabinovich's work \citep{rabinovich2007objects} shows that mis-labeled "Lemon" is refined to correct "Tennis" by enforcing semantic contextual constraints (in a scene of tennis match). To enforce spatial semantic context, the author uses external knowledge obtained from Google Sets (rather than visually from the image)  to generate semantic context constraints among object categories. Palmer \citep{Palmer} examined the influence of prior presentation of visual scenes on the identification of briefly presented drawings of real-world objects. He found that observers' accuracy at an object-categorization task was facilitated if the target (e.g. a loaf of bread) was presented after an appropriate scene (e.g. a kitchen counter) and impaired if the scene-object pairing was inappropriate (e.g. a kitchen counter and bass drum).}

\begin{figure}[ht]
  \centering
  \includegraphics[width=1\linewidth]{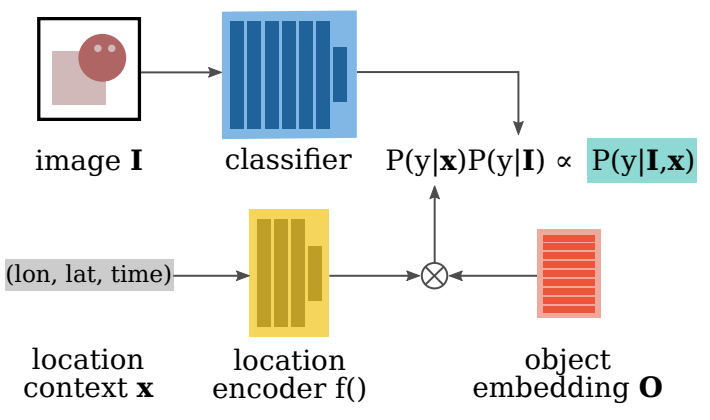}
  \caption{A framework of using geographic location as a spatial prior (from \citep{mac2019presence}).}
  \label{fig:spatialexample}
\end{figure}

\textcolor{black}{Geographic information can also be served as a spatial semantic context in certain tasks. Aodha et al. \citep{mac2019presence} proposed an approach to use geographical location as a spatial prior, to estimates the probability that a given object category occurs at that location. Fig. \ref{fig:spatialexample} shows the framework of \citep{mac2019presence}. In order to identify whether species are present in image, the author use spatial context extracted from metadata (lat, long, day). Further they use Bayesian model as the basic model and a different network to process image classification and spatial-temporal context from metadata.}

\subsection{Temporal Context} \label{subsectemporal}
In common sense, temporal context can be interpreted as the information in a video, such as nearby frames, previous clips or video captured recently. Many works use the temporal context within a video to improve the performance. However, for some computer vision tasks, such as species classification\citep{mac2019presence}, animal movement\citep{beery2020context}, temporal context from nearby frames or a recent video is not sufficient, a longer temporal context (over months or years) will be needed to help with these tasks. The longer temporal context sources can provide useful information, such as movement pattern of the species across different time periods, which will better indicate the presence of the objects in the scene. We first review temporal context in two categories: temporal context in videos and temporal context across months. Many works are focused on video by using nearby frames as temporal context. Temporal context across months are less used in computer vision tasks.  \textcolor{black}{We further review temporal semantic context where temporal information is provided mostly in non-visual forms to serve as a temporal cue for the task. In the following, we will describe different works in details in each of the three categories.} 

\subsubsection{Short-Term Temporal Context (in Videos)}
\begin{figure}[ht]
  \centering
  \includegraphics[width=1\linewidth]{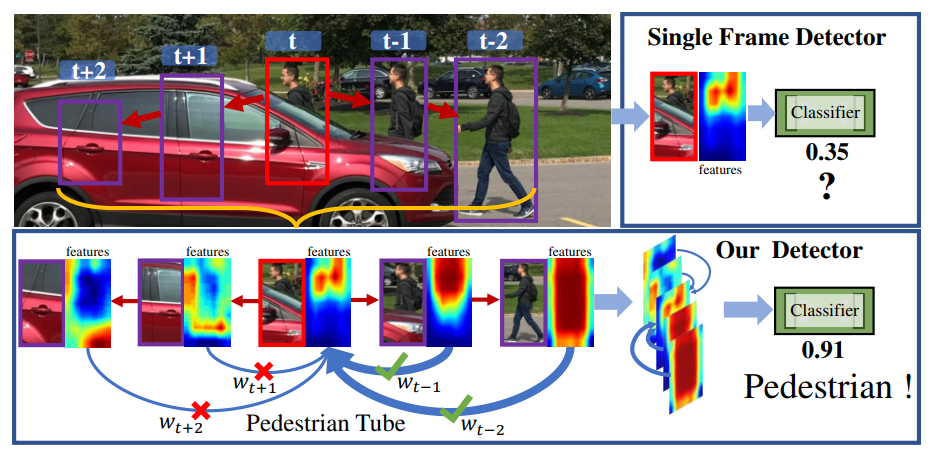}
  \caption{Top row: A heavily occluded pedestrian often leads to miss detection for a single frame detector due to incomplete and weak observations. Bottom row: local temporal context of a heavily occluded pedestrian is exploited. The figure was presented in \citep{Wu_2020_CVPR}.}
  \label{fig:Wu_2020_CVPR_1}
\end{figure}

Short-term temporal context refers to temporally proximal information, such as nearby frames of a video, images captured right before/after the given image, or video data from similar scenes and time of capture \citep{divvala2009empirical}. Sometimes pure temporal context cannot provide enough information, therefore other context cues such as spatial context and semantic context, are used simultaneously with temporal context. Temporal cues has been employed widely in video related tasks \citep{Wu_2020_CVPR, wang2015video, wang2016hierarchical, yan2019learning, yang2019step}. Since nearby frames of a video may have a better feature representation of the target, a recent work \citep{Wu_2020_CVPR} investigates how to utilize local temporal context to enhance the representations of heavily occluded pedestrians. The key idea is to search for non- or less-occluded pedestrian examples (which is refer to reliable pedestrians) with discriminative features along the temporal axis, and if they are present, to exploit them to compensate the missing information of the heavily occluded ones in the current frame, as shown in Fig. \ref{fig:Wu_2020_CVPR_1}. Since it is difficult to link heavily occluded pedestrians with non-/less-occluded ones, where the appearances of the pedestrians are substantially different, the authors resort to local spatial-temporal context to match pedestrians with different extents of occlusions not only embedding the temporal information, but also the spatial context of different parts between the occluded pedestrian and the less-occluded pedestrian.

\begin{figure}[ht]
  \centering
  \includegraphics[width=0.8\linewidth]{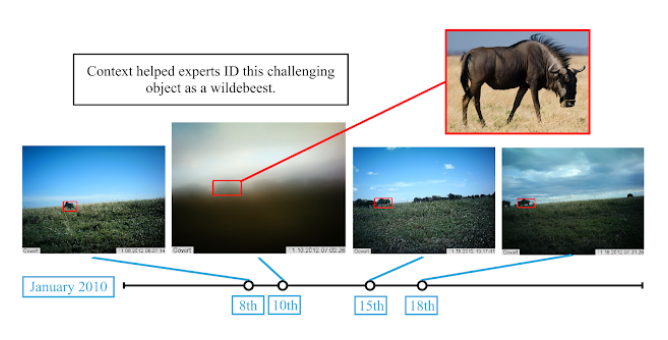}
  \caption{Here, additional examples from the same scene help experts determine that the object is an animal and not background. Context such as the shape \& size of the object, its attachment to a herd, and habitual grazing at certain times of day help determine that the species is a wildebeest. Useful examples occur throughout the month. The figure was presented in \citep{beery2020context}.}
  \label{fig:contextrcnn_example}
\end{figure}

Yan et al. \citep{yan2019learning} use temporal context for the person search task. The authors propose a graph learning framework to employ contextual person pairs from frames contains target person, to model the similarity of the target person. Video event recognition aims to recognize the spatio-temporal visual patterns of events from videos. Wang et al. \citep{wang2015video, wang2016hierarchical} propose a serial work for video even recognition. The authors propose a hierarchical context model built on temporal context. The combination of the prior context, semantic context and feature-level context from previous event provides the temporal support for the prediction for the current event. It also outperforms the existing context approaches, and utilizes multiple level contexts. 

\subsubsection{Long-Term Temporal Context (across Months or Years)}

Long-term temporal context is used as the neighbor frames or a large temporal scale within a video. From a broader perspective of temporal information, temporal context is not limited to nearby frames in a short period of time in video-based tasks, it can also be leveraged in a long period of time such as months or years, which can provide long-term temporal consistency for video-based tasks. These kind of temporal information are used in species recognition tasks \citep{mac2019presence, beery2020context}. Beery et al. \citep{beery2020context} propose Context R-CNN, which leverages temporal context for improving object detection regardless of frame rate or sampling irregularity. Context R-CNN leverages up to a month’s worth of images from the same camera for context to determine what objects might be present and identify them. An example is shown in Fig. \ref{fig:contextrcnn_example} \textcolor{black}{for determining the species in challenging data where the animal in images is small and hard to identify}. It takes advantages of the high degree of correlation within images taken by a static camera to boost performance on challenging data and improve generalization to new camera deployments without additional human data labeling. Aodha et al. \citep{mac2019presence} introduce a spatio-temporal framework that jointly models spatial context (the relationship between location), temporal context (time of year) and semantic context ( photographer, and the presence of multiple different object categories). At test time, given an image and where and when it was taken, the model aims to estimate which category it contains i.e. $P(y|I, x)$. This framework incorporates with the long term temporal context (months to years), which carries rich historical information about the species, to help the model successfully distinguish the species with similar appearance.

\subsubsection{Temporal Semantic Context} \label{subsectempsemantic}

\begin{figure}[ht]
  \centering
  \includegraphics[width=1\linewidth]{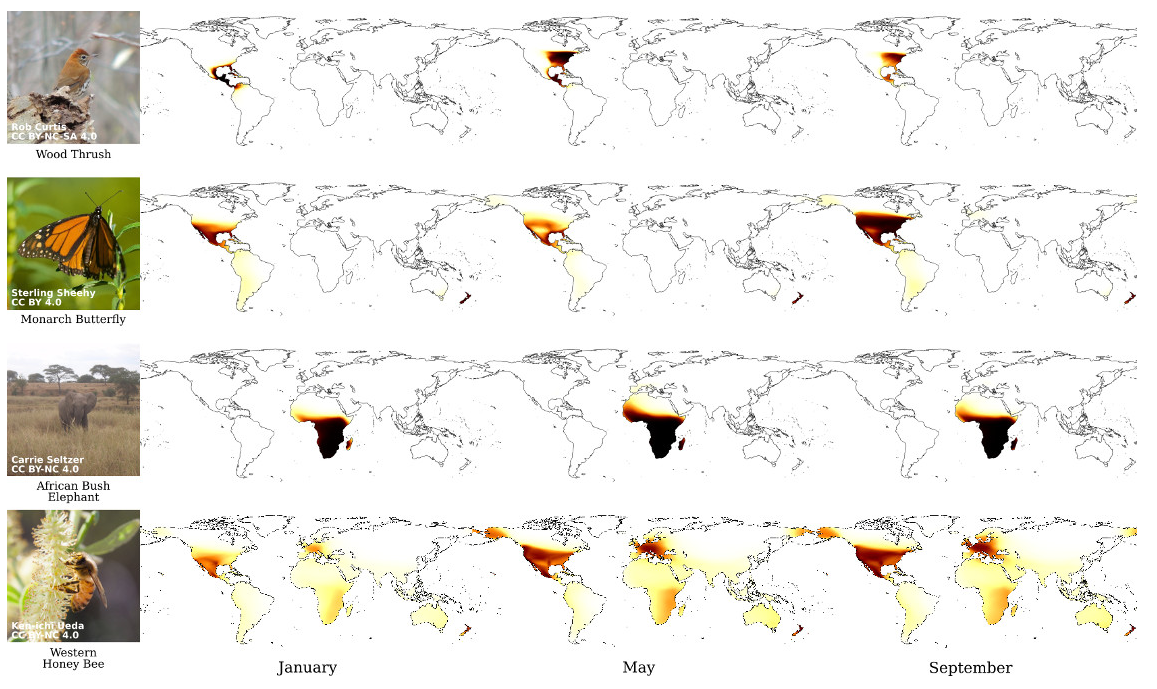}
  \caption{The example of \citep{mac2019presence}. The framework uses temporal information extracted from metadata up to a year to help predict the species present in the image.}
  \label{fig:temporalsemanticexample}
\end{figure}

\textcolor{black}{Semantic context can also be temporal information. These context are usually provided by the dataset or embedded in the metadata. The iNatualist dataset \citep{van2018inaturalist} consists not only the species' images, but also comes along with descriptions, locations, time and dates, and observer identifications, which are embedded in the metadata. The time and date information can be served as a temporal prior to help identify the species in the image, and also track the movement of the species. A work \citep{mac2019presence} uses up to a year of the iNaturalist data to help recognize the species in specific location, and also track the movement of the species. An example is shown in Fig. \ref{fig:temporalsemanticexample}.}

\begin{figure}[ht]
  \centering
  \includegraphics[width=1\linewidth]{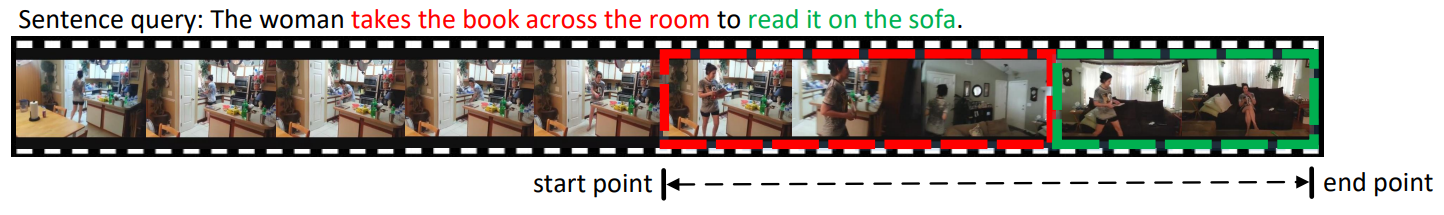}
  \caption{An example of temporal grounding in videos task. The semantic context indicates the temporal boundaries between two activities. The figure was presented in \citep{yuan2019semantic}.}
  \label{fig:temporal_semantic_example_1}
\end{figure}

\textcolor{black}{Semantic context can also serve as temporal cue to help find activities in video task. Yuan et al. \citep{yuan2019semantic} use semantic context to determine the temporal boundaries in temporal grounding in videos task. As shown in Fig. \ref{fig:temporal_semantic_example_1}, there are two activities in the video sequence: Woman walks cross the room and woman reads the book on the sofa. Without referring the semantic description, these two activities are difficult to be associated as one event. As the semantic description provided: The woman takes the book across the room to read it on the sofa, human can easily correlate the two activities together and determine the temporal boundaries accurately. Furthermore, the semantic guidance is also important for activity localization in the videos, due to diverse visual appearances and different temporal scales. In these situation, semantic context serves as temporal indicator to help correlate relevant video segments over time.}

\subsection{Other Context} \label{subsecother}

\textcolor{black}{
As mentioned in spatial context (Section \ref{subsecspatial}) and temporal context (Section \ref{subsectemporal}), semantic context usually provides constraints of the presence of the objects in a scene. For example, if we know the event a basketball match, we are expecting to see certain objects present in a certain scene: Basketballs and basketball stands in a basketball court. We are expecting snow in winter. These kind of semantic context also indicates the spatial information (Section \ref{subsecspatialsemantic}) and temporal information (Section \ref{subsectempsemantic}). On the other hand, there are also other semantic context that are neither spatial nor temporal. Context in other relations such as functionalities, purposes or intention can indicate the occurrence of certain actions or objects. There are also contextual information from (or for) other modalities, such as audio, text, thermal and weather, etc, which can be helpful in computer vision tasks.}

\textcolor{black}{
In this section, we categorize other contexts into other semantic context, context in other relations and context in other modalities. semantic context can be neither spatial context nor temporal context. Other semantic context only describes the dependency of the objects without any spatial information or temporal information. There are also context in other relations and other modalities, which can also provide critical cues in computer vision tasks. Different works are reviewed in each of the three categories.}

\subsubsection{Other Semantic Context}

\begin{figure}[ht]
  \centering
  \includegraphics[width=1\linewidth]{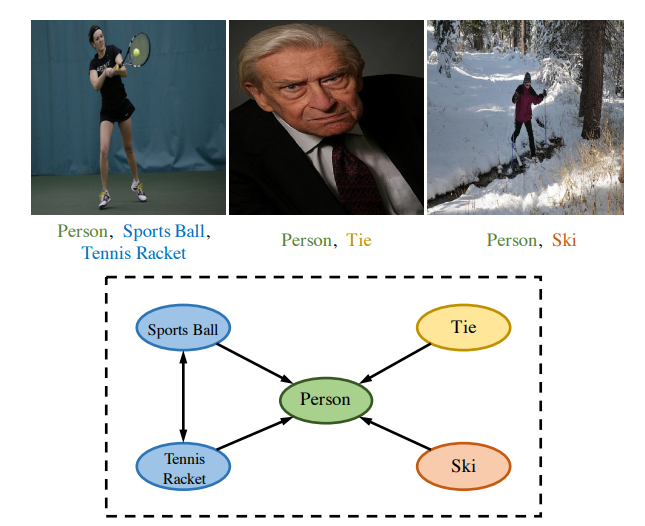}
  \caption{A directed graph is built over the object labels to model label dependencies in multi-label image recognition. The figure was presented in \citep{chen2019multi}.}
  \label{fig:other_semantic_example_1}
\end{figure}

\textcolor{black}{
There are other semantic contexts which are neither spatial context or temporal context. These kind of semantic context only indicate the presence of the objects, without any spatial information or temporal information. A work \citep{chen2019multi} in a multi-label image recognition task uses the label dependency to model the semantic relations. As shown in Fig. \ref{fig:other_semantic_example_1}, the author uses the labels in training dataset to build the relationship between each object in the image, without using any spatial or temporal information. The labels in training dataset provides the exclusive semantic relations of the objects without knowing the location of the object or the scene of the image.} 

\textcolor{black}{
Another work \citep{zhu2021semantic} in an object detection task uses semantic space projection to model the semantic relation to aid the learning of the visual information of the objects. The framework builds a semantic space from the word embeddings of all classes in the dataset, then projects the learned visual feature of the object into the semantic space to enhance the detection accuracy. Instead of using  spatial relations between objects, the framework only uses semantic context without any spatial or temporal information to build the semantic space.}

\subsubsection{Context in Utilization}

\begin{figure}[ht]
  \centering
  \includegraphics[width=0.9\linewidth]{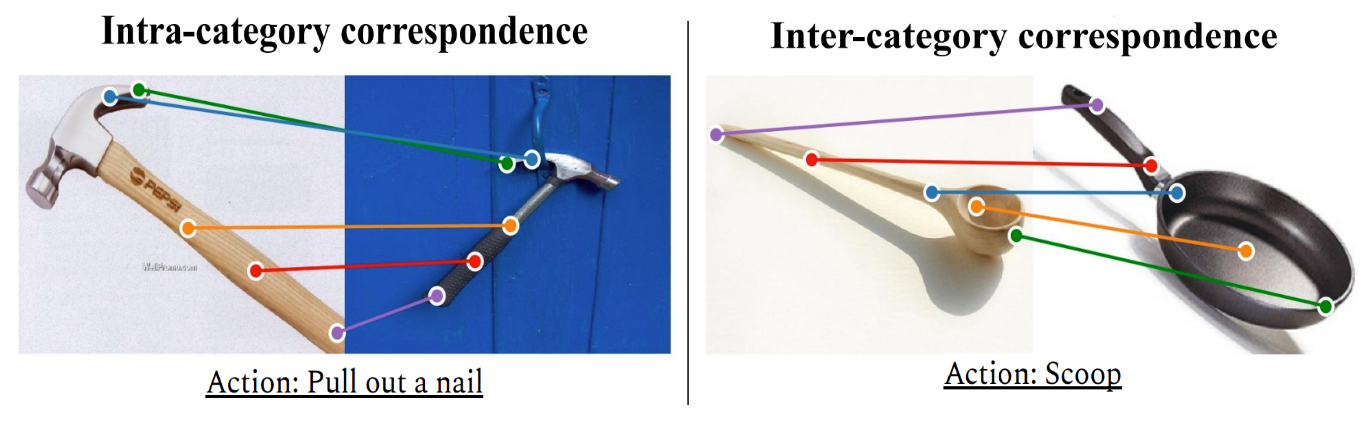}
  \caption{An example of intre-category functional correspondence and inter-category correspondence. The figure was presented in \citep{lai2021functional}.}
  \label{fig:other_relation_example}
\end{figure}

\textcolor{black}{
The function of a scissor is cutting. A cup is used to hold drinks. These kind of context provide the functionality or purposes of an object or an action. Even though they are important, these relations are less used in computer vision tasks. A recent work \citep{lai2021functional} introduces the problem of functional correspondence, which is aimed to find the set of correspondence between two objects for a given task. Any two objects that can be used to perform an action are then used to establish a correspondence relationship. As shown in Fig. \ref{fig:other_relation_example}, since both hammers (intra-category) can be used to pull out a nail, functional correspondence relationship could be established between the two objects. Same functional correspondence can be generated to inter-categorical objects like a spoon and a frying pan, since both can be used to scoop things. Human are good at predicting secondary functionalities (e.g. screwdriver can be used to clean paper jam in a printer), beyond the primary functionalities (e.g. screwdriver is used for screwing). Modeling functional correspondence using the context of functional relations could help in predicting novel use of the objects.}

\begin{figure}[ht]
  \centering
  \includegraphics[width=0.8\linewidth]{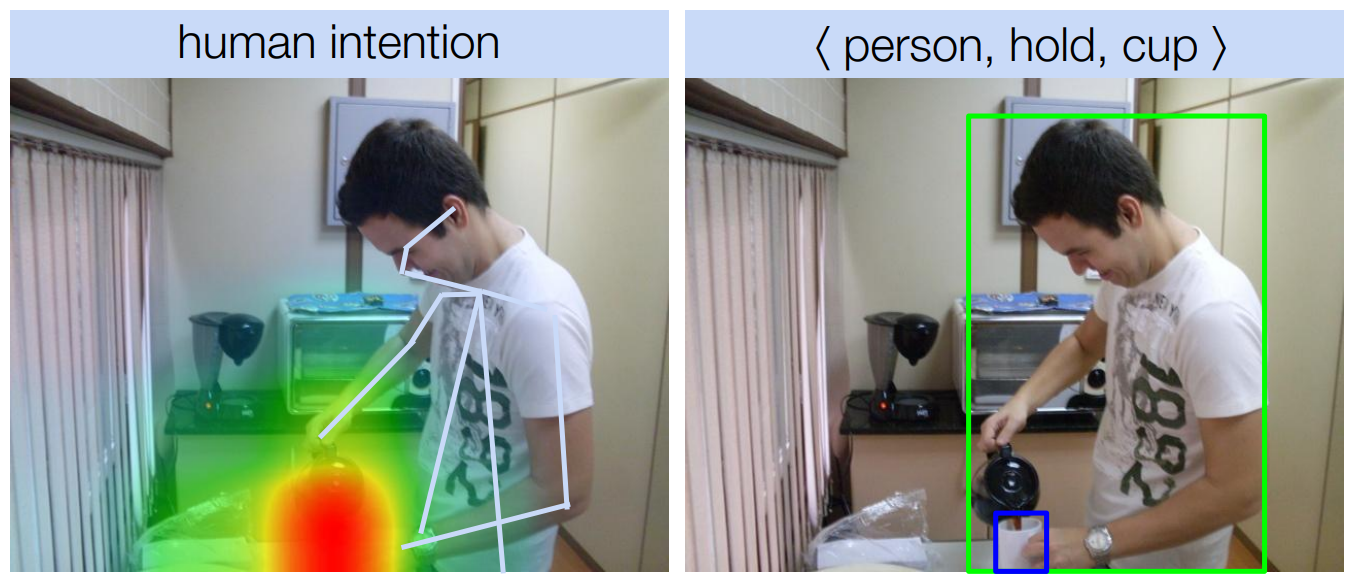}
  \caption{An example of the human intention of the human-object interaction. The figure was presented in \citep{xu2019interact}.}
  \label{fig:other_relation_example_2}
\end{figure}

\textcolor{black}{
Another work \citep{xu2019interact} uses human intention for detecting human-object interactions (HOIs) in social scene images. Human usually direct their attention and move their body based on their intention. The intention is also informative of the human-object interactions. An example is shown in Fig.\ref{fig:other_relation_example_2} where the intention is very informative for understanding the HOI. The person in the scene is fixating at the HOI regions around the cup that he is interacting with. Furthermore, his posture implicitly conveys his intention. The work leverages the context regions and incorporate the human pose information using spatial context (the relative distances from body joints to the instances).}

\subsubsection{Context in Other Modalities}

\begin{figure}[ht]
  \centering
  \includegraphics[width=0.8\linewidth]{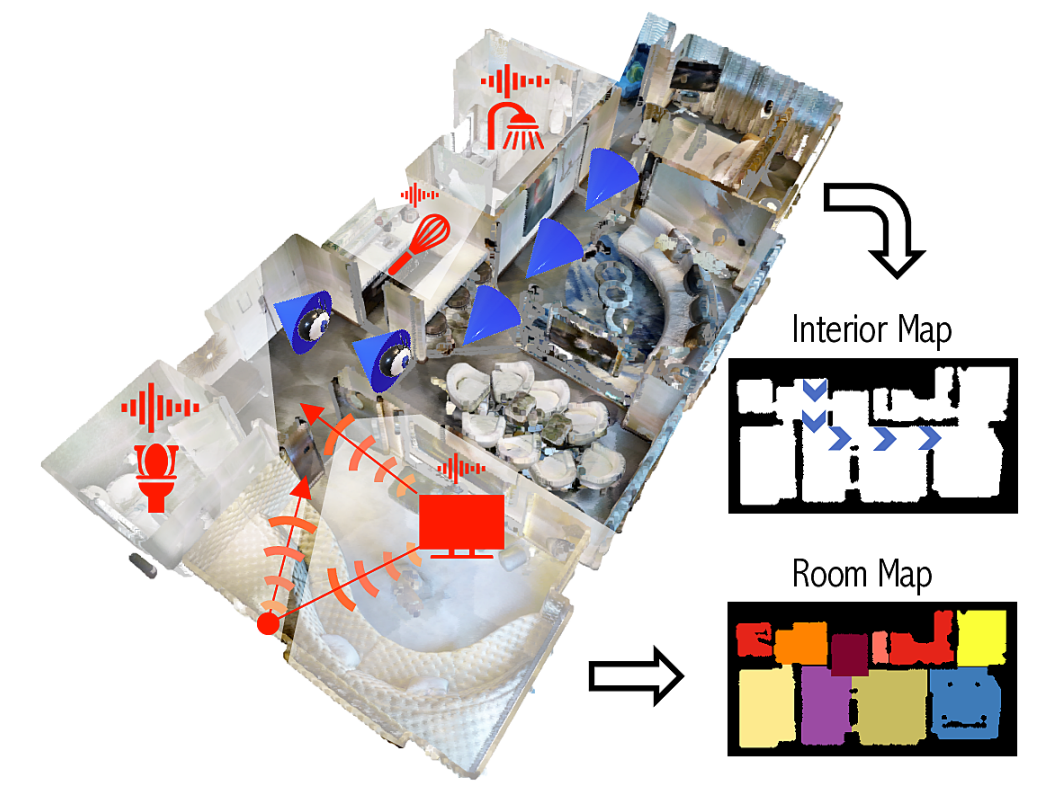}
  \caption{An example of audio-visual floorplan reconstruction, where sounds in the environment help infer both the geometric properties of the hidden areas as well as the semantic labels of the unobserved rooms. The figure was presented in \citep{purushwalkam2021audio}.}
  \label{fig:other_modalities_example_1}
\end{figure}

\textcolor{black}{
There are also context in other modalities, which can be used in computer vision tasks. When we hear a dog barking, we can estimate how far and which direction the dog is located. When we listen to a speech without video, we could also find video segments for the speech through lip reading, since the correlations between speech (audio) and lip movements provide a strong cue for linguistic understanding. Audio has been used in event localization task \citep{tian2018audio} and floorplan reconstruction task \citep{purushwalkam2021audio}. An example from \citep{purushwalkam2021audio} is shown in Fig. \ref{fig:other_modalities_example_1}. The authors use a short video walk through the house to reconstruct the visible portions of the floorplan, without going into each rooms. Audios from different rooms are used as a context to infer the geometric properties of the blind areas as well as the functionality of the rooms (e.g., washing machine sounds behind a wall to the camera's right suggest the laundry room).}

\textcolor{black}{
Besides audio, thermal can also be informative. When people drive in a dark road, it is hard to see wild animals and pedestrians on the roadside. A thermal sensor combined with the visual camera can easily detect and locate them, to lower the risk of causing accident. Thermal can also provide critical information for wildlife management, such as estimating animal populations. A work \citep{seymour2017automated} employs thermal as the context along with imagery to estimate seals in eastern Canada. The proposed method improves upon shortcomings of computer vision by effectively recognizing seals in aggregations while keeping minimum model setup time.}

\begin{figure}[ht]
  \centering
  \includegraphics[width=1\linewidth]{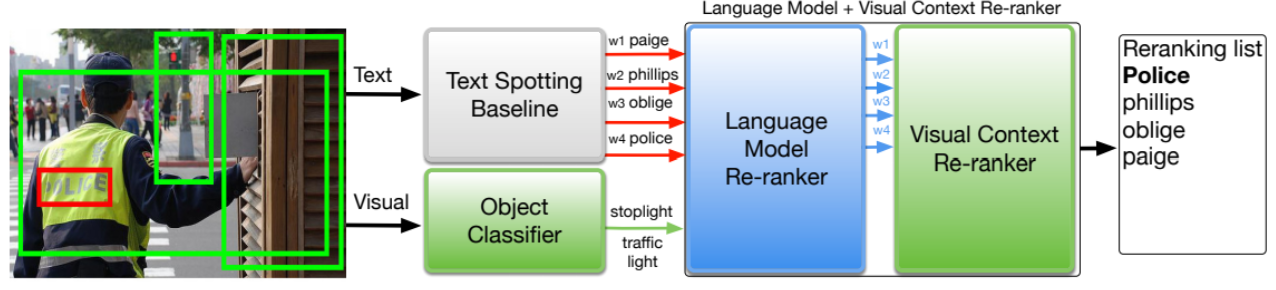}
  \caption{The semantic framework of \citep{sabir2018enhancing}. The word “police” is correctly ranked in the first position of the most likely predicted words based on the semantic relation between visual context information (traffic light) and the spotted words.}
  \label{fig:semanticexample3}
\end{figure}

\textcolor{black}{
Visual information can also be context for other modalities. Sabir et al. \citep{sabir2018enhancing} use semantic relations between text and visual context information to perform a text recognition task in natural scene. The author first uses a off-the-shelf text detector to propose the predicted text, and then uses a object classifier to predict objects appears in the scene. Visual context serves as a prior to refine the rank of the predicted text based on the semantic relation between visual context and predicted words. As shown in Fig. \ref{fig:semanticexample3}, the final top rank word \textit{police} is biased by the visual context information \textit{stoplight}. }

\subsection{Summary of Major Context Types}

\begin{figure}[ht]
  \centering
  \includegraphics[width=1\linewidth]{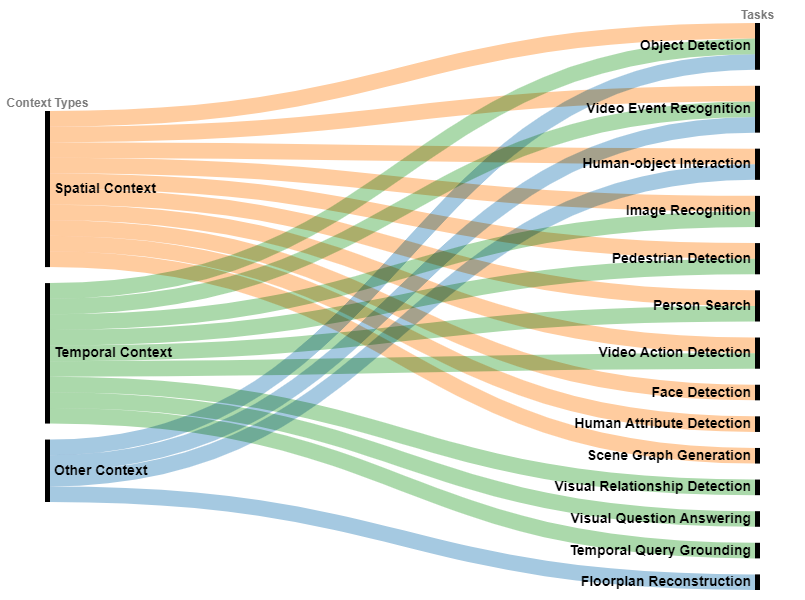}
  \caption{Summary of the three major context types and what tasks they are employed in.}
  \label{fig:context_summary}
\end{figure}

\textcolor{black}{
In this section, we mainly review three major types of context: spatial context, temporal context and other context. Spatial context can be defined as the likelihood of finding an object in some positions and not others with respect to other objects in the scene. We further split the spatial context representations into three categories: co-occurrence, 2D spatial relations, and semantic relations. Temporal context refers to temporally related information and it can be separated into a short term, a long term, or a semantic relation over time. In general, semantic context corresponds to the likelihood of an object to be found in some scenes but not others. Semantic context can be both spatial context and temporal context across time, and can also be neither of them. Context in utilization can reveal the functionalities or purposes of the objects or the actions. Context in other modalities, such as audio and thermal can also be informative for some computer vision tasks. All these types of context are employed in various combinations in existing context based approaches. Fig. \ref{fig:context_summary} shows the major context types and the related tasks when employing context information.}

\section{Levels of Context} \label{contextlevel}

\begin{figure}[ht]
  \centering
  \includegraphics[width=0.85\linewidth]{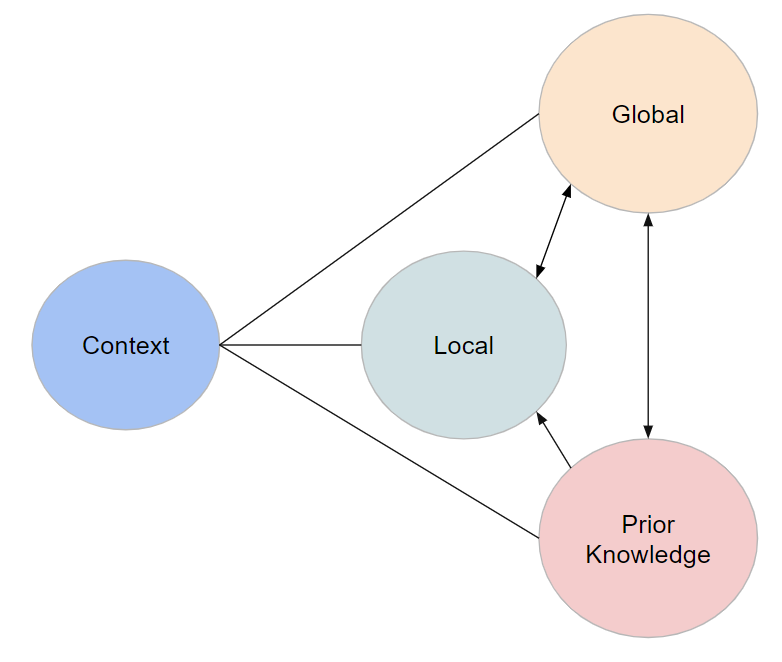}
  \caption{Three different context levels and their relations.}
  \label{fig:context_levels}
\end{figure}

Context can also be represented in different levels. We separate context into three levels: prior knowledge level, global level and local level. Prior knowledge refers to the knowledge obtained \textit{before} seeing the scenes or events, such as location, time and weather etc., serves as \textit{a prior} knowledge for computer vision tasks. Global context exploits the visual scene as global information, which could provide context such as spatial layout and semantic relations between objects. Local context contains the intrinsic context (of the object itself) and the extrinsic context (surrounding regions or objects of the target). As shown in Fig. \ref{fig:context_levels}, global level context can include local level context from the object, which could be further extracted from the local regions. Prior knowledge can serve as \textit {a prior} for a global scene, and global context can also be in the form prior knowledge to indicate the occurrence of the object or the event. Prior knowledge can also provide important information for local context, which can serve as a cue for computer vision tasks such as object recognition and object detection. In this section, we provide details for each context level and how they are employed in different computer vision tasks. 

\subsection{Prior Knowledge Level} \label{priorlevel}
\textcolor{black}{Context at the prior knowledge level refers to the knowledge obtained before seeing the scenes or events. It reflects the environment such as location and time, that can serve as prior to predict whether certain events would occur or certain object would be detected in the visual scene. For example, if we know there are a hotel building and a bus stop in the scene before we see the scene, we can easily guess the text appeared on a hotel building is probably different from the text appeared on a bus stop advertisement. These context information are treated as high level context information. Furthermore, context between neighboring images can also provide high level information. Both context will provide prior information for the inference of the task. Prior knowledge may not be directly extracted in the image or video in consideration. It may come from previous event for temporal support or metadata, which will serve as prior information to analyze the current scene.} 

\begin{figure}[ht]
  \centering
  \includegraphics[width=1\linewidth]{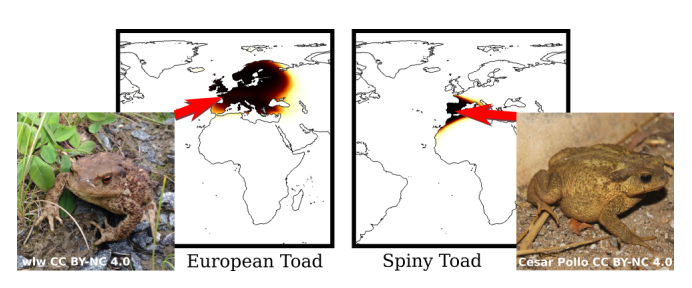}
  \caption{Differentiating between two visually similar categories such as the European toad (left) and Spiny toad (right)  can be challenging without additional context. To address this problem,  a spatio-temporal prior has been proposed that encodes where, and when, a given category is likely to occur. For a known test location, the prior predicts how likely it is for each category to be present. Darker colors indicate locations that are more likely to contain the object of interest. The figure was presented in \citep{mac2019presence}.}
  \label{fig:priorcontext_example}
\end{figure}

Events happen in a parking lot are different from events happen in a playground, a slide should appear in the playground, a car should appear in the parking lot , etc. Prior knowledge from the whole scene can provide a prior probability of the events. A series of works  \citep{wang2016hierarchical, wang2015video, wang2012incorporating} on video event recognition employ prior level context from two aspects: the context from current scene, and temporal support from previous event. Prior context from current scene reflects the environment such as locations (e.g. a parking lot, a shop entrance) and times (e.g. at noon, in dark) that can serve as prior to dictate whether certain events would occur. Prior context from previous event can provide temporal support for the prediction of the current event. These prior context provides critical cues for event recognition task.

Prior level context can be vary in different scenes. For example, we can easily expect that there may be chairs or tables in an office. On the other hand, we expect traffic lights, roads and buildings in an outdoor scene. Spatial priors can be another prior level context. For example, street lights along roads usually have regular shapes and sizes, which informs that local context of the street lights (appearances, objects surround traffic lights, etc.) could correlate.

Prior level context can also come from other sources, like metadata, scene description, etc. than the whole scene itself. As shown in Fig. \ref{fig:priorcontext_example}, the work \citep{mac2019presence} uses geographic location information extracted from metadata as spatial information, and it also serves as prior knowledge for the species recognition. These kind of prior knowledge can also provide useful context for recognizing the species appears in certain areas. 

\subsection{Global Context Level} \label{globalcontextlevel}
\begin{figure}[ht]
  \centering
  \includegraphics[width=1\linewidth]{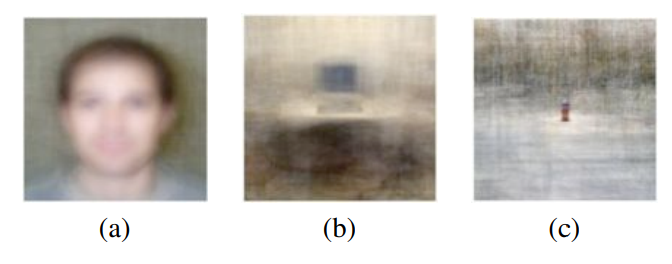}
  \caption{The structure of objects and their backgrounds. The figure was presented in \citep{torralba2003contextual}. Each image has been created by averaging hundreds of images containing a particular object in the center (a face, a keyboard and a fire hydrant) at a fixed scale and pose. The averages can reveal the regularities existing in the color/brightness patterns across all the images. However, this regularities is only visible for the keyboard in (b).}
  \label{fig:globalcontext_example}
\end{figure}

Global context exploits scene configuration (image as a whole) as an extra source of global information across categories. The structure of a scene image can be estimated by the mean of global image features, providing a statistical summary of the spatial layout properties. For example, a keyboard should appear next to the monitor in Fig. \ref{fig:hvexample}. Rabinovich's study \citep{rabinovich2009scenes} shows that by incorporating the statistics of the background, context becomes a global feature of the object category. For example, refrigerators usually appear in a kitchen, thus the usual background of refrigerators is similar. Having learned such a global feature of an object category, one can infer a potential object label: if the background resembles a living room, then the patch of interest may be a TV. The background or scene provides a likelihood of finding an object in the scene (e.g. it is unlikely to find a car in living room). It can also indicate the relative positions at which an object might appear (e.g. car on the road, pedestrians on walkways, etc.)

The knowledge of global context can help recognize individual objects in a scene, but identifying individual objects in global context can also help estimate the scene category. One way to incorporate this idea in object recognition is to learn how likely each object appears in a specific scene category. By using the global context information, object recognition performance can be greatly enhanced. However, there will be limitation for current human selected scene dataset. One example is the Cityscapes dataset \citep{Cordts2016Cityscapes}. The main source of the dataset is the street view from cities located in Europe. It is good for different computer vision tasks such as pedestrian detection and semantic segmentation. For real world applications, the global context is limited, and sharing the context among similar scenes such as streets in different cities, can be challenging. For example, there will be very different global context features between locations (Europe and America), and the structure of the streets could be very different.

\begin{figure}[ht]
  \centering
  \includegraphics[width=1\linewidth]{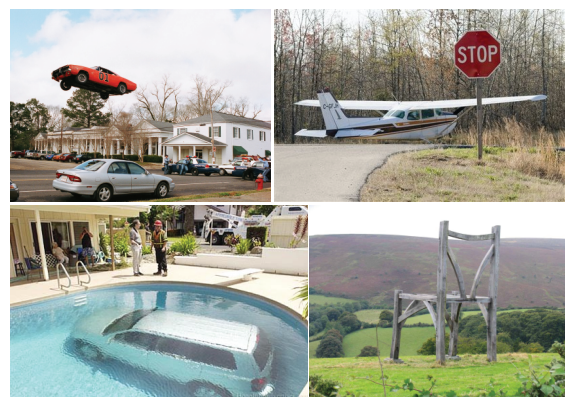}
  \caption{Examples of objects out of context (violations of support, probability, position, or size). The figure is presented in \citep{choi2012context}.}
  \label{fig:globalcontext_example2}
\end{figure}

If you see a microwave oven, the location is most likely in a kitchen, and if you are in a kitchen, you will have a high probability to see a microwave oven. The work by Oliva and Torralba \citep{torralba2003contextual} discusses how context influence on object recognition task. Global context and the objects within it can influence each other. However, some objects (faces, cars, persons, etc.) may have various background scenes based on the locations of their appearance. E.g., a car can be on the road, or in the parking lot. A person can appear indoor or outdoor, day or night. Under these circumstances, the background or the scene cannot indicate the object correctly. The demonstration of this limitation from \citep{torralba2003contextual} is shown in Fig. \ref{fig:globalcontext_example}(a)(c). Another limitation of global context is that it could be misleading if an object present in irrelevant scenes (Fig.\ref{fig:globalcontext_example2}). Choi et al. \citep{choi2012context} present a context model for out-of-context detection, where the object is unusual for a given scene in a image. \textcolor{black}{Another recent work \citep{bomatter2021pigs} introduces a out-of-context dataset and propose a context reasoning model for out-of-context object recognition.}

\subsection{Local Context Level} \label{localcontextlevel}

\begin{figure}[ht]
  \centering
  \includegraphics[width=1\linewidth]{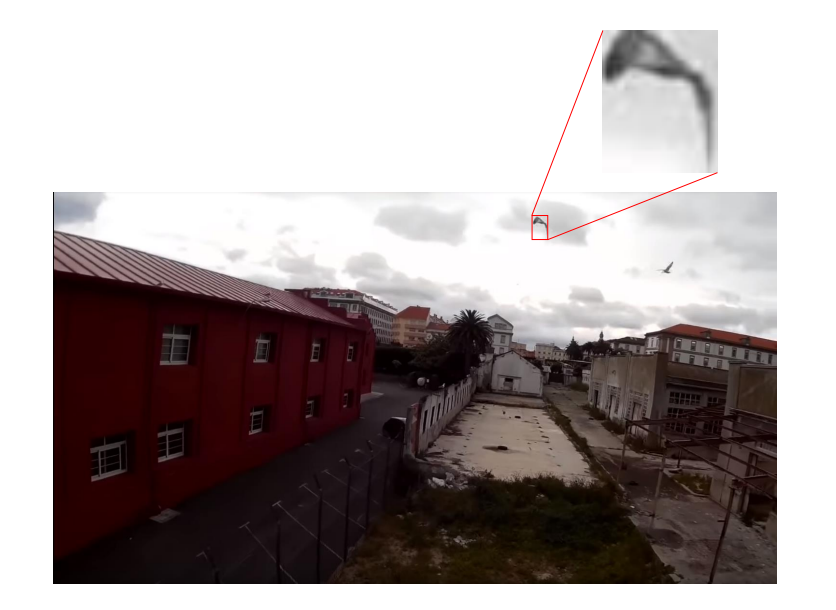}
  \caption{ Context is necessary to recognize small objects such as birds in this picture. The figure was presented in \citep{lim2021small}.}
  \label{fig:localcontext_example}
\end{figure}

Local context indicates the context from objects itself and surrounding local regions, such as color, shape, contrast with background, aspect ratio and other objects etc. Local context features can capture different local relations such as pixel, region and object interactions. As aforementioned in global context level, context could misleading if the object presents in irrelevant scenes. Therefore, rather than measuring global level features, local context can better impose on potential object presence in the image. For big objects like cars, the local feature from object itself like a car will be more obvious even if the car is in a swimming pool (left bottom image in Fig. \ref{fig:globalcontext_example2}). For smaller object like a flying bird, the surrounding area can provide important information in the context of sky (example in Fig. \ref{fig:localcontext_example}).

Local context can also indicate the location of the objects. This information can be captured using spatial context. Spatial context between objects within surrounding area can help because: (1) most objects are supported by other objects, e.g. car is on the road, pedestrians are supported by sidewalk or ground; (2) objects are not appeared in isolation. Objects that have a common function tend to appear nearby and have a certain spatial relation, e.g. a mouse appears next to a keyboard, a dining chair appears besides the dining table, etc; and (3) The structure of the global scene tend to have a common layout, e.g. a stair should appears under a door, and it should appear at at the lower half of the scene. Sky should appear above buildings, and it should appear at the upper half of the scene. Torralba et al. \citep{torralba2003contextual} also show that the vertical spatial relation indicated by local context is usually more informative than the horizontal spatial relation.

Local context can also help with small object detection. It is challenging to detect small objects from complex scenes since small objects have few visual features and simple structures. Current popular detectors mainly improve detection performance by extracting better internal features. Such approaches achieve impressive detection performance on large/medium sized objects, but the detection performance of small objects is still not satisfactory. Exploiting local context in the surrounding scene can be highly beneficial in such cases. When the object is in a simple scene and there is no serious change in the appearance, the object can be well located and identified by its features. However, when the visual information is damaged, ambiguous, or incomplete (e.g. an image contains noise, poor illumination conditions, or the object is occluded or truncated), it is difficult to complete the detection task by relying solely on the own features of the object. At this time, visual context information becomes an important source of information. For smaller object like a flying bird, the surrounding area can provide important information in the context of sky (example in Fig. \ref{fig:localcontext_example}).

Although global context can help indicate the presence of object and spatial representation between objects, if the number of objects increase in the scene, global context cannot discriminate well between scenes, since many objects may share the same scenes, and scenes may look similar to each other. Local context representation is still object-centered and it requires object recognition as a first step, which is different from global context.

\subsection{Summary of Context Levels}

\begin{figure}[ht]
  \centering
  \includegraphics[width=1\linewidth]{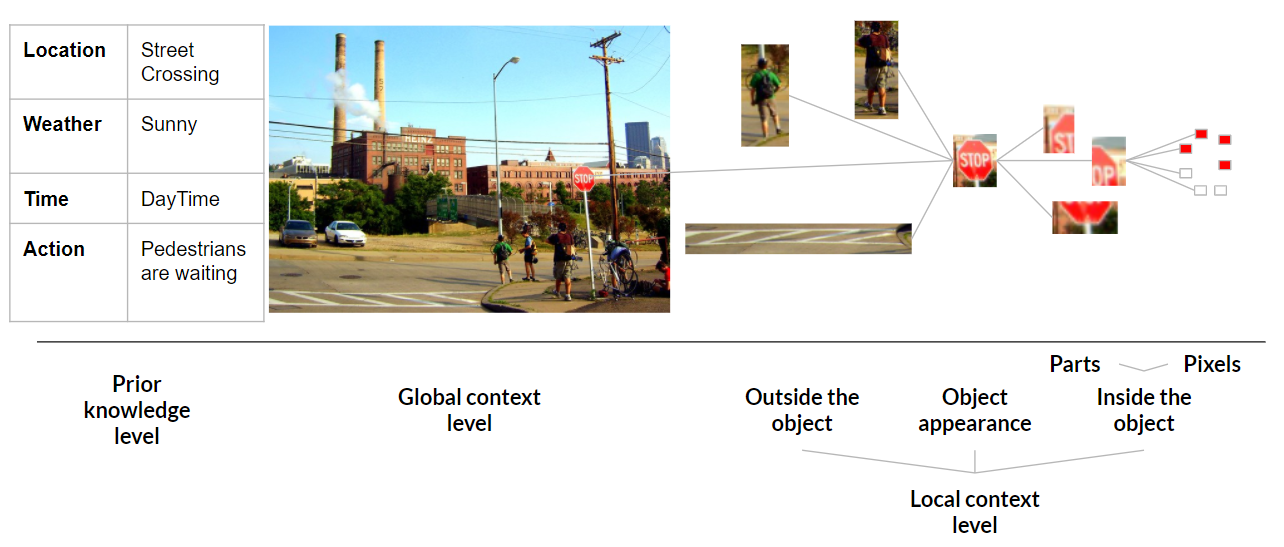}
  \caption{ Summary of context levels. The image in the middle is from COCO \citep{lin2014microsoft} dataset.}
  \label{fig:contextlevelsummary}
\end{figure}

In this section, we review context in three different levels: prior knowledge level, global context level and local context level. Fig. \ref{fig:contextlevelsummary} illustrates these three levels with a real image example. Prior knowledge level refers to the context information obtained usually before seeing the images in consideration. It reflects the environment such as location and time, that can serve as prior to predict whether certain events would occur or certain object would be detected. Global context exploits scene configuration itself (image as a whole) as an extra source of global information across categories. The structure of a scene image can be estimated by the mean of global image features, providing a statistical summary of the spatial layout properties. Local feature level context indicates the context from objects itself and surrounding local regions, such as color, shape, contrast with background, aspect ratio and other objects etc. All there context levels are employed by various approaches in different computer vision tasks, and they all have limitations.

\section{ConvNets in Context Based Approaches}\label{secconvnets}
\textcolor{black}{
Most of the approaches which employ context information are using deep learning methods. Thanks to some pioneers in the machine learning area, convolutional networks come into computer vision researches in both image-based tasks and video-based tasks. Different network architectures were introduced and served as backbones not only for state-of-art context-free approaches, but also for context-based approaches. These deep models are mainly used for feature extraction from images, or relation modeling between object or event categories. Many works integrated multiple context information by either using  existing network architectures, or modified versions of these architectures. In context based approaches, different context types, like spatial context (spatial relation), semantic context (semantic reasoning) have been integrated on these network architectures.
}
\begin{figure}[ht]
  \centering
  \includegraphics[width=1\linewidth]{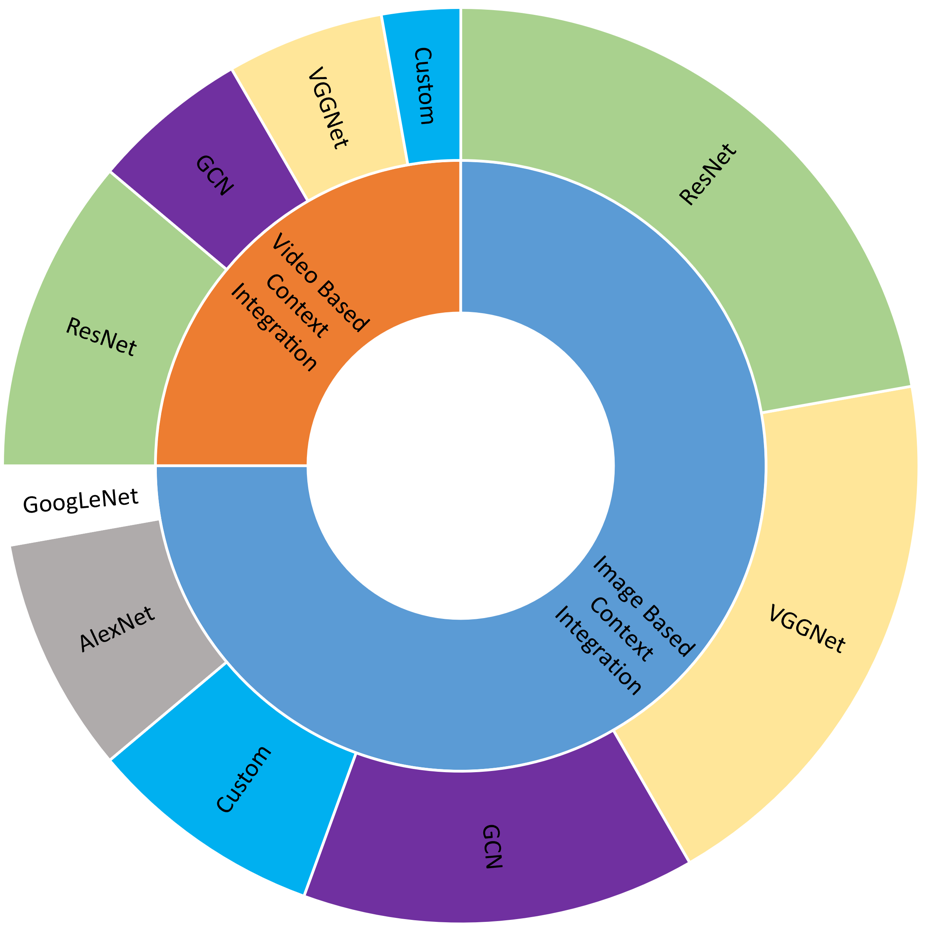}
  \caption{ Summary of convolutional network architectures used in context based approaches. The size of each region shows the proportion of number of works with the named architecture/integration.}
  \label{fig:summaryconvnets}
\end{figure}

The use of different convolutional network architectures in context based approaches is shown in Fig.\ref{fig:summaryconvnets}. Context information are widely integrated in image-based tasks. ResNet is the most used convolutional network architecture in context based approaches. One of the reasons is that ResNet skips the connections of convolutional layers to avoid the gradient vanishing problems. This could extract better context features from the network and benefits the performance of the integration of context. \textcolor{black} {In this section, we review how the base convolutional network architectures have been used in context based approaches in different computer tasks.}

\begin{figure}[ht]
  \centering
  \includegraphics[width=1\linewidth]{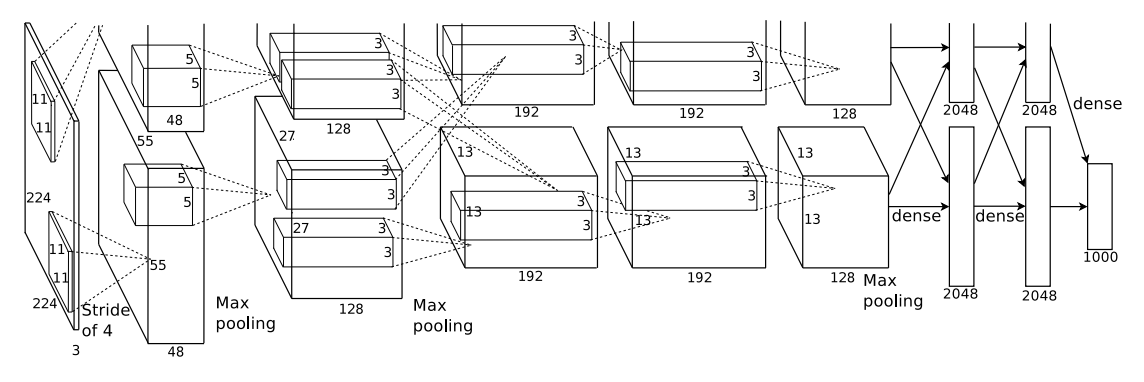}
  \caption{ An illustration of the architecture of AlexNet \citep{krizhevsky2012imagenet}. }
  \label{fig:alexnet}
\end{figure}

\textcolor{black}{
\textbf{AlexNet (2012): }AlexNet \citep{krizhevsky2012imagenet} uses ReLu activation function to accelerated the speed of training. It also uses dropout regularisation to reduce overfitting problem. Another feature of AlexNet is that it overlaps pooling to reduce the size of the network. It is one of the classic convolutional neural network architecture. AlexNet is mostly used when global context (Section \ref{globalcontextlevel}) for whole image feature extraction is applied.
}
\begin{figure}[ht]
  \centering
  \includegraphics[width=1\linewidth]{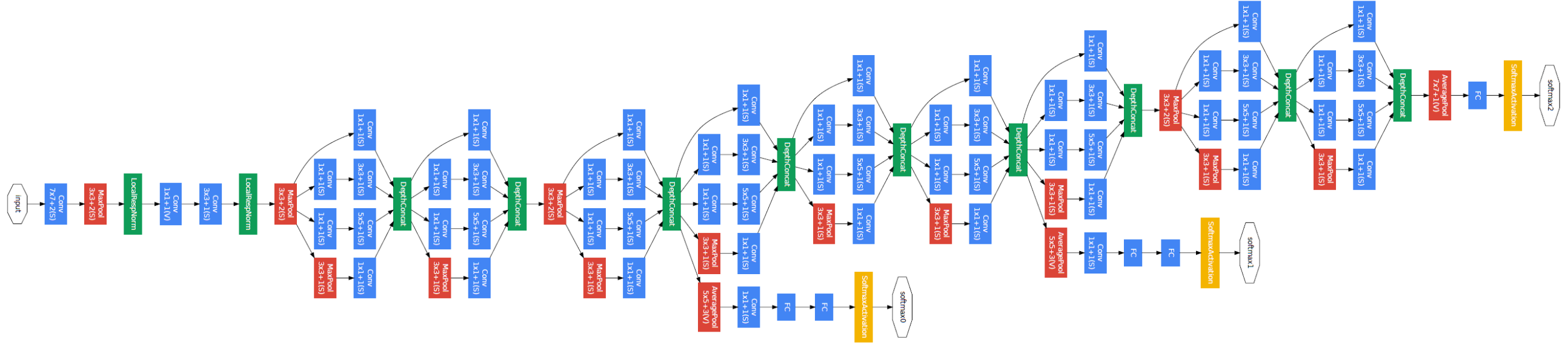}
  \caption{ An illustration of the architecture of GoogLeNet \citep{szegedy2015going}.}
  \label{fig:googlenet}
\end{figure}

\textcolor{black}{
\textbf{GoogLeNet (2014): }GoogLeNet (Fig. \ref{fig:googlenet}) consists of a 22 layer deep CNN architecture. Similar to AlexNet, GoogleNet is mainly used when global context is applied for whole image feature extraction (Section \ref{globalcontextlevel}).
}  
% \begin{figure}[ht]
%   \centering
%   \includegraphics[width=1.0\linewidth]{figures/googlenet_inception.PNG}
%   \caption{ An illustration of the inception module used in GoogLeNet \citep{szegedy2015going}. }
%   \label{fig:googlenet_inception}
% \end{figure}

\textcolor{black}{
\textbf{VGGNet (2014):} VGGNet \citep{simonyan2014very} consists of 13 convolutional layers and 3 fully connected layers in a very uniformed architecture. It is mostly used as a feature extractor from images. VGGNet uses global context (Section \ref{globalcontextlevel}) for image feature extraction and local context (Section \ref{localcontextlevel}) for region feature extraction. VGGNet has been used in context based approaches in different computer vision tasks such as image recognition \citep{zhang2020putting}, object detection \citep{fang2017object} and scene graph generation \citep{xu2017scene}. 
}
\begin{figure}[ht]
  \centering
  \includegraphics[width=1\linewidth]{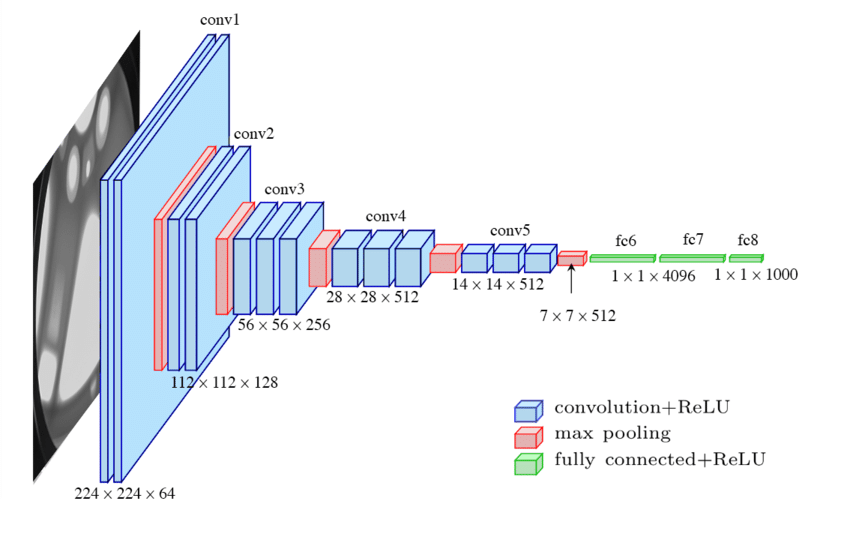}
  \caption{ An illustration of VGGNet \citep{simonyan2014very}. }
  \label{fig:vggnet}
\end{figure}

\begin{figure}[ht]
  \centering
  \includegraphics[width=1\linewidth]{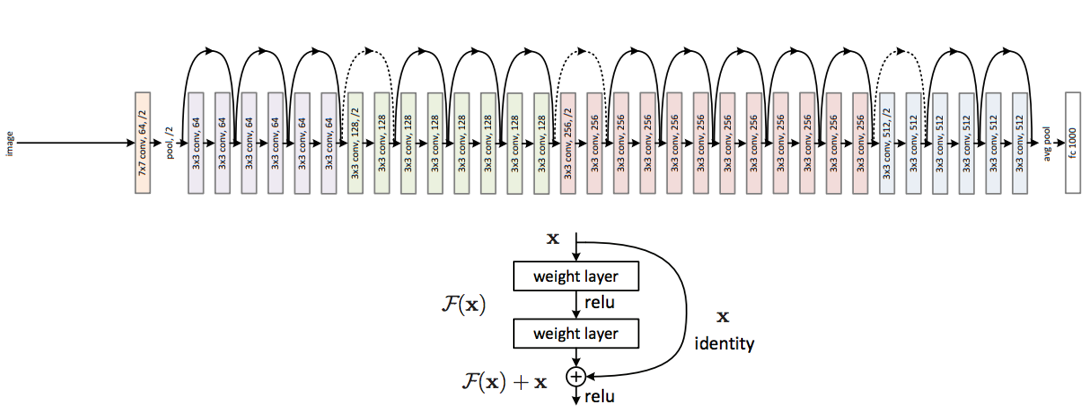}
  \caption{ An illustration of ResNet and its residual block \citep{he2016deep}. }
  \label{fig:resnet_architecture}
\end{figure}

\textcolor{black}{
\textbf{ResNet (2015):} The main idea of ResNet \citep{he2016deep} is to introduce an architecture with skip connections by using a residual block, to overcome the gradient vanishing problems. ResNet with various depth (18 to 152) has been used as a backbone feature extractor in computer vision tasks. It uses global context (Section \ref{globalcontextlevel}) and local context (Section \ref{localcontextlevel}) for image feature extraction and local region feature extraction. ResNet has been widely used in both image-based context integration \citep{yang2019step, dvornik2018modeling, leng2021realize, lim2021small, sabir2018enhancing, zhu2021semantic}, and video-based context integration \citep{beery2020context, Wu_2020_CVPR, yan2019learning}. An Example was shown in Fig. \ref{fig:resnet_architecture}.
}

\textcolor{black}{
\textbf{Graph Convolutional Network (2017):} Kipf and Welling \citep{kipf2016semi} first introduced the Graph Convolutional Network (GCN) to perform semi-supervised classification of nodes in a graph. GCN has also been used to solve computer vision tasks, such as image classification \citep{chen2019multi}, visual relationship detection\citep{cui2018context} and scene graph generation \citep{yang2018graph, johnson2018image}, etc. Because of the unique structure of GCN, it has been used in modelling spatial semantic relation \citep{zhu2021semantic, xu2017scene, zellers2018neural}, 2D spatial representation\citep{yang2019step} and co-occurrence \citep{yan2019learning} in different computer vision tasks, by mainly using spatial context.(Section \ref{subsecspatial}).
}
\begin{figure}[ht]
  \centering
  \includegraphics[width=1\linewidth]{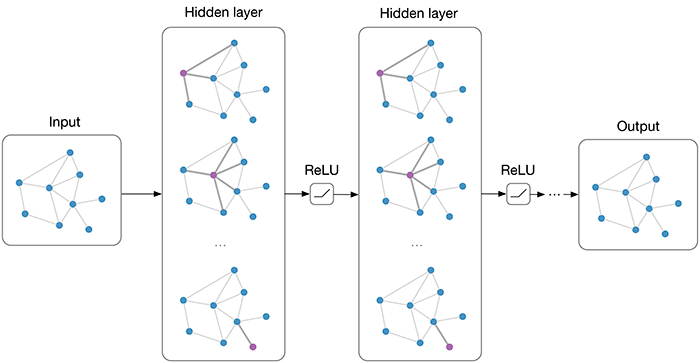}
  \caption{An illustration of Graph Convolutional Network \citep{kipf2016semi}.}
  \label{fig:gcnnetwork}
\end{figure}

%In this section, we reviewed convolutional networks that has been frequently used in context based approaches. These 
\textcolor{black}{In summary these convolutional networks have shown great potential in using context in various computer vision tasks.} However, all the networks have millions of parameters, hence, the network can be easily over fitting when the training data is limited. As shown in Fig. \ref{fig:summaryconvnets}, we pie-charted how the context based approaches use these convolutional network architecture. ResNet and VGGNet are the most used networks in both image-based context integration and video-based context integration. They are always used as the backbone for visual feature extraction. GCN is heavily used in scene graph generation tasks \citep{xu2017scene, zellers2018neural, johnson2018image}, to model the semantic relation of objects in a scene. It has been also used in person search task \citep{yan2019learning}, to model the spatial relation between target person and the person around target. These models can also be pretrained on large-scale labeled datasets to initial parameters for certain computer vision tasks.

\section{Datasets Used in Context Based Approaches} \label{datasets}
This section is the summary of the datasets that has been used in context-based approaches. We separate the datatsets into two categories: image datasets and video datasets. Image datasets have been used in various tasks such as object detection, image recognition, text detection, visual relationship detection, scene graph generation, semantic segmentation and face detection, etc. Video datasets are mainly used for video action recognition, video event recognition, person search, pedestrian detection and object detection, etc. \textcolor{black}{Here we list the two types of datasets in Table \ref{tab:datasets}, with information of sizes, numbers of classes, label types, and  context types/levels used.}

\begin{table*}[!t]
    \centering
    \caption{Summary of datasets used by context-based approaches in both image-based and video-based tasks.}
    \resizebox{\linewidth}{!}{
    \begin{tabular}{|c|c|c|c|c|c|}
    \hline
    \textbf{Datasets} & \textbf{Data Types} & \textbf{Sizes} & \textbf{\# Classes} & \textbf{Label Types} & \textbf{Context Types/Levels}  \\ \hline
    Caltech Camera Traps \citep{beery2018recognition}    & Image               &   196K       &    21  & Object class label  &   Prior knowledge, Spatial, Temporal   \\
    CelebFaces Attributes \citep{liu2015faceattributes}           & Image               &  200K   &   40   &  Attribute class label   &  Spatial  \\
    Chars74K \citep{deCampos09}           & Image               &    74K      &    64     &  Digit class label   & Local \\
    Cityscapes \citep{Cordts2016Cityscapes}          & Image    &    25K      &   30  &     Semantic, instance-wise, pixel level label   & Spatial, Temporal \\ 
    COCO \citep{lin2014microsoft}            & Image               &    328k      &    80   &   Semantic, object class label  &  Spatial, Other \\
    CUHKSYSU \citep{xiao2017joint}               & Image               &     18K     &  /    &    Pedestrian label    &  Spatial \\
    Curb Ramp \citep{hara2014tohme}           & Image               &    1086      &    1   &  Object class label   &  Spatial  \\
    ImageNet \citep{deng2009imagenet}          & Image               &   1.3M       &    1000    &  Image class label   &  Global, Local \\
    iNaturalist \citep{van2018inaturalist}           & Image               &   670K       & 5089  &  Image class label   &  Prior knowledge, Global, Local   \\   
    Labelme \citep{russell2008labelme}           & Image               &     2688     &   8     &   Object class label  &  Global, Local \\
    MSRC-21 \citep{shotton2009textonboost}            & Image               &    591      & 21   &   Semantic label  & Spatial, Other  \\    
    PascalVOC \citep{Everingham10}           & Image               &    4369      &      20        & Pixel level label  & Spatial, Other \\
    SUN 09 \citep{choi2011tree}           & Image               &   12K       &     200+    &   Object class label  & Spatial, Other \\
    SVT \citep{wang2011end}           & Image               &     350     &    /     &   Object class label  & Spatial \\
    Visual Genome \citep{krishna2017visual}            & Image               &    101K      &    /    & Object class, object relation, question-answering pairs   &  Prior knowledge, Spatial  \\
    VQA \citep{antol2015vqa}           & Image               &    265K      &   /     &   Question-answering pairs  &  Prior knowledge, Spatial \\
    Wider Faces \citep{yang2016wider}           & Image               &    32K      &    61   &   Object class label    &  Prior knowledge, Spatial\\ \hline
    AVA \citep{gu2018ava}                & Video               &   430 15-min videos       &   80   & Action label   &   Spatial, Temporal    \\
    City Cam \citep{zhang2017understanding}               & Video               &    60M frames      & 19     &  Semantic, object class label     &  Spatial, Temporal \\
    KAIST \citep{hwang2015multispectral}               & Video               &     95K frames     &   3     &   Object class label   & Spatial, Temporal \\
    Snapshot Serengeti \citep{swanson2015snapshot}           & Video               &     2.65M frames     &  61    &    Image label   &  Prior knowledge, Spatial, Temporal \\
    UCF101 \citep{soomro2012ucf101}           & Video               &   13K videos      &    101     &   Human action label  & Spatial, Temporal \\
    UT-Interaction \citep{UT-Interaction-Data}           & Video               &    20 1-min videos      &     6     &   Human action label  & Spatial, Temporal \\
    VIRAT \citep{oh2011large}           & Video               &    8 hours videos      &    11   &   Interaction event label   &  Spatial, Temporal \\ \hline
    \end{tabular}
    }
    \label{tab:datasets}
\end{table*}

\subsection{Image Datasets}
    \textbf{Caltech Camera Traps \citep{beery2018recognition}}: Caltech Camera Traps (CCT) is a dataset collected from 140 camera locations in the Southwestern United States, with labels for 21 animal categories, primarily at the species level (most common labels are opossum, raccoon, and coyote). There are approximately 66K bounding annotations. 
    
    \textbf{CelebFaces Attributes \citep{liu2015faceattributes}}: CelebFaces Attributes (CelebA) is a large-scale face attributes dataset with more than 200K celebrity images, each with 40 attribute annotations. The images in this dataset cover large pose variations and background clutter. CelebA has large diversities, large quantities, and rich annotations, including 10,177 number of identities, 202,599 number of face images, and 5 landmark locations, 40 binary attributes annotations per image.
    
    \textbf{Chars74K \citep{deCampos09}}: Chars74K is a benchmark dataset for character recognition in natural images. The dataset consists of 64 classes (0-9, A-Z, a-z), 7705 characters obtained from natural images, 3410 hand drawn characters using a tablet PC, 62992 synthesised characters from computer fonts, which gives a total of over 74K images.
    
    \textbf{Cityscapes \citep{Cordts2016Cityscapes}}: Cityscapes is a large-scale database for semantic understanding of urban street scenes. The dataset provides semantic, instance-wise and dense pixel annotations for 30 classes in 8 categories. The Cityscapes dataset consists of 5000 fined annotated images and 20000 coarse annotated ones. All the data was captured in 50 cities with daytime and good weather conditions. It also provides bounding box annotations of pedestrians and image augmentations with fog and rain conditions.
    
    \textbf{COCO \citep{lin2014microsoft}}: Common Objects in Context, known as COCO, is a large-scale dataset for object detection, segmentation and captioning. The dataset consists of 328K images, where over 220K images are labeled. It has 1.5M object instances, 80 object categories (person, car ,chair, etc.) and 91 stuff categories (sky, street, grass, etc.). There are also 5 captions per image.
    
    \textbf{CUHKSYSU \citep{xiao2017joint}}: CUHKSYSU is a large-scale benchmark dataset for person search. It covers hundreds of scenes from street and movie snapshots. The dataset contains over 18K frames and 8432 identities. The dataset also contains two subsets: low-resolution subset and occlusion subset, for evaluating the influence of various factors on person search.
    
    \textbf{Curb Ramp \citep{hara2014tohme}}: Curb ramp dataset is a small dataset for curb ramp detection. It contains  1086 Google Street View panoramas which come from four cities in North America: Washington DC, Baltimore, Los Angeles and Saskatoon (Canada). Each panorama image has 1024x2048 pixels. It provides bounding box labels for existing curb ramps. On average there are four curb ramps per image. The dataset also contains bounding box labels for missing curb ramps regions.
    
    \textbf{ImageNet \citep{deng2009imagenet}}: ImageNet is an image dataset designed for use in visual object recognition research. It has more than 14M annotated images and more than 1M images have the object level bounding box annotation. Each image has been assigned with one class label.
    
    \textbf{iNaturalist \citep{van2018inaturalist}}: The iNaturalist dataset contains more than 670K images from 5089 natural fine-grained categories. Those categories belong to 13 super-categories including Plant, Insect, Bird, Mammal etc. It is a highly imbalanced dataset from largest super category Plant (196K images from 2101 categories) to smallest super category Protozoa (Only 381 images from 4 categories).
    
    \textbf{Labelme \citep{russell2008labelme}}: Labelme is a small dataset for image classification, which consists of 2688 images from 8 classes. 1000 of them are labeled by annotators from Amazon Mechanical Turk (AMT) and remaining are used for validation and testing.
    
    \textbf{MSRC-21 \citep{shotton2009textonboost}}: MSRC-21 is an image dataset for object segmentation. It contains 591 images of 21 object classes. All images are approximately 320 × 240 pixels.
    
    \textbf{Pascal VOC \citep{Everingham10}}: PASCAL Visual Object Classes (VOC) dataset contains 20 object categories in total. Each image in this dataset has pixel-level segmentation annotations, bounding box annotations, and object class annotations. This dataset has been widely used as a benchmark for object detection, semantic segmentation, and classification tasks. It has 1464 images for training, 1449 images for validation and a private testing set.
    
    \textbf{SUN 09 \citep{choi2011tree}}: SUN 09 is a dataset for context based recognition. The dataset contains 12.000 annotated images covering a large number of scene categories (indoor and outdoors) with more than 200 object categories and 152.000 annotated object instances. Each image contains an average of 7 different annotated objects and the average occupancy of each object is 5 percent of image size.
    
    \textbf{SVT \citep{wang2011end}}: Street View Text dataset, known as SVT, is a street imagery dataset for text detection. The SVT data set consists of images collected from Google Street View, where each image is annotated with bounding boxes around words from businesses around where the image was taken. The dataset contains 350 total images (from 20 different cities) and 725 total labeled words. The dataset also has three subset: SVT-SPOT (word locating), SVT-WORD (word recognition) and SVT-CHAR (character recognition).
    
    \textbf{Visual Genome \citep{krishna2017visual}}: Visual Genome is a dataset, a knowledge base, an ongoing effort to connect structured image concepts to language. It is the largest dataset with descriptions of images, objects, attributes and relationships. There are 35 objects marked on each image on average. The dataset contains 5.4M object descriptions, 1.7M question-answer pairs, 2.8M attributes and 2.3M relationships.
    
    \textbf{VQA \citep{antol2015vqa}}: Visual Question and Answering, known as VQA, is a dataset containing open-ended questions about images. These questions require an understanding of vision, language and commonsense knowledge to answer. The VQA dataset consists more than 265K images and at least 3 questions (5.4 questions on average) per image. There are 10 ground truth answers per questions. VQA also provides automatic evaluation metric for use.
    
    \textbf{Wider Faces \citep{yang2016wider}}: WIDER FACE dataset is a face detection benchmark dataset, of which images are selected from the publicly available WIDER dataset \citep{xiong2015recognize}. The Wider Faces dataset contains 32K images and more than 393K labeled faces with a high degree of  variability in scale, pose and occlusion as depicted in the sample images. WIDER FACE dataset is organized based on 61 event classes.

\subsection{Video Datasets}
    \textbf{AVA \citep{gu2018ava}}: The AVA dataset is a video dataset of atomic visual action. It annotates 80 atomic visual actions in 430 15-minute movie clips, where actions are localized in space and time, resulting in 1.62M action labels in total, with multiple labels per human occurring frequently.
    
    \textbf{City Cam \citep{zhang2017understanding}}: City Cam is a public dataset for large-scale city camera videos, which have low resolution (352x240), low frame rate (1 frame per second), and high occlusion. Bounding box are available for 60K for vehicle. It covers multiple cameras and different weather conditions.
    
    \textbf{KAIST \citep{hwang2015multispectral}}: KAIST is a multispectral pedestrian detection benchmark. The KAIST dataset consists of 95K color-thermal pairs (640x480, 20Hz) taken from a vehicle. All pairs are manually annotated (person, people, cyclist) for 103K annotations and 1182 unique pedestrians. The annotations also include temporal correspondence between bounding boxes.
    
    \textbf{Snapshot Serengeti \citep{swanson2015snapshot}}: Snapshot Serengeti dataset is a dataset for animal classification. The dataset contains approximately 2.65M sequences of camera trap images. There are 61 categories labeled, which are primarily at the species level. The common labels are wildebeest, zebra and Thomson's gazelle.
    
    \textbf{UCF101 \citep{soomro2012ucf101}}: UCF is the largest dataset for action recognition. The UCF101 dataset, which is collected from YouTube, contains 101 action categories, with over 13K videos. The 101 categories are devided into give types: Human-object Interaction, Body-Motion Only, Human-Human Interaction, Playing Musical Instruments and Sports. ALL the videos have a fixed frame rate of 25 FPS and resolution of 320x240.
    
    \textbf{UT-Interaction \citep{UT-Interaction-Data}} UT-Interaction is a video dataset for human interaction recognition. The UT-Interaction dataset contains videos of continuous executions of 6 classes of human-human interactions: shake-hands, point, hug, push, kick and punch. There are 20 video sequences with around 1 minute length for each. Each video contains at least one execution per interaction, with average 8 execution of human activities per video. The video has a frame rate of 30 fps and resolution of 720x480.
    
    \textbf{VIRAT \citep{oh2011large}}: VIRAT dataset is a video dataset for video event recognition. It contains over 8 hours of videos from surveillance cameras from school parking lots, as well as shop entrance, outdoor dining area and construction sites. There are six person-vehicle interaction events types (Loading a vehicle, Unloading a vehicle, Opening a Trunk, Closing a trunk, Getting into a vehicle and Getting out of a vehicle) and five other interaction event types (Gesturing, Carrying a object, Running, Entering a facility and Exiting a facility) defined in the dataset.

\section{Context Integration} \label{contextintegration}
In this section, we review how context information has been integrated in various computer vision tasks in two main categories: image-based tasks and video-based tasks. We focus on the mechanisms in integrating various context information in some representative tasks. We further compare the reviewed context integration in the following aspects: tasks, backbone models, employed context types, employed context levels. Performance comparison is also provided for some of the tasks with and without using context.

Spatial context are heavily used in image-based tasks. Either spatial relation between different objects or different parts within an object are integrated to extract context features. Spatial semantic context, such as location, is served as prior level knowledge. Other semantic context such as object relation description, object appearance, label co-occurrence is served as object presence constraints for tasks like image recognition and object detection. A few works use temporal context from long term (months to years) as a historical information for predicting current object appearance.

Temporal context is the main context sources for video-based tasks. Temporal context not only provides previous clues for current scene, it also carries semantic context and spatial context in both the language form and the visual form. These context can help to solve some challenges in video-based tasks, such as heavy occluded pedestrian detection, video event recognition, temporal query grounding, etc. We review details for representative context integration tasks in these video-based tasks.  

\subsection{Image-based Integration} \label{subsecimagebased}

\begin{table*}[ht]
    \centering
    \caption{Image-based context integration.}
    \resizebox{\linewidth}{!}{
    \renewcommand{\arraystretch}{1}
    \begin{tabular}{|c|c|c|c|c|c|c|}
    \hline
    \textbf{Methods} & \textbf{Tasks} & \textbf{DNN Models} & \textbf{Context Types} & \textbf{Context Levels} & \textbf{Mechanisms}  \\ \hline
    Faceness-Net\citep{yang2015facial}   & Face detection   &  AlexNet &  Spatial   &   Local      & Faceness-Net \\ \hline
    Hierarchical Context Net\citep{li2016human}      & Human attribute recognition    & VGGNet   & Spatial  &  Global, Local  & Hierarchical context net    \\ \hline
    Hierarchical Random Field\citep{yao2010modeling}  & Human-object interaction   &  Custom  & Spatial  &  Local & Graph model     \\ \hline
    ML-GCN\citep{chen2019multi} & Image recognition & GCN & Spatial, Temporal  & Global, Local, Prior knowledge & GCN      \\ \hline
    Spatio-temporal Prior Model\citep{mac2019presence} & Image recognition & ResNet & Spatial, Temporal  & Global, Local, Prior knowledge & Bayesian model    \\ \hline
    CATNet\citep{zhang2020putting} & Image recognition & VGGNet & Spatial, Temporal & Global, Local & Two-stream context net      \\ \hline
    Context Encoder\citep{pathak2016context}  &  Image inpainting    &  AlexNet  & Other  &  Local  & Context encoder    \\ \hline
    Co-occurrence Tree Model\citep{choi2012context}           &   Object detection     &               /         &  Spatial  &  Global  & Tree-structured model   \\ \hline
    Context Data Augmentation\citep{dvornik2018modeling}           &   Object detection                   &    ResNet       &  Spatial  &  Local  & Context CNN   \\ \hline
    Knowledge-aware Model\citep{fang2017object}           &   Object detection                     &   VGGNet      &  Other  &    Global, Local, Prior knowledge & Knowledge graphs   \\ \hline
    Internal-External Context Model\citep{leng2021realize}           &   Object detection                    &    ResNet      &  Spatial  &  Local & Internal-External Network    \\ \hline
    Feature Fusion Attention Model\citep{lim2021small}           &   Object detection                 &     ResNet    &  Other  &   Global, Local & Feature fusion SSD  \\ \hline
    Deformable Part-based Model\citep{mottaghi2014role}           &   Object detection                &      /        &  Spatial  &  Global, Local & Markov random field    \\ \hline
    Siamese Context Network\citep{sun2017seeing}           &   Object detection                       &   Custom    &  Spatial  &   Global &  Siamese CNN     \\ \hline
    Bayes Probabilistic Model\citep{torralba2010using}           &   Object detection                  &   /         &  Spatial  &  Global, Local & Bayesian Model     \\ \hline
    Semantic Relation Reasoning Model\citep{zhu2021semantic}           &   Object detection            &      ResNet            &  Other  &   Global & SSD     \\ \hline
    Cascaded Refinement Network\citep{johnson2018image}            &   Scene graph generation           &       GCN           & Spatial  &    Global, Local & GCN   \\ \hline
    Iterative Message Passing\citep{xu2017scene}            &   Scene graph generation                 &     VGGNet       & Spatial  &  Global, Local & Conditional random fields     \\ \hline
    Graph R-CNN\citep{yang2018graph}            &   Scene graph generation   &   GCN   & Spatial  &   Global, Local  & GCN    \\ \hline
    MOTIFNET\citep{zellers2018neural}            &   Scene graph generation                       &   ResNet   & Spatial  &   Global  &  Bayesian model  \\ \hline
    Conditional Random Field (CRF)\citep{mottaghi2013analyzing}            &   Semantic segmentation                &    /      & Other &  Global & Conditional random field     \\ \hline
    Context-based SVM\citep{du2012context}            &   Text detection                        &  Custom &  Spatial &    Local & SVM    \\ \hline
    Visual-language Re-ranker\citep{sabir2018enhancing}            &   Text detection                 &    ResNet/GoogLeNet     &  Other  &  Global &  Language model    \\ \hline
    PLEX\citep{wang2011end}            &   Text detection                 &   /      &  Spatial  &   Local  & Trie structure   \\ \hline
    Scene Context-based Model\citep{zhu2016could}            &   Text detection                   &  Custom       &  Other  &  Global, Local & CNN/SVM      \\ \hline
    Context-dependent Diffusion Network\citep{cui2018context}           &   Visual relationship detection               & VGGNet & Spatial  &  Global & Graph model     \\ \hline
    Dynamic Tree Structure\citep{tang2019learning}           &   Visual Q\&A                      &   VGGNet   & Spatial  &   Global, Local  & Tree-structured model    \\ \hline
    \end{tabular}
    }
    \label{tab:imagebasedintegration}
\end{table*}

Spatial context and semantic context are heavily used in image-based context integration, even though some works \citep{mac2019presence, zhang2020putting} also use temporal context as a prior to improve the performance. In this section, we review some representative works in different image-based tasks in details, and provide a summary of all the reviewed approaches and the performance comparison of some of them. Table \ref{tab:imagebasedintegration} provides a summary of all the reviewed works in image-based context integration, in terms of tasks, backbone deep NN (DNN) models, employed context types, employed context levels, and mechanisms for using context.

\subsubsection{Face Detection}

Yang et al. \citep{yang2015facial} proposed Faceness-Net for face detection. Instead of using the whole face, the Faceness-Net considered the use of spatial structure and arrangement of face parts as a context cue to detect faces. Each facial parts was scored separately in case of the occlusion and pose variation.

\begin{figure}[ht]
  \centering
  \includegraphics[width=1\linewidth]{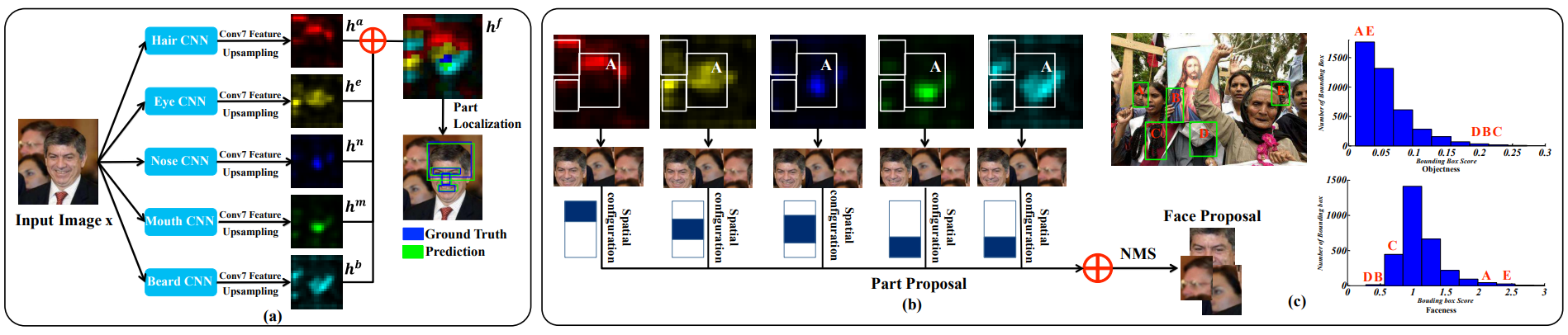}
  \caption{ The pipeline of Faceness-Net \citep{yang2015facial}. Faceness-Net uses spatial structure and arrangements of face parts as context cues to detect faces. The figure was presented in \citep{yang2015facial}.}
  \label{fig:facenessnet}
\end{figure}

Faceness-Net’s pipeline consists of three stages, i.e. generating contextual partness  maps, ranking candidate windows by faceness scores, and refining face proposals for face detection. In first stage as shown in Fig. \ref{fig:facenessnet}(a), a full image is used as input for 5 CNNs: Hair, Eye, Nose, Mouth, Beard. Each CNN outputs a contextual partness map to indicate the spatial location of a specific facial component presented in the image. In second stage, given a set of candidate bounding boxes, the net ranks these bounding boxes according to the contextual partness map with respect to spatial relation of different facial parts. For instance, the hair should appear above the eyes, and the mouth should only appear below the nose, etc. In the last stage, the proposed candidate bounding boxes are refined by training a multitask CNN, where face classification and bounding box regression are jointly optimized. The Faceness-Net achieves a high recall rate of 90.99\% on the challenging FDDB benchmark, outperforming the state-of-the-art method \citep{mathias2014face} by a large margin of 2.91\%. \textcolor{black}{Faceness-Net employs spatial relation between facial components to optimize face detection.  However, tiny-sized faces are still challenging for Faceness-Net, since the facial parts such as eyes, nose or mouth can be barely distinguished from the tiny faces. In this case, the efficiency of the network is comparable with previous CNN face detector \citep{li2015convolutional}. Faceness-Net also uses more data and increases the computational cost because of the multiple CNN architecture.}

Built on the observation that context can unveil more clues to make recognition easier, Li et al. \citep{li2016human} proposed hierarchical context model for human attribute recognition task. Similar to Faceness-Net \citep{yang2015facial}, the hierarchical context model incorporated with both global level context (whole scene) and local level context (human body parts) for final human attribute recognition, which we previously discussed in section \ref{contextlevel}. 

\subsubsection{Image Recognition}
Image recognition is a fundamental and practical task in computer vision, where the aim is to predict the object present in an image. Several recent works \citep{chen2019multi, zhang2020putting, mac2019presence} had employed context information as important cues for recognizing the object. Aodha et al. \citep{mac2019presence} stated that appearance information alone was often not sufficient to accurately differentiate between fine-grained visual categories. They further proposed not only the spatial context, but also temporal context as prior for fine-grained image classification task, which is to classify the object species in the image. As shown in Fig. \ref{fig:spatialexample}, the authors made use of additional spatio-temporal context information in the form of where and when the image was taken. The model also naturally captures the relationships between locations and objects, objects and objects, photographers and objects, and photographers and locations in an interpretable manner. Their method shows a large improvement when combining the prior with image classifiers.

\begin{figure}[ht]
  \centering
  \includegraphics[width=1\linewidth]{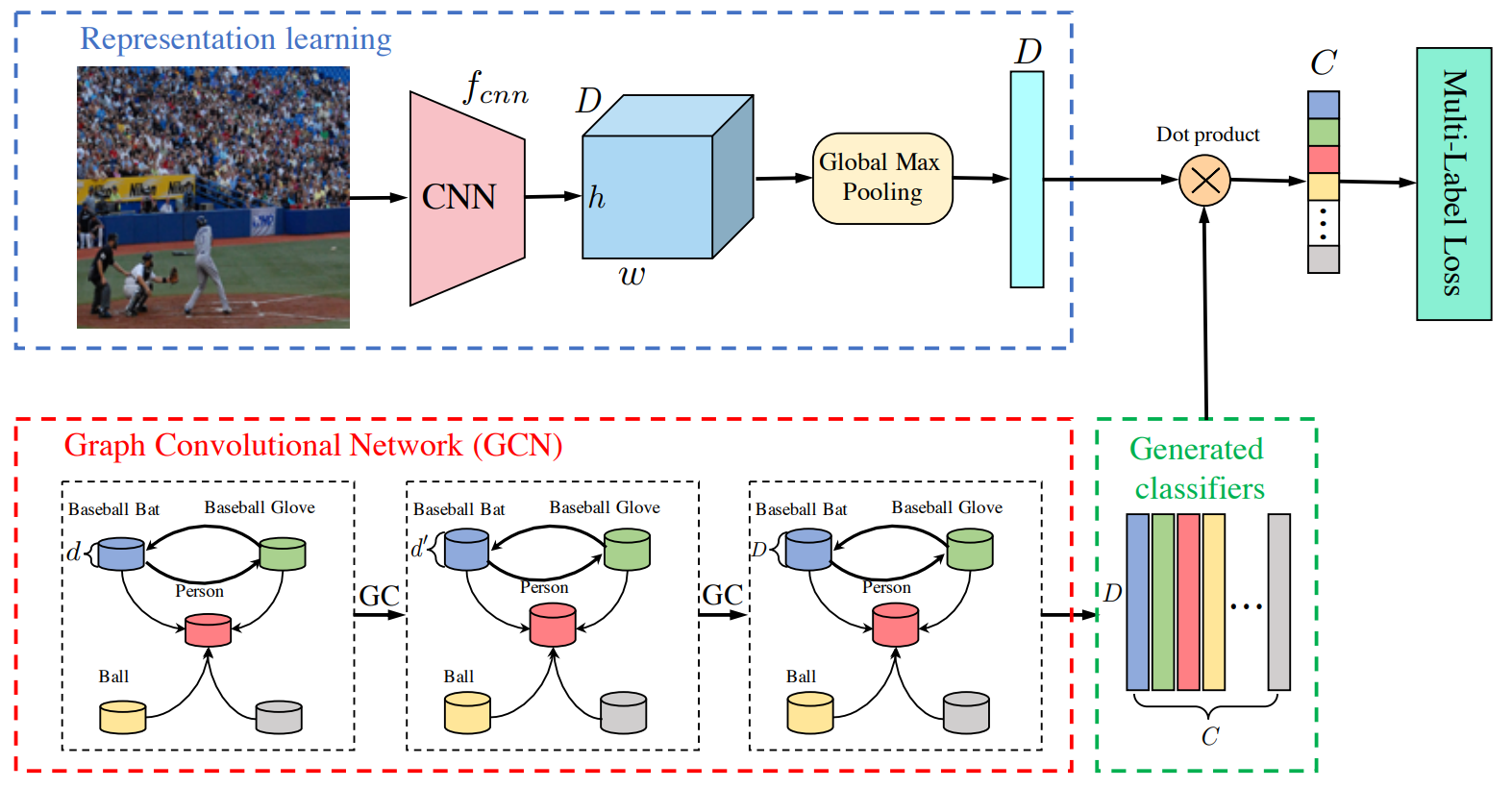}
  \caption{ The architecture of Multi-label image classification \citep{chen2019multi}. Graph Convolutional Network is used to model the co-occurrence relation from prior label dependencies. The figure was presented in \citep{chen2019multi}.}
  \label{fig:multilabel_architecture}
\end{figure}

Multi-label image classification is to predict a set of objects present in an image. The multi-label task is more challenging. As the objects normally co-occur in the physical world,  Chen et al. \citep{chen2019multi} used Graph Convolutional Network to model the co-occurrence relation from prior label dependencies. Graph Convolutional Network uses relation descriptor $A$ to propagate information between nodes. The authors modeled the label correlation dependency in the form of conditional probability, and then fed into Graph Convolutional Network. \textcolor{black}{By applying the co-occurrence relation using label appearance, the model consistently achieves superior performance over previous competing approaches.The model takes advantages of spatial semantic context to achieve better performance. However, the label co-occurrence may describe the object relation preciously when the dataset is large enough and the objects are highly correlated. If the dataset is small, it may not work well since the co-occurrence cannot provide accurate object relations.} The architecture is shown in Fig. \ref{fig:multilabel_architecture}.

A recent work \citep{zhang2020putting} investigated ten critical properties of where, when, and how context modulates recognition, including the amount of context, context and object resolution, geometrical structure of context, context congruence, and temporal dynamics of contextual modulation, to model the role of contextual information in image classification. The authors further proposed a two-stream architecture to dynamically incorporate object and contextual information, and sequentially reason about the class label for the target object.

\subsubsection{Image Inpainting}

\begin{figure}[ht]
  \centering
  \includegraphics[width=1\linewidth]{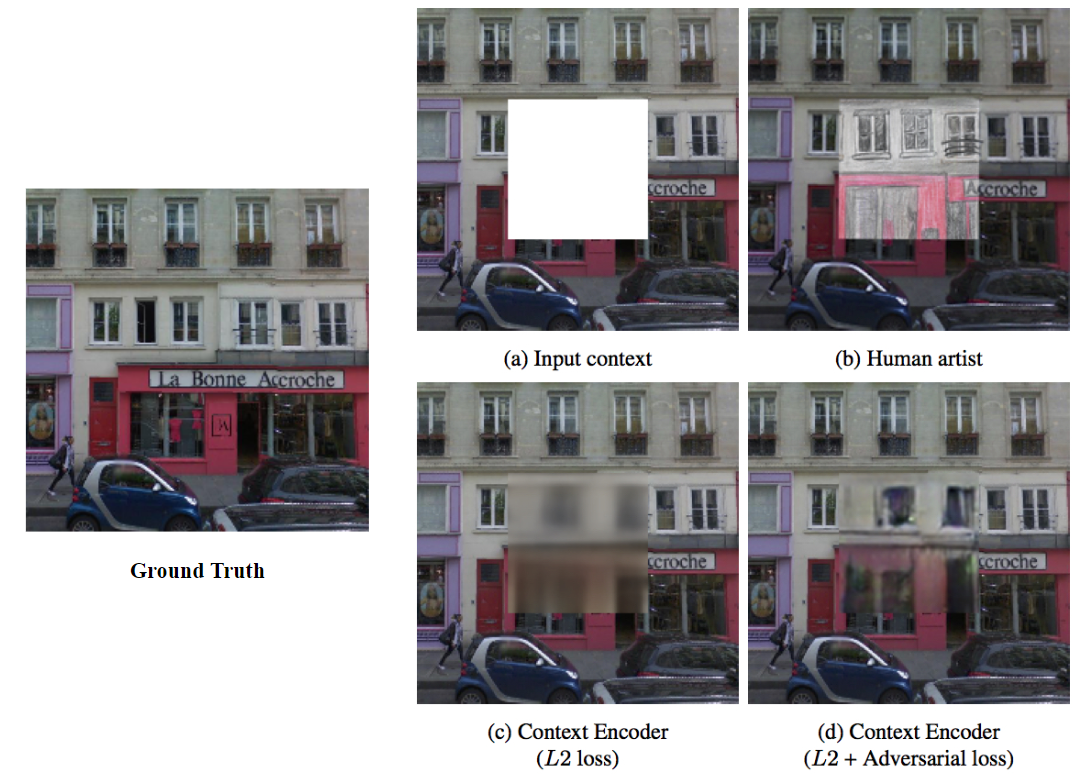}
  \caption{Qualitative illustration of the image inpainting task. Given an image with a missing region (a), a human artist has no trouble inpainting it (b). Automatic inpainting using a context encoder is shown in (c) and (d). The mechanism of employing context is to train a Context Encoder to generate the contents of an arbitrary image region conditioned on its surrounding local context.  The figure was presented in \citep{pathak2016context}.}
  \label{fig:image_inpainting}
\end{figure}

Image inpainting is a task of predicting the arbitrary missing region based on the rest of the image. To correctly predict a missing region, the network is required to learn the common knowledge including the colors and the structures of common objects. %By analogy with auto-encoders, 
Pathak et al. \citep{pathak2016context} trained a convolutional neural network to generate the contents of an arbitrary image region conditioned on its surrounding context. The context encoder learns a representation that captures not only appearance but also the semantics of visual structures. The overall pipeline is a encoder-decoder architecture. The encoder is a convolutional neural network that predict missing parts of a scene from their surroundings. The decoder then generates pixels of missing regions of the image using the features learned from encoder. \textcolor{black}{In order to accomplish the task, both encoder and decoder are required to learn the surrounding local context of the missing parts. The model's efficiency is bit lower due to the large number of parameters of the encoder-decoder architecture.} 

\subsubsection{Object Detection}

The context of an image encapsulates rich information about how natural scenes and objects are related to each other. Such contextual information has the potential to enable a coherent understanding of natural scenes and images. However, context models have been evaluated mostly based on the improvement of object recognition performance even though it is only one of many ways to exploit contextual information. Choi et al. \citep{choi2012context} presented a new scene understanding problem, which is interested in finding scenes and objects that are “out-of-context”. Detecting “out-of-context” objects and scenes is challenging because context violations can be detected only if the relationships between objects are carefully and precisely modeled.

\begin{figure}[ht]
  \centering
  \includegraphics[width=1\linewidth]{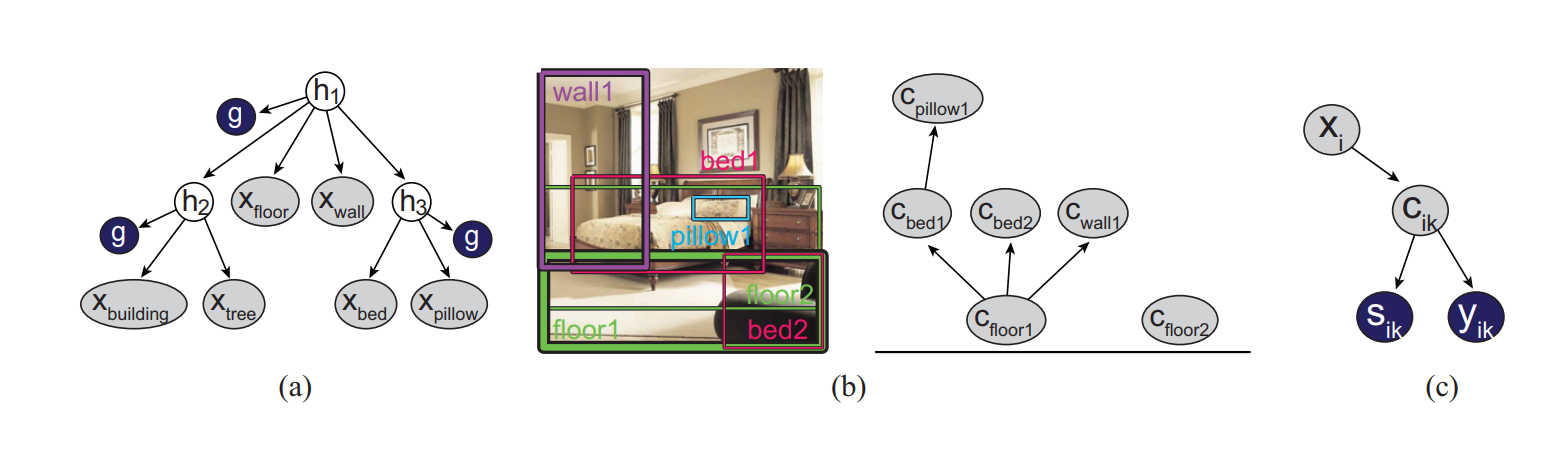}
  \caption{An illustrative example of a support context model for 6 object categories. The mechanism of employing context in this work is to use a graph model to combine different context information such as global context, object co-occurrence and spatial relations between objects.  The figure was presented in \citep{choi2012context}.}
  \label{fig:choi2012context}
\end{figure}

As shown in Fig. \ref{fig:choi2012context}, the author presented a graph model to combine different context information such as global context, object co-occurrence and spatial relations between objects. Their context mechanism computes the probability of each object’s presence and the likelihood of each detection being correct. The results on SUN09 \citep{choi2011tree} dataset demonstrates that context information plays very important role in scene understanding, for both object recognition and out-of-context object detection. \textcolor{black}{However, the major limitation of this work is that when the model is used to detect out-of-context objects, it heavily replied on the support context and ignore the detection with high confidence score, consequently the model makes incorrect detection results.}

\begin{figure}[ht]
  \centering
  \includegraphics[width=1\linewidth]{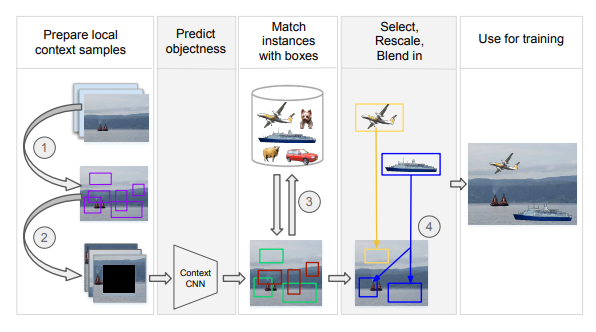}
  \caption{Illustration of the data augmentation approach presented in \citep{dvornik2018modeling}. The work used a context CNN to estimate the likelihood of a particular category of object to be present in certain local context.}
  \label{fig:dvornik2018modeling}
\end{figure}

Another work \citep{dvornik2018modeling} shows that modeling appropriately the visual context surrounding objects is crucial to place them in the right environment. The model estimates the likelihood of a particular category of object to be present inside a box given its neighborhood, and then automatically finds suitable locations on images to place new objects and perform data augmentation. The model (Fig. \ref{fig:dvornik2018modeling}) selects an image for augmentation and 1) generates 200 candidate boxes that cover the image. Then, 2) for each box, it finds a neighborhood that contains the box entirely, crops this neighborhood and mask all pixels falling inside the bounding box; this “neighborhood” with masked pixels is then fed to the context neural network module and, 3) object instances are matched to boxes that have high confidence scores for the presence of an object category. Finally, 4) it selects at most two instances that are rescaled and blended into the selected bounding boxes. The resulting image is then used for training the object detector. The authors further evaluate their context model for data augmentation the subset of the VOC12 dataset. Their experiment demonstrates that context-driven data augmentation has more impact on categories for which visual context is crucial (aeroplane, bird, boat, bus, cat, cow, horse) than some general categories (chair, table, persons, train), since general categories like person, table etc. could appear in various different scenes. \textcolor{black}{However, the limitation of this work is that it uses a CNN network to perform data augmentation, which could increase computational cost potentially and introduce misrepresented objects in the scene.}

%Not only local context, 
Prior knowledge also plays an important role for object detection task. Fang et al. \citep{fang2017object} proposed a novel framework of knowledge-aware object detection, which enables the integration of external knowledge such as knowledge graphs into any object detection algorithm. Background knowledge can often be organized as a knowledge graph, which is a data structure capable of modeling both real-world concepts and their interactions. The framework considers a knowledge graph for modeling semantic consistency, which can better generalize to a pair of concepts even if they are not connected by any edge. The framework employs the notion of semantic consistency to quantify and generalize knowledge, which improves object detection through a reoptimization process to achieve better consistency with prior knowledge. It also provides context-aware approach for object detection task, which not only considers the visual context, but also considers the prior knowledge context.

Mottaghi et al.\citep{mottaghi2014role} studied the role of context in existing state-of-the-art detection and segmentation approaches. They designed a novel category level object detector which exploits both local context around each candidate detection as well as global context at the level of the scene. The model exploits both appearance and semantic segmentation as potentials. It also incorporates global context by scoring context classes present in the full image.

\begin{figure}[ht]
  \centering
  \includegraphics[width=1\linewidth]{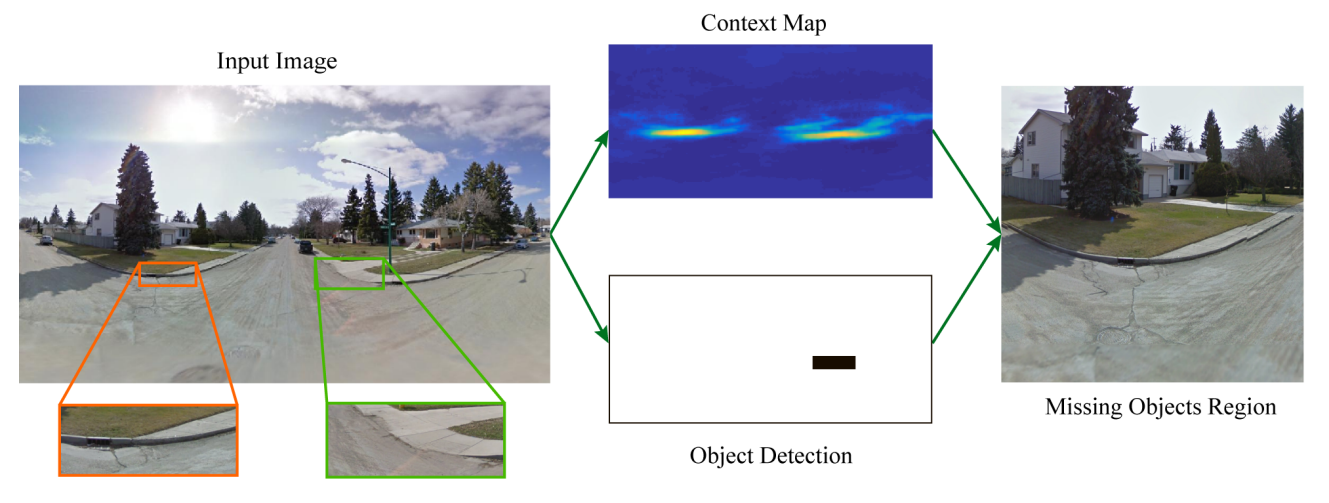}
  \caption{ The approach to determine where objects are missing by learning a context model so that it can be combined with object detection results. The mechanism of employing context in this work is to train a Siamese Context Network to learn the pair-wise existence of curb ramps. The figure was presented in \citep{sun2017seeing}.}
  \label{fig:sun2017seeing_2}
\end{figure}

Context not only can be used to detect the object, but also helps to predict where objects should exist, even when no object instances are present. Sun \citep{sun2017seeing} performed a novel vision task: finding where objects are missing in an image. The author proposes a Siamese trained Fully convolutional Context network (SFC) (Fig. \ref{fig:sun2017seeing_2}). The  network first generates a context heat map using the context network $Q$. This map shows where an object should appear. Then object detection results are generated by using any object detector. The next step is to convert detection boxes into a binary map by assigning 0 to the detected box region, 1 otherwise. This binary map shows where no objects are found. Furthermore, element-wise multiplication is performed between the context heatmap and the binary map. The resulting map shows the regions where an object should occur according to its context but the detector finds nothing. Finally by cropping the high scored regions (above a preset threshold) from the image according to the resulting map, these are the regions where objects are missing. With the local context and co-occurrence of the curbs at street crossings, context map from SFC network and detection results from object detector can be generated in parallel, which provides a more efficient and effective way to combine context information with target objects. \textcolor{black}{This work heavily employs local context and spatial context without any global context, which could indicate the location of the curb ramps otherwise.}

\begin{figure}[ht]
  \centering
  \includegraphics[width=1\linewidth]{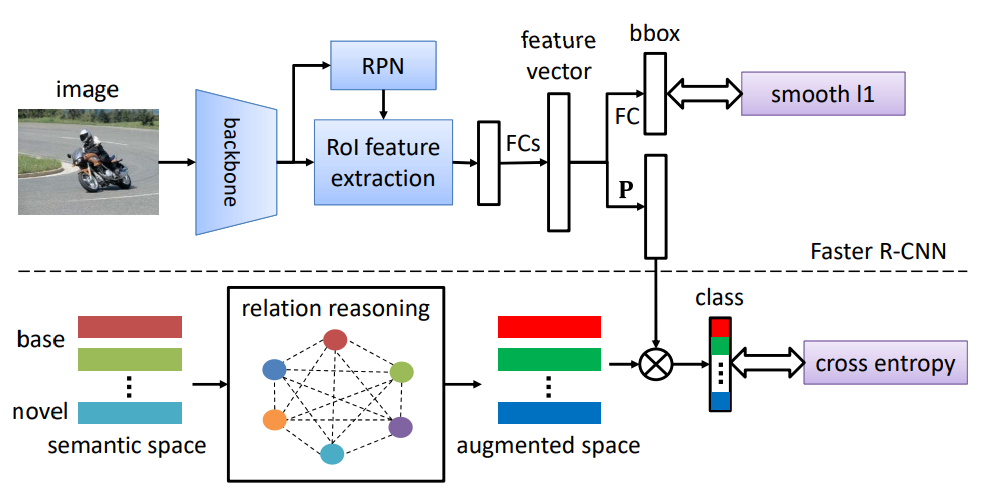}
  \caption{ Overview of the SRR-FSD. The figure was presented in \citep{zhu2021semantic}. The mechanism of employing context in SRR-FSD is to model the semantic relation by building a semantic space, using exclusive semantic context from word embeddings.}
  \label{fig:zhu2021semantic_2}
\end{figure}

A recent work \citep{zhu2021semantic} investigated utilizing the semantic context together with the visual information and introduce explicit relation reasoning into the learning of few-shot object detection. Word embedding is used to represent each class label. Semantic relation consistency is embedded between base class and novel class. If prior knowledge is given that the novel class “bicycle” looks similar to “motorbike”, can have interaction with “person”, and can carry a “bottle”, it would be easier to learn the concept “bicycle” than solely using a few images. Such explicit semantic relation context is even more crucial when visual context is hard to access. This few-shot detector is built on top of Faster R-CNN. A semantic space is built from the word embeddings of all corresponding classes in the dataset and is augmented through a relation reasoning module. The overall framework is shown in Fig. \ref{fig:zhu2021semantic_2}. \textcolor{black}{The work tries to reduce the domain gap between visual information and language information. But when more images are
available, the visual information becomes more precise then the language information starts to be misleading. It is still challenging on how to properly model visual-language relation to reduce the domain gap.}

\subsubsection{Scene Graph Generation}

Today's state-of-the-art deep learning models, such as Faster R-CNN \citep{ren2015faster}, Yolo \citep{bochkovskiy2020yolov4}, etc., have mostly tackled detecting and recognizing individual objects in isolation. However, Even a perfect object detector would struggle to perceive the subtle difference between a man feeding a horse and a man standing by a horse. The rich semantic relationships between these objects have been largely untapped by these models. To understand the visual scenes, one crucial step is to build a structured representation, i.e, a scene graph, which captures objects and their semantic relationships. Scene graph not only offers contextual cues for recognition tasks, but also provides values in a larger variety of high-level visual tasks. Different context has been widely used in scene graph generation task. The goal of scene graph generation is to generate a visually-grounded scene graph from an image. 

\begin{figure}[ht]
  \centering
  \includegraphics[width=1\linewidth]{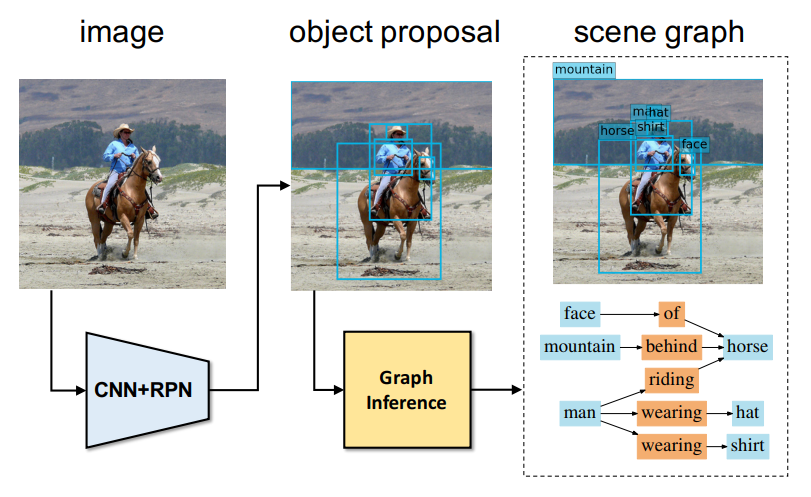}
  \caption{ An overview of model architecture for \citep{xu2017scene}. The figure was presented in \citep{xu2017scene}. The mechanism of employing context is to learn spatial context and semantic context to model the object relationships over a graph inference module.}
  \label{fig:xu2017scene}
\end{figure}

Xu et al. \citep{xu2017scene} proposed a model to generate scene graph by using RNNs and learn to iteratively improve its predictions via message passing. The model can take advantage of spatial context and semantic relations as cues to make better predictions on objects and their relationships. Given an image as input, the model first produces a set of object proposals using a Region Proposal Network (RPN) \citep{ren2015faster}, and then passes the extracted features of the object regions to a novel graph inference module. The output of the model is a scene graph \citep{johnson2015image}, which contains a set of localized objects, the category of each object, and relationship types between each pair of objects. The model also employ both global level context and local level context for localizing the objects. \textcolor{black}{However, as mentioned in their paper, the performance decreases if the number of iterations increases, because of the noisy messages start to permeate through the graph and hamper the prediction. Thus this framework needs a solution to solve the noise message passing issue.}

\begin{figure}[ht]
  \centering
  \includegraphics[width=1\linewidth]{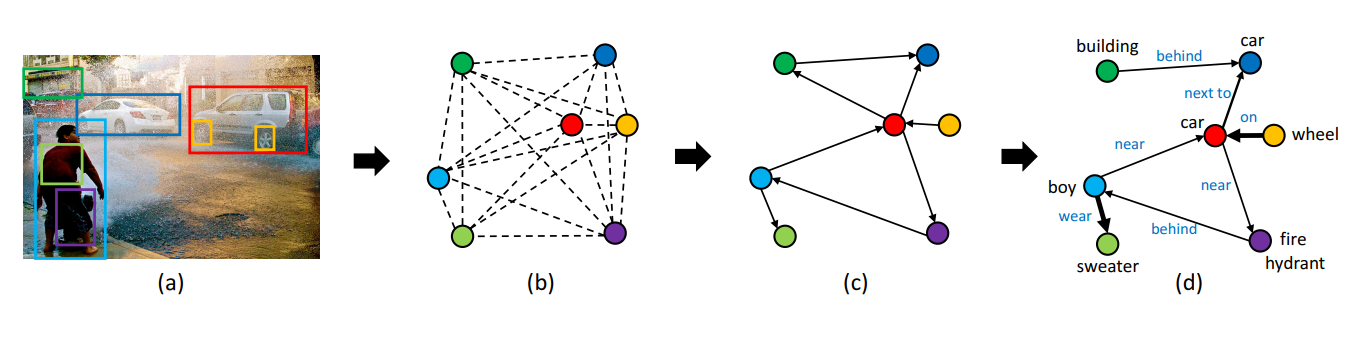}
  \caption{The pipeline of the graph R-CNN framework proposed in \citep{yang2018graph}. A Graph Convolution Network is used to capture contextual information between objects and relations.  }
  \label{fig:yang2018graph}
\end{figure}

Yang, et al. \citep{yang2018graph} used a graph convolution network to capture contextual information between objects and relations. In Fig. \ref{fig:yang2018graph}, it is much more likely for a ‘car’ and a ‘wheel’ to have a relationship than a ‘wheel’ and a ‘building’. Furthermore, the types of relationships that typically occur between objects are also highly dependent on those objects, e.g the wheel is on the car. An attentional graph convolution network (aGCN) is applied to propagate higher-order context throughout the graph, by updating each object and relationship representation based on local neighbor context. In this work, the author employed spatial context types for modeling the relationship between objects. Global and local level context are employed for region proposal and localize the objects.

\begin{figure}[ht]
  \centering
  \includegraphics[width=1\linewidth]{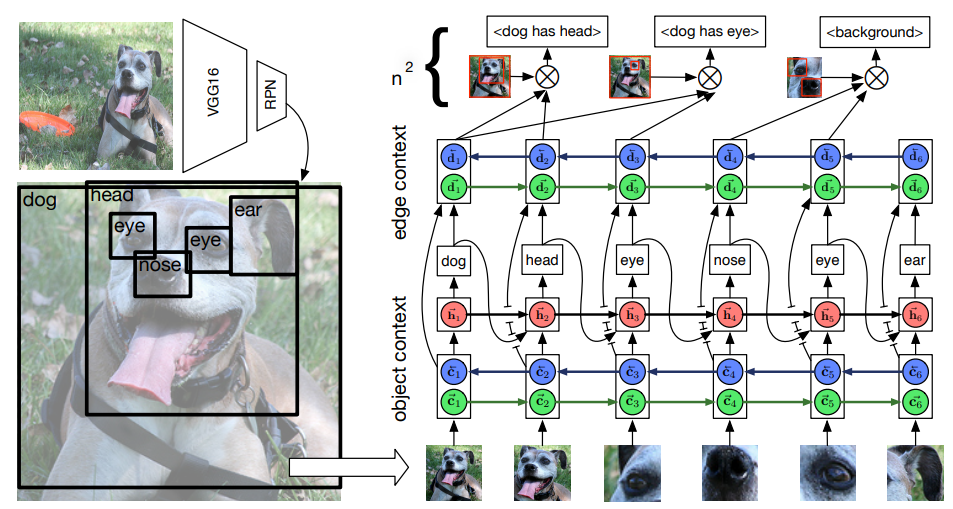}
  \caption{ A diagram of a Stacked Motif Network (MOTIFNET). The mechanism of employing context in this work is to use a MOTIFNET to encode the global context that can directly inform the local context (i.e., objects and relations). The figure was presented in \citep{zellers2018neural}.}
  \label{fig:zellers2018neural}
\end{figure}

%Scene graph, or motifs, is the structural representation of the object relations in a visual scene. 
One of the challenges in modeling scene graphs (or motifs as called by some researchers) lies in devising an efficient mechanism to encode the global context that can directly inform the local predictors (i.e., objects and relations). To overcome this challenge, Zellers et al. \citep{zellers2018neural} introduced the Stacked Motif Network(MOTIFNET). The network builds on Faster R-CNN \citep{ren2015faster} for predicting bounding box regions. Global context across bounding regions is computed and propagated through bidirectional LSTMs, which is then used by another LSTM that labels each bounding region conditioned on the overall context and all previous labels. Between each stage, global context is computed using bidirectional LSTMs and is then used for subsequent stages. In the first stage, a detector proposes bounding regions and then contextual information among bounding regions is computed and propagated (object context). The global context is used to predict labels for bounding boxes. Given bounding boxes and labels, the model constructs a new representation (edge context) that gives global context for edge predictions. Finally, edges are assigned labels by combining contextualized head, tail, and union bounding region information with an outer product. The overall structure is shown in Fig. \ref{fig:zellers2018neural}. \textcolor{black}{As the authors stated in their paper, if the detector fails, it will result in cascading failure to predict any relation edge to the object. Therefore model needs to overcome the detection accuracy for better scene graph generation. }

\subsubsection{Image-based Context Integration Performance}\label{secimagebasedperformance}

We compare the performance of different context integration on the object detection tasks, which is performed on two dataset: PASCALVOC07 \citep{Everingham10} and MSCOCO\citep{lin2014microsoft} datasets. The performance of different context integration is compared against context free baseline model. PASCALVOC, know as PASCAL Visual Object Classes contains 20 object categories. Several object detection model with context integration has evaluated their performance on this dataset. 

\begin{table}[ht]
    \centering
    \caption{The performance of different methods for object detection task in image-based context integration. ($\uparrow$: Higher is better.)}
    \resizebox{\linewidth}{!}{
    \renewcommand{\arraystretch}{2}
    \begin{tabular}{|c|c|c|c|}
    \hline
    \textbf{Datasets} & \textbf{Methods}  & \textbf{mAP$\uparrow$}  & \textbf{AR$\uparrow$}  \\ \hline
    \multirow{4}{*}{PASCALVOC07\citep{Everingham10}}   & Context Free Baseline       & 58.0 &   -  \\ \cline{2-2} \cline{3-4} 
          &      Context Data Augmentation\citep{dvornik2018modeling}                   & 62.0 &  -   \\ \cline{2-2} \cline{3-4} 
                   &    Knowledge-aware Model \citep{fang2017object}                      & 66.6 & 85.0  \\ \cline{2-2} \cline{3-4} 
          &         Feature Fusion Attention Model \citep{lim2021small}                & 78.1 &  -     \\ \hline
    \multirow{2}{*}{MSCOCO\citep{lin2014microsoft}} & Context Free Baseline         & 42.7 &  -    \\ \cline{2-2} \cline{3-4} 
                 &        Internal-External Context Model \citep{leng2021realize}                  & 69.3 & -   \\ \hline
    \end{tabular}
    }
    \label{tab:imagebasedperformance}
\end{table}

The performance of the PASCALVOC is shown in Table \ref{tab:imagebasedperformance}. Overall the performance of all context integration outperform the context free baseline model (+4\% to +20.1\%). \textcolor{black}{ The context integration is from integrating spatial semantic context only \citep{dvornik2018modeling}  to the combination of the context integration (of spatial semantic context, global level context and local level context) \citep{lim2021small}.} All the models are using ConvNets to integrate context information into the context free model. Feature Fusion Attention Model\citep{lim2021small} achieves the best performance on PASCALVOC dataset. We also compared context integration with context free model on MSCOCO \citep{lin2014microsoft} dataset. Internal-External context model\citep{leng2021realize} integrated three components for contextual learning. Feature fusion component is used to capture local context of objects. Context reasoning component is used to improve the region proposals by using semantic relation between easily detected objects and hard ones. Context feature augmentation component is used to learn spatial pair-wise relation between region proposals from context reasoning component. The model then produce global feature information associated with region proposals for the final classification. As shown in Table \ref{tab:imagebasedperformance}, the performance has a significant improvement over the context free model.

From the reviewed context based integration, we can learn that context information can help with object detection. Different context information can also be combined to achieve better performance.

\subsection{Video-based Integration} \label{subsecvideobased}

\begin{table*}[ht]
    \centering
    \caption{Video-based context integration.}
    \resizebox{\linewidth}{!}{
    \renewcommand{\arraystretch}{2}
    \begin{tabular}{|c|c|c|c|c|c|c|}
    \hline
    \textbf{Methods} & \textbf{Tasks} & \textbf{DNN Models} & \textbf{Context Types} & \textbf{Context Levels} & \textbf{Mechanisms} \\ \hline
    Context R-CNN\citep{beery2020context}   & Object detection                   &    ResNet        & Spatial, Temporal &  Global, Local &  Memory bank    \\ \hline
    Tube Feature Aggregation Network\citep{Wu_2020_CVPR}                & Pedestrian detection               &       ResNet         & Spatial, Temporal & Local  & Feature aggregation      \\ \hline
    Contextual Graph Representation Learning\citep{yan2019learning}                & Person search                   &     ResNet/GCN       & Spatial, Temporal & Global, Local &  GCN      \\ \hline
    Contextual Boundary-aware Framework\citep{wang2020temporally}               & Temporal query grounding         & Custom & Spatial, Temporal  & Global, Local  & Self attention      \\ \hline
    Hierarchical Temporal Network\citep{yuan2019semantic}   & Temporal query grounding   & Custom  &  Temporal   &  Global, Local  & Semantic conditioned model    \\ \hline
    Spatio-temporal Progressive Learning\citep{yang2019step}               & Video action detection             & VGGNet & Spatial, Temporal  &   Global, Local  & GCN   \\ \hline
    Spatio-temporal Structural Model\citep{zhu2013context}            & Video event recognition  & / &  Temporal  &  Local & Structural activity model      \\ \hline
    Hierarchical Context Learning\citep{wang2015video, wang2016hierarchical}            & Video event recognition   & Custom  & Spatial, Temporal   &  Prior knowledge, Global, Local  & Hierarchical context model    \\ \hline
    
    \end{tabular}
    }
    \label{tab:videobasedintegration}
\end{table*}

In video-based context integration, spatial context and semantic context are carried in the temporal dimension. Video-based tasks heavily use temporal context with spatial relations between target objects or events to make better prediction. In this section, we review recent representative works that employed context information, and provide an overview of all the reviewed video-based context integration. Table \ref{tab:videobasedintegration} provides a summary of all the reviewed works in video-based context integration, in terms of tasks, backbone deep NN (DNN) models, employed context types, employed context levels, and mechanisms for using context.

\subsubsection{Pedestrian Detection}

\begin{figure}[ht]
  \centering
  \includegraphics[width=1\linewidth]{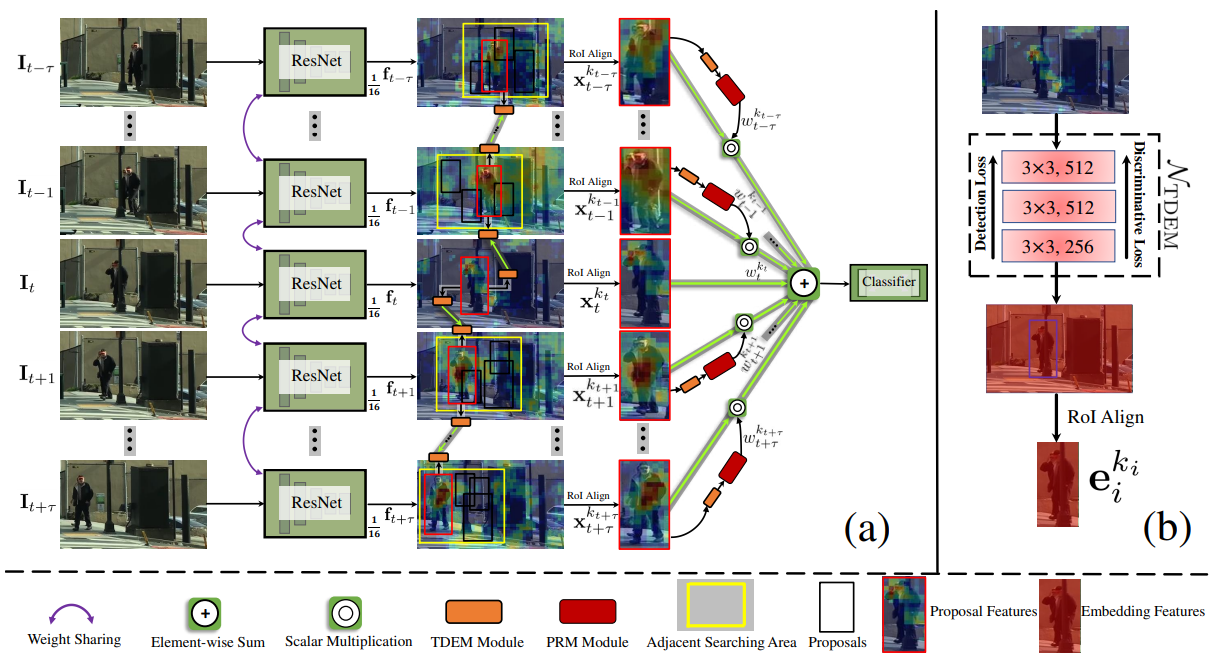}
  \caption{Overall framework of the TFAN. The mechanism of employing context is to iteratively search for its relevant local context along temporal order to form a context tube, by training a Feature Aggregation Network. The figure was presented in \citep{Wu_2020_CVPR}.}
  \label{fig:Wu_2020_CVPR_2}
\end{figure}

Detecting heavily occluded pedestrians is crucial for real-world applications, such as autonomous driving systems. There are two main challenges for this task: (1) Heavily occluded pedestrians are hard to be distinguished from background due to missing/incomplete observations; (2) Detectors seldom have a clue about how to focus on the visible parts of partially occluded pedestrians. Although there are works try to solve the occlusion issue using attention, feature transformation, and part-based detection, they have not leveraged additional context information beyond a single image. A recent work \citep{Wu_2020_CVPR} exploited the local context through temporal context of pedestrians in videos by aggregating local context features to enhance pedestrian detectors against occlusions. The model iteratively searches for its relevant local context along temporal order to form a context tube. Furthermore, the model resorts to local spatial-temporal context to match pedestrians with different extents of occlusions using a new temporally discriminative embedding module and a part-based relation module. Overall, the work employs spatial context, temporal context and global level context in combination to overcome the pedestrian occlusion issue, which also outperforms the context-free methods on benchmark datasets.\textcolor{black}{ However, the framework is sensitive to some of the hyper-parameters. If the parameters becomes larger, the result becomes worse. A better training process could be designed to find the best hyper-parameter combination.}

\subsubsection{Temporal Query Grounding}

\begin{figure}[ht]
  \centering
  \includegraphics[width=1\linewidth]{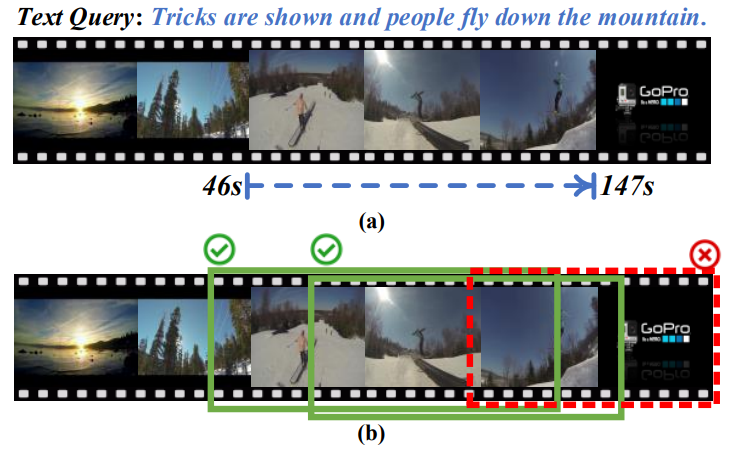}
  \caption{ (a) The task of temporally grounding language queries in videos. (b) Positive and negative training segments defined in anchor-based approaches given the sentence query in (a). The figure was presented in \citep{wang2020temporally}.}
  \label{fig:wang2020temporally}
\end{figure}

The task of temporally grounding language queries in videos is to temporally localize the best matched video segment corresponding to a given language (sentence), as shown in Fig. \ref{fig:wang2020temporally}. This requires both visual understandings and linguistic understandings. Previous works use predefined sliding window to scan the video, which could compromise the precision of semantic boundaries. By using  temporal semantic context and short-term temporal context, it could provide more accurate boundaries. In order to cooperate both temporal semantic context and short-term temporal context in a video, Wang et al. \citep{wang2020temporally} propose an end-to-end contextual boundary-aware model for temporally grounding language queries task, which aggregates temporal semantic context and short-term temporal context, by modeling the relationship between the current frame and its neighbors. The proposed context module operates on the layer which already integrates query and video information. It thus enables the network to “perceive” the surrounding local context and collect reliable contextual evidences before making predictions at the current step. This is different from previous context modeling, which only considers visual context but ignores the impact of semantic context. The temporal context dependency provides both semantic relation between objects and different local visual context compared to the background. With the aid of its local context, the activity is better localized. Temporal context and temporal semantic context played as crucial cues for better precision in this framework.

\begin{table*}[!t]
    \centering
    \caption{Performance of methods in different tasks in video-based context integration.($\uparrow$: Higher is better. $\downarrow$: Lower is better.)}
    \resizebox{\linewidth}{!}{
    \begin{tabular}{|c|c|c|c|c|c|c|c|c|}
    \hline
    \textbf{Tasks}                                    & \textbf{Methods}                               & \textbf{Datasets}                 & \textbf{mAP$\uparrow$}  & \textbf{AR$\uparrow$}   & \textbf{Top-1$\uparrow$} & \textbf{HO$\downarrow$}   & \textbf{PO$\downarrow$}  & \textbf{R$\downarrow$}    \\ \hline
    \multirow{2}{*}{Object Detection} & Context Free Baseline & \multirow{2}{*}{Caltech Camera Traps\citep{beery2018recognition}} & 56.8 & 53.8 & - & - & - & - \\ \cline{2-2} \cline{4-9} 
    & Context R-CNN \citep{beery2020context}                       &                         & 76.3 & 62.3 & -     & -    & -    & -    \\ \hline 
    \multirow{2}{*}{Occluded Pedestrian Detection}   & Context Free Baseline          & \multirow{2}{*}{KAIST \citep{hwang2015multispectral}}  & -    & -    & -     & 76.6 & 55.4 & 35.9 \\ \cline{2-2} \cline{4-9} 
    & Tube Feature Aggregation Network \citep{Wu_2020_CVPR}    &                         & -    & -    & -     & 71.3 & 49.0 & 34.6 \\ \hline 
    \multirow{2}{*}{Person Search}          & Context Free Baseline                             & \multirow{2}{*}{PRW \citep{zheng2016person}}    & 21.3 & -    & 49.9  & -    & -    & -    \\ \cline{2-2} \cline{4-9} 
    & Contextual Graph Representation \citep{yan2019learning}     &                         & 33.4 & -    & 73.6  & -    & -    & -    \\ \hline 
    \multirow{2}{*}{Video Action Detection} & Context Free Baseline                   & \multirow{2}{*}{UCF101\citep{soomro2012ucf101}} & 65.7 & -    & -     & -    & -    & -    \\ \cline{2-2} \cline{4-9} 
    & Spatio-temporal Progressive Learning \citep{yang2019step} &                         & 75.0 & -    & -     & -    & -    & -    \\ \hline
    \end{tabular}
    }
    \label{tab:videobasedperformance}
\end{table*}

\subsubsection{Video Event Recognition}

\begin{figure}[ht]
  \centering
  \includegraphics[width=1\linewidth]{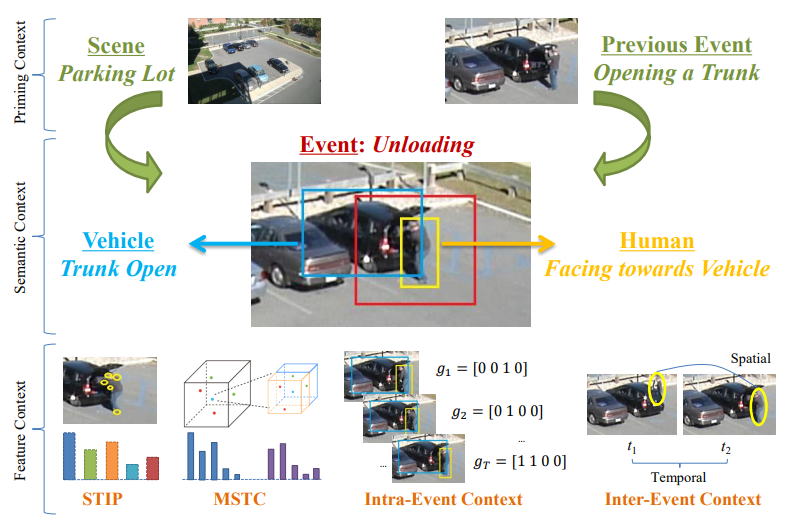}
  \caption{An example of incorporating contexts at different levels. A hierarchical context model is built to incorporate different context (prior knowledge, spatial, temporal, semantic) for a better video event prediction and recognition. The figure was presented in \citep{wang2016hierarchical}.}
  \label{fig:wang2016hierarchical}
\end{figure}

Context also plays a crucial role in video event recognition tasks. Video event recognition aims to recognize the spatio-temporal visual patterns of events from videos. Recognizing events in surveillance videos is still challenging due to problems such as intra-class variation and low image resolution, to name a few. Different context information could help on solving these challenges. Here, context can be regarded as information that is not directly related to event recognition task, but it can be utilized to improve the traditional data-driven and target-centered event recognition. Wang and Ji published a serial work \citep{wang2015video, wang2016hierarchical}, focus on video event recognition task, by integrating multiple levels of contexts.

Wang et al. \citep{wang2015video, wang2016hierarchical} define three levels of context in video event recognition task: prior level, semantic level, and feature level. The prior level contexts capture the prior information of events, which is the prior knowledge such as location, time and weather, etc. These prior knowledge can indicate the possible scene states in the video. Temporal context is also treated as prior knowledge support in the event, which can provides support for the prediction of the current event given previous event. Spatial semantic context can capture the semantic interactions among event entities, such as person get off the car, person open the trunk, etc. Feature level context in this work is defined as local level context (visual appearance) and other semantic context (interaction). Temporal context is used to connect feature level context through the event. Wang et al.\citep{wang2015video, wang2016hierarchical} also introduce a hierarchical context model to learn all these feature for a better video event prediction and recognition.

\subsubsection{Video-based Context Integration Performance}\label{secvideobasedperformance}

For video-based context integration, we compare with context-free model on four different tasks: object detection, pedestrian detection, person search and video action recognition. For these tasks, the evaluation is performed on different video datasets: Caltech Camera Traps\citep{beery2018recognition} for object detection, KAIST\citep{hwang2015multispectral} for occluded pedestrian detection, PRW\citep{zheng2016person} for person search and UCF101\citep{soomro2012ucf101} for video action detection.

The performance result for video-based context integration is shown in Table \ref{tab:videobasedperformance}. The main evaluation metric is mAP (mean average precision). From the result, all methods with context integration have a great improvement over the context-free model for object detection (+19.5\%), person search (+12.1\%) and video action detection (+9.3\%). For occluded pedestrian detection, the evaluation metric is the missing rate of the pedestrian, which is the lower rate, the better performance. Tube feature aggregation network\citep{Wu_2020_CVPR} gain large improvement on heavy occluded (HO) and partial occluded (PO) category, and slight improvement on reasonable category. Context also perform as a key role in video-based tasks.

\subsection{The Merits of Context Integration}
\label{secmeritsofcontextintegration}
\textcolor{black}{
We further provide a study of some key merits for all the reviewed works. Table \ref{tab:contextintegration_merit} provides a summary of the merits of all the reviewed works in both image-based context integration and video-based context integration, in terms of \textit{human likeness}, \textit{performance accuracy} and \textit{efficiency for data and time}. For image-based context integration, we observe that all the works can achieve same or even better performance over previous state-of-art methods. For tasks like face detection, human attribute recognition and human-object interaction, context information is more effective, resulting in much better performance over context-free methods. Faceness-Net produces more human-like behaviors: When human look at an image, we also try to find obvious characteristics like eyes, nose, hair and mouth for finding the correct face in the image. Another work on image inpainting \citep{pathak2016context} is also human-like. When we try to fill a hole in the image, we look into the surrounding areas of the hole, and try to guess the shape and color of the hole. The result of this method is also similar to a human artist (Fig. \ref{fig:image_inpainting}). Most of the works in image-based context integration implement context information into deep learning models and aggregate with the features from context-free methods, in order to achieve better performance. Although the deep learning models are try to mimic our brain neurons, most of the works are not obvious human-like behaved methods. In Table \ref{tab:contextintegration_merit}, the human likeness of each method is categorized as High, Medium, Low.   In terms of accuracy versus efficiency, we can see that most of image-based works can achieve better performance (Medium or High in the Accuracy column in the table), by using the same amount data and computational cost (denoted by $\rightarrow$ in the  Efficiency column in  the table) comparing to the context-free methods. A number of the works need more data and/or time (denoted by $\uparrow$ in the table) to achieve the same performance, even though a few need less data ($\downarrow$). For the latter, utilization of context information with data augmentation and semantic relation reasoning helps the models to achieve high performance with less data.}

\textcolor{black}{
For video-based context integration, all the reviewed works significantly outperform the state-of-art context-free methods (as denoted by High in the Accuracy column in Table \ref{tab:contextintegration_merit} ). Context R-CNN \citep{beery2020context} leverages long term temporal context for improving object detection on challenging data. These kind context information also helps experts to recognize species in challenging scene in real world. Another work \citep{yan2019learning} observes the co-occurrence of a target person and a context person (the person appears in different scenes with the target person), which can provide more confidence score on identifying the target person. We human use similar way to find a target person in real world. In order to aggregate context information with temporal support in video, all of these methods use more data during training and a longer training time (as denoted by $\uparrow$ in the table).}

\begin{table*}[ht]
    \centering
    \caption{The merits of context integration. First part: Image-based context integration. Second part: Video-based context integration. (In the Efficiency column: $\uparrow$: more, $\rightarrow$: equal, $\downarrow$: less, comparing with context-free baseline methods.)}
    \resizebox{\linewidth}{!}{
    \renewcommand{\arraystretch}{1}
    \begin{tabular}{|c|c|c|c|c|}
    \hline
    \textbf{Methods} & \textbf{Tasks}  & \textbf{Human likeness} & \textbf{Accuracy} & \textbf{Efficiency (data/time)}  \\ \hline
    Faceness-Net\citep{yang2015facial}   & Face detection   &  High &  High   &   Data: $\uparrow$, Time: $\uparrow$   \\ \hline
    Hierarchical Context Net\citep{li2016human}      & Human attribute recognition    & Medium   & High  &  Data: $\rightarrow$, Time: $\uparrow$  \\ \hline
    Hierarchical Random Field\citep{yao2010modeling}  & Human-object interaction   &  Low  & High   &  Data: $\rightarrow$, Time: $\rightarrow$ \\ \hline
    ML-GCN\citep{chen2019multi} & Image recognition & Low & Medium & Data:$\uparrow$, Time: $\rightarrow$    \\ \hline
    Spatio-temporal Prior Model\citep{mac2019presence} & Image recognition & Medium & High & Data: $\uparrow$, Time: $\uparrow$ \\ \hline
    CATNet\citep{zhang2020putting} & Image recognition & Medium & High & Data: $\rightarrow$, Time: $\rightarrow$ \\ \hline
    Context Encoder\citep{pathak2016context}  &  Image inpainting    &  High  &  Medium  &  Data: $\rightarrow$, Time: $\uparrow$ \\ \hline
    Co-occurrence Tree Model\citep{choi2012context}           &   Object detection     &               Medium       &  Medium &  Data: $\rightarrow$, Time: $\rightarrow$ \\ \hline
    Context Data Augmentation\citep{dvornik2018modeling}           &   Object detection                   &    Medium       &  High  &  Data: $\downarrow$, Time: $\uparrow$  \\ \hline
    Knowledge-aware Model\citep{fang2017object}           &   Object detection                     &   Low      &  Medium  &    Data: $\uparrow$, Time: $\rightarrow$ \\ \hline
    Internal-External Context Model\citep{leng2021realize}           &   Object detection                    &    Low      &  High  &  Data: $\uparrow$, Time: $\uparrow$ \\ \hline
    Feature Fusion Attention Model\citep{lim2021small}           &   Object detection                 &     Medium    &  Medium  &   Data: $\rightarrow$, Time: $\rightarrow$ \\ \hline
    Deformable Part-based Model\citep{mottaghi2014role}           &   Object detection                &      Medium        &  Medium  &  Data: $\rightarrow$, Time: $\rightarrow$ \\ \hline
    Siamese Context Network\citep{sun2017seeing}           &   Object detection                       &   Low    &  High  &   Data: $\rightarrow$, Time: $\uparrow$ \\ \hline
    Bayes Probabilistic Model\citep{torralba2010using}           &   Object detection                  &   Low        &  Medium  &  Data: $\rightarrow$, Time: $\rightarrow$ \\ \hline
    Semantic Relation Reasoning Model\citep{zhu2021semantic}           &   Object detection            &      Low            &  High  &   Data: $\downarrow$, Time: $\rightarrow$ \\ \hline
    Cascaded Refinement Network\citep{johnson2018image}            &   Scene graph generation           &       Low           & High  &    Data: $\rightarrow$, Time: $\rightarrow$ \\ \hline
    Iterative Message Passing\citep{xu2017scene}            &   Scene graph generation                 &     Low       & High  &  Data: $\rightarrow$, Time: $\rightarrow$ \\ \hline
    Graph R-CNN\citep{yang2018graph}            &   Scene graph generation   &   Medium   & High  &   Data: $\rightarrow$, Time: $\rightarrow$ \\ \hline
    MOTIFNET\citep{zellers2018neural}            &   Scene graph generation                       &   Low   & Medium  &   Data: $\rightarrow$, Time: $\rightarrow$  \\ \hline
    Conditional Random Field (CRF)\citep{mottaghi2013analyzing}            &   Semantic segmentation                &    Low      & Medium &  Data: $\rightarrow$, Time: $\rightarrow$ \\ \hline
    Context-based SVM\citep{du2012context}            &   Text detection                        &  Low &  Medium &    Data: $\rightarrow$, Time: $\rightarrow$ \\ \hline
    Visual-language Re-ranker\citep{sabir2018enhancing}            &   Text detection                 &   Medium   &  Medium  &  Data: $\rightarrow$, Time: $\uparrow$ \\ \hline
    PLEX\citep{wang2011end}            &   Text detection                 &   Low      &  Medium  &   Data: $\rightarrow$, Time: $\rightarrow$  \\ \hline
    Scene Context-based Model\citep{zhu2016could}            &   Text detection                   &  High       &  High  &  Data: $\rightarrow$, Time: $\rightarrow$ \\ \hline
    Context-dependent Diffusion Network\citep{cui2018context}           &   Visual relationship detection               & Low & Medium &  Data: $\rightarrow$, Time: $\rightarrow$ \\ \hline
    Dynamic Tree Structure\citep{tang2019learning}           &   Visual Q\&A                      &   Low   & Medium  &   Data: $\rightarrow$, Time: $\rightarrow$ \\ \hline \hline \hline
     Context R-CNN\citep{beery2020context}   & Object detection     &    High    & High & Data: $\uparrow$, Time: $\uparrow$    \\ \hline
    Tube Feature Aggregation Network\citep{Wu_2020_CVPR}                & Pedestrian detection               &       Low         & High  & Data: $\uparrow$, Time: $\uparrow$      \\ \hline
    Contextual Graph Representation Learning\citep{yan2019learning}                & Person search                   &     High       & High & Data: $\uparrow$, Time: $\uparrow$    \\ \hline
    Contextual Boundary-aware Framework\citep{wang2020temporally}               & Temporal query grounding         & Low & High  & Data: $\uparrow$, Time: $\uparrow$      \\ \hline
    Hierarchical Temporal Network\citep{yuan2019semantic}   & Temporal query grounding   & Low  &  High   &  Data: $\uparrow$, Time: $\uparrow$    \\ \hline
    Spatio-temporal Progressive Learning\citep{yang2019step}               & Video action detection             & Low & High  &  Data: $\uparrow$, Time: $\uparrow$   \\ \hline
    Spatio-temporal Structural Model\citep{zhu2013context}            & Video event recognition  & Low &  High  &  Data: $\uparrow$, Time: $\uparrow$     \\ \hline
    Hierarchical Context Learning\citep{wang2015video, wang2016hierarchical}            & Video event recognition   & Medium  & High   &  Data: $\uparrow$, Time: $\uparrow$ \\ \hline
    \end{tabular}
    }
    \label{tab:contextintegration_merit}
\end{table*}

\section{Conclusions and Future Directions}\label{secfuturedirection}
In this survey, we reviewed how context has been understand and integrated in context-based approaches for computer vision tasks. This survey covers recent context integration in both image-based tasks and video-based tasks. We categorized context in three major types and three levels, and reviewed basic deep learning architectures and datasets used in context integration. We classified the datasets according to various criteria. In the end, we compared the results of context integration. In conclusion, the context has achieved great success and outperformed over context free methods in different computer vision tasks. Context information has been integrated and utilized over context-free methods, and it has been achieved great success and surpass the performance of context-free methods, in both image-based tasks and video-based tasks. However, there are still space for further improvement and better way to incorporate the context in various tasks. Here are some potential future directions on how we can make better use of context in computer vision research.

\textbf{Contextual Data Augmentation:} We have reviewed many works using context information of different types and different levels. Context has been used over different tasks such as object detection, image recognition, pedestrian detection and so on. Most of the context integration is focused on how to aggregate context features into context-free methods and potentially improve the performance. Although context has achieved great success over context-free methods, there are less work focus on data augmentation using context information. To our best knowledge, there is only one published work \citep{dvornik2018modeling} using semantic context and local level context to augment data for small objects. Current data augmentation techniques include flipping, rotation, crop and translation, can not solve the small object detection challenges, and many state-of-art context-free methods \citep{ren2015faster, bochkovskiy2020yolov4, liu2016ssd} are still facing this challenge. Many investigations in context integration \citep{lim2021small, zhang2020putting} have shown context can really help to detect small objects, but there is no better way to augment data for small objects. We expect more methods for data augmentation by using context.

\textcolor{black}{
\textbf{Filling the Gaps in the Context Taxonomy:} 
The context taxonomy offered in this survey includes three types: spatial, temporal and other, and three levels: prior, global and local. By looking into the taxonomy, we can identify areas that have not been fully explored. For example, there are fewer works in both long-term temporal context and temporal semantic context. Other types of context than spatial and temporal need a much greater attention, especially other modalities and functionalities/intention/purpose. In terms of system architectures, convNets \citep{krizhevsky2012imagenet, szegedy2015going, simonyan2014very, he2016deep} are mainly used for visual feature extractions, neglecting non-visual features. Although there are a number of works \citep{chen2019multi, zhu2021semantic, yang2019step, tian2018audio, purushwalkam2021audio} that model relations between visual context and non-visual context, it is still challenging to better represent visual-other context relations to narrow the domain gaps. New architectures particularly designed for context learning and utilization are still needed.}

\textbf{A General Context Integration Pipeline:} Context has been integrated in different ways in both image-based tasks and video-based tasks. Many image-based context integration \citep{yang2015facial, chen2019multi, pathak2016context, leng2021realize, li2016human, mac2019presence, yang2018graph} and video-based context integration \citep{beery2020context, Wu_2020_CVPR, yan2019learning, yang2019step} implement context information into the backbone models and aggregate with the features extracted from context-free methods. Deep learning methods mainly have four stages: data pre-processing (including labeling), model training, post-processing, and result evaluation. Context information has either been aggregated during the training stage or used in the post-processing stage. No general pipelines have been proposed on how we can incorporate context through the whole process stages. Although different context integration can be used in a single stage or in multiple stages, a general pipeline is needed to guide the integration for context.

\textbf{Contextual Evaluation:} Many computer vision tasks use standard evaluation metrics to evaluate the model's performance, such as IOU for object detection. It is not necessarily equivalent to describe the accuracy in the real world. For example, if a person try to find the knob on the door, an estimated knob location (left side or right side of the door) could have more benefits than the exact knob location (1.5m high of the door on the left). A contextual evaluation based on the requirements of real-world applications is needed not only for object detection task, but may also benefit other computer vision tasks.

\section*{Acknowledgements}
\label{sec:acknowledgements}
The work is supported by the National Science Foundation (NSF) through Awards \#2131186 (CISE-MSI),  \#1827505 (PFI), and \#1737533 (S\&CC). The work is also supported by the US Air Force Office of Scientific Research (AFOSR) via Award \#FA9550-21-1-0082 and the ODNI Intelligence Community Center for Academic Excellence (IC CAE) at Rutgers University (\#HHM402-19-1-0003 and \#HHM402-18-1-0007).

\bibliographystyle{model2-names}
\bibliography{references.bib}

\begin{thebibliography}{96}
\expandafter\ifx\csname natexlab\endcsname\relax\def\natexlab#1{#1}\fi
\providecommand{\url}[1]{\texttt{#1}}
\providecommand{\href}[2]{#2}
\providecommand{\path}[1]{#1}
\providecommand{\DOIprefix}{doi:}
\providecommand{\ArXivprefix}{arXiv:}
\providecommand{\URLprefix}{URL: }
\providecommand{\Pubmedprefix}{pmid:}
\providecommand{\doi}[1]{\href{http://dx.doi.org/#1}{\path{#1}}}
\providecommand{\Pubmed}[1]{\href{pmid:#1}{\path{#1}}}
\providecommand{\bibinfo}[2]{#2}
\ifx\xfnm\relax \def\xfnm[#1]{\unskip,\space#1}\fi
%Type = Inproceedings
\bibitem[{Antol et~al.(2015)Antol, Agrawal, Lu, Mitchell, Batra, Zitnick and
  Parikh}]{antol2015vqa}
\bibinfo{author}{Antol, S.}, \bibinfo{author}{Agrawal, A.},
  \bibinfo{author}{Lu, J.}, \bibinfo{author}{Mitchell, M.},
  \bibinfo{author}{Batra, D.}, \bibinfo{author}{Zitnick, C.L.},
  \bibinfo{author}{Parikh, D.}, \bibinfo{year}{2015}.
\newblock \bibinfo{title}{Vqa: Visual question answering}, in:
  \bibinfo{booktitle}{Proceedings of the IEEE international conference on
  computer vision}, pp. \bibinfo{pages}{2425--2433}.
%Type = Article
\bibitem[{Bar and Aminoff(2003)}]{bar2003cortical}
\bibinfo{author}{Bar, M.}, \bibinfo{author}{Aminoff, E.}, \bibinfo{year}{2003}.
\newblock \bibinfo{title}{Cortical analysis of visual context}.
\newblock \bibinfo{journal}{Neuron} \bibinfo{volume}{38},
  \bibinfo{pages}{347--358}.
%Type = Inproceedings
\bibitem[{Beery et~al.(2018)Beery, Van~Horn and Perona}]{beery2018recognition}
\bibinfo{author}{Beery, S.}, \bibinfo{author}{Van~Horn, G.},
  \bibinfo{author}{Perona, P.}, \bibinfo{year}{2018}.
\newblock \bibinfo{title}{Recognition in terra incognita}, in:
  \bibinfo{booktitle}{Proceedings of the European conference on computer vision
  (ECCV)}, pp. \bibinfo{pages}{456--473}.
%Type = Inproceedings
\bibitem[{Beery et~al.(2020)Beery, Wu, Rathod, Votel and
  Huang}]{beery2020context}
\bibinfo{author}{Beery, S.}, \bibinfo{author}{Wu, G.}, \bibinfo{author}{Rathod,
  V.}, \bibinfo{author}{Votel, R.}, \bibinfo{author}{Huang, J.},
  \bibinfo{year}{2020}.
\newblock \bibinfo{title}{Context r-cnn: Long term temporal context for
  per-camera object detection}, in: \bibinfo{booktitle}{Proceedings of the
  IEEE/CVF Conference on Computer Vision and Pattern Recognition}, pp.
  \bibinfo{pages}{13075--13085}.
%Type = Article
\bibitem[{Bochkovskiy et~al.(2020)Bochkovskiy, Wang and
  Liao}]{bochkovskiy2020yolov4}
\bibinfo{author}{Bochkovskiy, A.}, \bibinfo{author}{Wang, C.Y.},
  \bibinfo{author}{Liao, H.Y.M.}, \bibinfo{year}{2020}.
\newblock \bibinfo{title}{Yolov4: Optimal speed and accuracy of object
  detection}.
\newblock \bibinfo{journal}{arXiv preprint arXiv:2004.10934} .
%Type = Inproceedings
\bibitem[{Bomatter et~al.(2021)Bomatter, Zhang, Karev, Madan, Tseng and
  Kreiman}]{bomatter2021pigs}
\bibinfo{author}{Bomatter, P.}, \bibinfo{author}{Zhang, M.},
  \bibinfo{author}{Karev, D.}, \bibinfo{author}{Madan, S.},
  \bibinfo{author}{Tseng, C.}, \bibinfo{author}{Kreiman, G.},
  \bibinfo{year}{2021}.
\newblock \bibinfo{title}{When pigs fly: Contextual reasoning in synthetic and
  natural scenes}, in: \bibinfo{booktitle}{Proceedings of the IEEE/CVF
  International Conference on Computer Vision}, pp. \bibinfo{pages}{255--264}.
%Type = Inproceedings
\bibitem[{de~Campos et~al.(2009)de~Campos, Babu and Varma}]{deCampos09}
\bibinfo{author}{de~Campos, T.E.}, \bibinfo{author}{Babu, B.R.},
  \bibinfo{author}{Varma, M.}, \bibinfo{year}{2009}.
\newblock \bibinfo{title}{Character recognition in natural images}, in:
  \bibinfo{booktitle}{Proceedings of the International Conference on Computer
  Vision Theory and Applications, Lisbon, Portugal}, pp.
  \bibinfo{pages}{273--280}.
%Type = Inproceedings
\bibitem[{Carbonetto et~al.(2004)Carbonetto, Freitas and
  Barnard}]{carbonetto2004statistical}
\bibinfo{author}{Carbonetto, P.}, \bibinfo{author}{Freitas, N.d.},
  \bibinfo{author}{Barnard, K.}, \bibinfo{year}{2004}.
\newblock \bibinfo{title}{A statistical model for general contextual object
  recognition}, in: \bibinfo{booktitle}{European conference on computer
  vision}, \bibinfo{organization}{Springer}. pp. \bibinfo{pages}{350--362}.
%Type = Inproceedings
\bibitem[{Chen et~al.(2019)Chen, Wei, Wang and Guo}]{chen2019multi}
\bibinfo{author}{Chen, Z.M.}, \bibinfo{author}{Wei, X.S.},
  \bibinfo{author}{Wang, P.}, \bibinfo{author}{Guo, Y.}, \bibinfo{year}{2019}.
\newblock \bibinfo{title}{Multi-label image recognition with graph
  convolutional networks}, in: \bibinfo{booktitle}{Proceedings of the IEEE/CVF
  Conference on Computer Vision and Pattern Recognition}, pp.
  \bibinfo{pages}{5177--5186}.
%Type = Article
\bibitem[{Choi et~al.(2011)Choi, Torralba and Willsky}]{choi2011tree}
\bibinfo{author}{Choi, M.J.}, \bibinfo{author}{Torralba, A.},
  \bibinfo{author}{Willsky, A.S.}, \bibinfo{year}{2011}.
\newblock \bibinfo{title}{A tree-based context model for object recognition}.
\newblock \bibinfo{journal}{IEEE transactions on pattern analysis and machine
  intelligence} \bibinfo{volume}{34}, \bibinfo{pages}{240--252}.
%Type = Article
\bibitem[{Choi et~al.(2012)Choi, Torralba and Willsky}]{choi2012context}
\bibinfo{author}{Choi, M.J.}, \bibinfo{author}{Torralba, A.},
  \bibinfo{author}{Willsky, A.S.}, \bibinfo{year}{2012}.
\newblock \bibinfo{title}{Context models and out-of-context objects}.
\newblock \bibinfo{journal}{Pattern Recognition Letters} \bibinfo{volume}{33},
  \bibinfo{pages}{853--862}.
%Type = Inproceedings
\bibitem[{Cordts et~al.(2016)Cordts, Omran, Ramos, Rehfeld, Enzweiler,
  Benenson, Franke, Roth and Schiele}]{Cordts2016Cityscapes}
\bibinfo{author}{Cordts, M.}, \bibinfo{author}{Omran, M.},
  \bibinfo{author}{Ramos, S.}, \bibinfo{author}{Rehfeld, T.},
  \bibinfo{author}{Enzweiler, M.}, \bibinfo{author}{Benenson, R.},
  \bibinfo{author}{Franke, U.}, \bibinfo{author}{Roth, S.},
  \bibinfo{author}{Schiele, B.}, \bibinfo{year}{2016}.
\newblock \bibinfo{title}{The cityscapes dataset for semantic urban scene
  understanding}, in: \bibinfo{booktitle}{Proc. of the IEEE Conference on
  Computer Vision and Pattern Recognition (CVPR)}.
\newblock \DOIprefix\doi{10.1109/CVPR.2016.350}.
%Type = Inproceedings
\bibitem[{Cui et~al.(2018)Cui, Xu, Zheng and Yang}]{cui2018context}
\bibinfo{author}{Cui, Z.}, \bibinfo{author}{Xu, C.}, \bibinfo{author}{Zheng,
  W.}, \bibinfo{author}{Yang, J.}, \bibinfo{year}{2018}.
\newblock \bibinfo{title}{Context-dependent diffusion network for visual
  relationship detection}, in: \bibinfo{booktitle}{Proceedings of the 26th ACM
  international conference on Multimedia}, pp. \bibinfo{pages}{1475--1482}.
%Type = Inproceedings
\bibitem[{Deng et~al.(2009)Deng, Dong, Socher, Li, Li and
  Fei-Fei}]{deng2009imagenet}
\bibinfo{author}{Deng, J.}, \bibinfo{author}{Dong, W.},
  \bibinfo{author}{Socher, R.}, \bibinfo{author}{Li, L.J.},
  \bibinfo{author}{Li, K.}, \bibinfo{author}{Fei-Fei, L.},
  \bibinfo{year}{2009}.
\newblock \bibinfo{title}{Imagenet: A large-scale hierarchical image database},
  in: \bibinfo{booktitle}{2009 IEEE conference on computer vision and pattern
  recognition}, \bibinfo{organization}{Ieee}. pp. \bibinfo{pages}{248--255}.
%Type = Inproceedings
\bibitem[{Divvala et~al.(2009)Divvala, Hoiem, Hays, Efros and
  Hebert}]{divvala2009empirical}
\bibinfo{author}{Divvala, S.K.}, \bibinfo{author}{Hoiem, D.},
  \bibinfo{author}{Hays, J.H.}, \bibinfo{author}{Efros, A.A.},
  \bibinfo{author}{Hebert, M.}, \bibinfo{year}{2009}.
\newblock \bibinfo{title}{An empirical study of context in object detection},
  in: \bibinfo{booktitle}{2009 IEEE Conference on computer vision and Pattern
  Recognition}, \bibinfo{organization}{IEEE}. pp. \bibinfo{pages}{1271--1278}.
%Type = Inproceedings
\bibitem[{Du et~al.(2012)Du, Duan and Ai}]{du2012context}
\bibinfo{author}{Du, Y.}, \bibinfo{author}{Duan, G.}, \bibinfo{author}{Ai, H.},
  \bibinfo{year}{2012}.
\newblock \bibinfo{title}{Context-based text detection in natural scenes}, in:
  \bibinfo{booktitle}{2012 19th IEEE International Conference on Image
  Processing}, \bibinfo{organization}{IEEE}. pp. \bibinfo{pages}{1857--1860}.
%Type = Inproceedings
\bibitem[{Dvornik et~al.(2018)Dvornik, Mairal and Schmid}]{dvornik2018modeling}
\bibinfo{author}{Dvornik, N.}, \bibinfo{author}{Mairal, J.},
  \bibinfo{author}{Schmid, C.}, \bibinfo{year}{2018}.
\newblock \bibinfo{title}{Modeling visual context is key to augmenting object
  detection datasets}, in: \bibinfo{booktitle}{Proceedings of the European
  Conference on Computer Vision (ECCV)}, pp. \bibinfo{pages}{364--380}.
%Type = Article
\bibitem[{Everingham et~al.(2010)Everingham, Van~Gool, Williams, Winn and
  Zisserman}]{Everingham10}
\bibinfo{author}{Everingham, M.}, \bibinfo{author}{Van~Gool, L.},
  \bibinfo{author}{Williams, C.K.I.}, \bibinfo{author}{Winn, J.},
  \bibinfo{author}{Zisserman, A.}, \bibinfo{year}{2010}.
\newblock \bibinfo{title}{The pascal visual object classes (voc) challenge}.
\newblock \bibinfo{journal}{International Journal of Computer Vision}
  \bibinfo{volume}{88}, \bibinfo{pages}{303--338}.
%Type = Inproceedings
\bibitem[{Fang et~al.(2017)Fang, Kuan, Lin, Tan and
  Chandrasekhar}]{fang2017object}
\bibinfo{author}{Fang, Y.}, \bibinfo{author}{Kuan, K.}, \bibinfo{author}{Lin,
  J.}, \bibinfo{author}{Tan, C.}, \bibinfo{author}{Chandrasekhar, V.},
  \bibinfo{year}{2017}.
\newblock \bibinfo{title}{Object detection meets knowledge graphs}, in:
  \bibinfo{booktitle}{Proceedings of the Twenty-Sixth International Joint
  Conference on Artificial Intelligence, {IJCAI-17}}, pp.
  \bibinfo{pages}{1661--1667}.
\newblock \URLprefix \url{https://doi.org/10.24963/ijcai.2017/230},
  \DOIprefix\doi{10.24963/ijcai.2017/230}.
%Type = Article
\bibitem[{Fink and Perona(2003)}]{fink2003mutual}
\bibinfo{author}{Fink, M.}, \bibinfo{author}{Perona, P.}, \bibinfo{year}{2003}.
\newblock \bibinfo{title}{Mutual boosting for contextual inference}.
\newblock \bibinfo{journal}{Advances in neural information processing systems}
  \bibinfo{volume}{16}.
%Type = Article
\bibitem[{Goh et~al.(2004)Goh, Siong, Park, Gutchess, Hebrank and
  Chee}]{goh2004cortical}
\bibinfo{author}{Goh, J.O.}, \bibinfo{author}{Siong, S.C.},
  \bibinfo{author}{Park, D.}, \bibinfo{author}{Gutchess, A.},
  \bibinfo{author}{Hebrank, A.}, \bibinfo{author}{Chee, M.W.},
  \bibinfo{year}{2004}.
\newblock \bibinfo{title}{Cortical areas involved in object, background, and
  object-background processing revealed with functional magnetic resonance
  adaptation}.
\newblock \bibinfo{journal}{Journal of Neuroscience} \bibinfo{volume}{24},
  \bibinfo{pages}{10223--10228}.
%Type = Inproceedings
\bibitem[{Gu et~al.(2018)Gu, Sun, Ross, Vondrick, Pantofaru, Li,
  Vijayanarasimhan, Toderici, Ricco, Sukthankar et~al.}]{gu2018ava}
\bibinfo{author}{Gu, C.}, \bibinfo{author}{Sun, C.}, \bibinfo{author}{Ross,
  D.A.}, \bibinfo{author}{Vondrick, C.}, \bibinfo{author}{Pantofaru, C.},
  \bibinfo{author}{Li, Y.}, \bibinfo{author}{Vijayanarasimhan, S.},
  \bibinfo{author}{Toderici, G.}, \bibinfo{author}{Ricco, S.},
  \bibinfo{author}{Sukthankar, R.}, et~al., \bibinfo{year}{2018}.
\newblock \bibinfo{title}{Ava: A video dataset of spatio-temporally localized
  atomic visual actions}, in: \bibinfo{booktitle}{Proceedings of the IEEE
  Conference on Computer Vision and Pattern Recognition}, pp.
  \bibinfo{pages}{6047--6056}.
%Type = Inproceedings
\bibitem[{Hara et~al.(2014)Hara, Sun, Moore, Jacobs and
  Froehlich}]{hara2014tohme}
\bibinfo{author}{Hara, K.}, \bibinfo{author}{Sun, J.}, \bibinfo{author}{Moore,
  R.}, \bibinfo{author}{Jacobs, D.}, \bibinfo{author}{Froehlich, J.},
  \bibinfo{year}{2014}.
\newblock \bibinfo{title}{Tohme: detecting curb ramps in google street view
  using crowdsourcing, computer vision, and machine learning}, in:
  \bibinfo{booktitle}{Proceedings of the 27th annual ACM symposium on User
  interface software and technology}, pp. \bibinfo{pages}{189--204}.
%Type = Inproceedings
\bibitem[{He et~al.(2016)He, Zhang, Ren and Sun}]{he2016deep}
\bibinfo{author}{He, K.}, \bibinfo{author}{Zhang, X.}, \bibinfo{author}{Ren,
  S.}, \bibinfo{author}{Sun, J.}, \bibinfo{year}{2016}.
\newblock \bibinfo{title}{Deep residual learning for image recognition}, in:
  \bibinfo{booktitle}{Proceedings of the IEEE conference on computer vision and
  pattern recognition}, pp. \bibinfo{pages}{770--778}.
%Type = Inproceedings
\bibitem[{Heitz and Koller(2008)}]{heitz2008learning}
\bibinfo{author}{Heitz, G.}, \bibinfo{author}{Koller, D.},
  \bibinfo{year}{2008}.
\newblock \bibinfo{title}{Learning spatial context: Using stuff to find
  things}, in: \bibinfo{booktitle}{European conference on computer vision},
  \bibinfo{organization}{Springer}. pp. \bibinfo{pages}{30--43}.
%Type = Inproceedings
\bibitem[{Hwang et~al.(2015)Hwang, Park, Kim, Choi and
  Kweon}]{hwang2015multispectral}
\bibinfo{author}{Hwang, S.}, \bibinfo{author}{Park, J.}, \bibinfo{author}{Kim,
  N.}, \bibinfo{author}{Choi, Y.}, \bibinfo{author}{Kweon, I.S.},
  \bibinfo{year}{2015}.
\newblock \bibinfo{title}{Multispectral pedestrian detection: Benchmark dataset
  and baselines}, in: \bibinfo{booktitle}{Proceedings of IEEE Conference on
  Computer Vision and Pattern Recognition (CVPR)}.
%Type = Inproceedings
\bibitem[{Johnson et~al.(2018)Johnson, Gupta and Fei-Fei}]{johnson2018image}
\bibinfo{author}{Johnson, J.}, \bibinfo{author}{Gupta, A.},
  \bibinfo{author}{Fei-Fei, L.}, \bibinfo{year}{2018}.
\newblock \bibinfo{title}{Image generation from scene graphs}, in:
  \bibinfo{booktitle}{Proceedings of the IEEE conference on computer vision and
  pattern recognition}, pp. \bibinfo{pages}{1219--1228}.
%Type = Inproceedings
\bibitem[{Johnson et~al.(2015)Johnson, Krishna, Stark, Li, Shamma, Bernstein
  and Fei-Fei}]{johnson2015image}
\bibinfo{author}{Johnson, J.}, \bibinfo{author}{Krishna, R.},
  \bibinfo{author}{Stark, M.}, \bibinfo{author}{Li, L.J.},
  \bibinfo{author}{Shamma, D.}, \bibinfo{author}{Bernstein, M.},
  \bibinfo{author}{Fei-Fei, L.}, \bibinfo{year}{2015}.
\newblock \bibinfo{title}{Image retrieval using scene graphs}, in:
  \bibinfo{booktitle}{Proceedings of the IEEE conference on computer vision and
  pattern recognition}, pp. \bibinfo{pages}{3668--3678}.
%Type = Article
\bibitem[{Kipf and Welling(2016)}]{kipf2016semi}
\bibinfo{author}{Kipf, T.N.}, \bibinfo{author}{Welling, M.},
  \bibinfo{year}{2016}.
\newblock \bibinfo{title}{Semi-supervised classification with graph
  convolutional networks}.
\newblock \bibinfo{journal}{arXiv preprint arXiv:1609.02907} .
%Type = Article
\bibitem[{Krishna et~al.(2017)Krishna, Zhu, Groth, Johnson, Hata, Kravitz,
  Chen, Kalantidis, Li, Shamma et~al.}]{krishna2017visual}
\bibinfo{author}{Krishna, R.}, \bibinfo{author}{Zhu, Y.},
  \bibinfo{author}{Groth, O.}, \bibinfo{author}{Johnson, J.},
  \bibinfo{author}{Hata, K.}, \bibinfo{author}{Kravitz, J.},
  \bibinfo{author}{Chen, S.}, \bibinfo{author}{Kalantidis, Y.},
  \bibinfo{author}{Li, L.J.}, \bibinfo{author}{Shamma, D.A.}, et~al.,
  \bibinfo{year}{2017}.
\newblock \bibinfo{title}{Visual genome: Connecting language and vision using
  crowdsourced dense image annotations}.
\newblock \bibinfo{journal}{International journal of computer vision}
  \bibinfo{volume}{123}, \bibinfo{pages}{32--73}.
%Type = Article
\bibitem[{Krizhevsky et~al.(2012)Krizhevsky, Sutskever and
  Hinton}]{krizhevsky2012imagenet}
\bibinfo{author}{Krizhevsky, A.}, \bibinfo{author}{Sutskever, I.},
  \bibinfo{author}{Hinton, G.E.}, \bibinfo{year}{2012}.
\newblock \bibinfo{title}{Imagenet classification with deep convolutional
  neural networks}.
\newblock \bibinfo{journal}{Advances in neural information processing systems}
  \bibinfo{volume}{25}.
%Type = Inproceedings
\bibitem[{Lai et~al.(2021)Lai, Purushwalkam and Gupta}]{lai2021functional}
\bibinfo{author}{Lai, Z.}, \bibinfo{author}{Purushwalkam, S.},
  \bibinfo{author}{Gupta, A.}, \bibinfo{year}{2021}.
\newblock \bibinfo{title}{The functional correspondence problem}, in:
  \bibinfo{booktitle}{Proceedings of the IEEE International Conference on
  Computer Vision}, pp. \bibinfo{pages}{15772--15781}.
%Type = Article
\bibitem[{Leng et~al.(2021)Leng, Ren, Jiang, Sun and Wang}]{leng2021realize}
\bibinfo{author}{Leng, J.}, \bibinfo{author}{Ren, Y.}, \bibinfo{author}{Jiang,
  W.}, \bibinfo{author}{Sun, X.}, \bibinfo{author}{Wang, Y.},
  \bibinfo{year}{2021}.
\newblock \bibinfo{title}{Realize your surroundings: Exploiting context
  information for small object detection}.
\newblock \bibinfo{journal}{Neurocomputing} \bibinfo{volume}{433},
  \bibinfo{pages}{287--299}.
%Type = Inproceedings
\bibitem[{Li et~al.(2015)Li, Lin, Shen, Brandt and Hua}]{li2015convolutional}
\bibinfo{author}{Li, H.}, \bibinfo{author}{Lin, Z.}, \bibinfo{author}{Shen,
  X.}, \bibinfo{author}{Brandt, J.}, \bibinfo{author}{Hua, G.},
  \bibinfo{year}{2015}.
\newblock \bibinfo{title}{A convolutional neural network cascade for face
  detection}, in: \bibinfo{booktitle}{Proceedings of the IEEE conference on
  computer vision and pattern recognition}, pp. \bibinfo{pages}{5325--5334}.
%Type = Inproceedings
\bibitem[{Li et~al.(2016)Li, Huang, Loy and Tang}]{li2016human}
\bibinfo{author}{Li, Y.}, \bibinfo{author}{Huang, C.}, \bibinfo{author}{Loy,
  C.C.}, \bibinfo{author}{Tang, X.}, \bibinfo{year}{2016}.
\newblock \bibinfo{title}{Human attribute recognition by deep hierarchical
  contexts}, in: \bibinfo{booktitle}{European Conference on Computer Vision},
  \bibinfo{organization}{Springer}. pp. \bibinfo{pages}{684--700}.
%Type = Inproceedings
\bibitem[{Lim et~al.(2021)Lim, Astrid, Yoon and Lee}]{lim2021small}
\bibinfo{author}{Lim, J.S.}, \bibinfo{author}{Astrid, M.},
  \bibinfo{author}{Yoon, H.J.}, \bibinfo{author}{Lee, S.I.},
  \bibinfo{year}{2021}.
\newblock \bibinfo{title}{Small object detection using context and attention},
  in: \bibinfo{booktitle}{2021 International Conference on Artificial
  Intelligence in Information and Communication (ICAIIC)},
  \bibinfo{organization}{IEEE}. pp. \bibinfo{pages}{181--186}.
%Type = Inproceedings
\bibitem[{Lin et~al.(2014)Lin, Maire, Belongie, Hays, Perona, Ramanan,
  Doll{\'a}r and Zitnick}]{lin2014microsoft}
\bibinfo{author}{Lin, T.Y.}, \bibinfo{author}{Maire, M.},
  \bibinfo{author}{Belongie, S.}, \bibinfo{author}{Hays, J.},
  \bibinfo{author}{Perona, P.}, \bibinfo{author}{Ramanan, D.},
  \bibinfo{author}{Doll{\'a}r, P.}, \bibinfo{author}{Zitnick, C.L.},
  \bibinfo{year}{2014}.
\newblock \bibinfo{title}{Microsoft coco: Common objects in context}, in:
  \bibinfo{booktitle}{European conference on computer vision},
  \bibinfo{organization}{Springer}. pp. \bibinfo{pages}{740--755}.
%Type = Inproceedings
\bibitem[{Liu et~al.(2016)Liu, Anguelov, Erhan, Szegedy, Reed, Fu and
  Berg}]{liu2016ssd}
\bibinfo{author}{Liu, W.}, \bibinfo{author}{Anguelov, D.},
  \bibinfo{author}{Erhan, D.}, \bibinfo{author}{Szegedy, C.},
  \bibinfo{author}{Reed, S.}, \bibinfo{author}{Fu, C.Y.},
  \bibinfo{author}{Berg, A.C.}, \bibinfo{year}{2016}.
\newblock \bibinfo{title}{Ssd: Single shot multibox detector}, in:
  \bibinfo{booktitle}{European conference on computer vision},
  \bibinfo{organization}{Springer}. pp. \bibinfo{pages}{21--37}.
%Type = Inproceedings
\bibitem[{Liu et~al.(2015)Liu, Luo, Wang and Tang}]{liu2015faceattributes}
\bibinfo{author}{Liu, Z.}, \bibinfo{author}{Luo, P.}, \bibinfo{author}{Wang,
  X.}, \bibinfo{author}{Tang, X.}, \bibinfo{year}{2015}.
\newblock \bibinfo{title}{Deep learning face attributes in the wild}, in:
  \bibinfo{booktitle}{Proceedings of International Conference on Computer
  Vision (ICCV)}.
%Type = Inproceedings
\bibitem[{Mac~Aodha et~al.(2019)Mac~Aodha, Cole and Perona}]{mac2019presence}
\bibinfo{author}{Mac~Aodha, O.}, \bibinfo{author}{Cole, E.},
  \bibinfo{author}{Perona, P.}, \bibinfo{year}{2019}.
\newblock \bibinfo{title}{Presence-only geographical priors for fine-grained
  image classification}, in: \bibinfo{booktitle}{Proceedings of the IEEE/CVF
  International Conference on Computer Vision}, pp.
  \bibinfo{pages}{9596--9606}.
%Type = Article
\bibitem[{Marques et~al.(2011)Marques, Barenholtz and
  Charvillat}]{marques2011context}
\bibinfo{author}{Marques, O.}, \bibinfo{author}{Barenholtz, E.},
  \bibinfo{author}{Charvillat, V.}, \bibinfo{year}{2011}.
\newblock \bibinfo{title}{Context modeling in computer vision: techniques,
  implications, and applications}.
\newblock \bibinfo{journal}{Multimedia Tools and Applications}
  \bibinfo{volume}{51}, \bibinfo{pages}{303--339}.
%Type = Inproceedings
\bibitem[{Mathias et~al.(2014)Mathias, Benenson, Pedersoli and
  Gool}]{mathias2014face}
\bibinfo{author}{Mathias, M.}, \bibinfo{author}{Benenson, R.},
  \bibinfo{author}{Pedersoli, M.}, \bibinfo{author}{Gool, L.V.},
  \bibinfo{year}{2014}.
\newblock \bibinfo{title}{Face detection without bells and whistles}, in:
  \bibinfo{booktitle}{European conference on computer vision},
  \bibinfo{organization}{Springer}. pp. \bibinfo{pages}{720--735}.
%Type = Inproceedings
\bibitem[{Mottaghi et~al.(2014)Mottaghi, Chen, Liu, Cho, Lee, Fidler, Urtasun
  and Yuille}]{mottaghi2014role}
\bibinfo{author}{Mottaghi, R.}, \bibinfo{author}{Chen, X.},
  \bibinfo{author}{Liu, X.}, \bibinfo{author}{Cho, N.G.}, \bibinfo{author}{Lee,
  S.W.}, \bibinfo{author}{Fidler, S.}, \bibinfo{author}{Urtasun, R.},
  \bibinfo{author}{Yuille, A.}, \bibinfo{year}{2014}.
\newblock \bibinfo{title}{The role of context for object detection and semantic
  segmentation in the wild}, in: \bibinfo{booktitle}{Proceedings of the IEEE
  conference on computer vision and pattern recognition}, pp.
  \bibinfo{pages}{891--898}.
%Type = Inproceedings
\bibitem[{Mottaghi et~al.(2013)Mottaghi, Fidler, Yao, Urtasun and
  Parikh}]{mottaghi2013analyzing}
\bibinfo{author}{Mottaghi, R.}, \bibinfo{author}{Fidler, S.},
  \bibinfo{author}{Yao, J.}, \bibinfo{author}{Urtasun, R.},
  \bibinfo{author}{Parikh, D.}, \bibinfo{year}{2013}.
\newblock \bibinfo{title}{Analyzing semantic segmentation using hybrid
  human-machine crfs}, in: \bibinfo{booktitle}{Proceedings of the IEEE
  Conference on Computer Vision and Pattern Recognition}, pp.
  \bibinfo{pages}{3143--3150}.
%Type = Inproceedings
\bibitem[{Oh et~al.(2011)Oh, Hoogs, Perera, Cuntoor, Chen, Lee, Mukherjee,
  Aggarwal, Lee, Davis et~al.}]{oh2011large}
\bibinfo{author}{Oh, S.}, \bibinfo{author}{Hoogs, A.}, \bibinfo{author}{Perera,
  A.}, \bibinfo{author}{Cuntoor, N.}, \bibinfo{author}{Chen, C.C.},
  \bibinfo{author}{Lee, J.T.}, \bibinfo{author}{Mukherjee, S.},
  \bibinfo{author}{Aggarwal, J.}, \bibinfo{author}{Lee, H.},
  \bibinfo{author}{Davis, L.}, et~al., \bibinfo{year}{2011}.
\newblock \bibinfo{title}{A large-scale benchmark dataset for event recognition
  in surveillance video}, in: \bibinfo{booktitle}{CVPR 2011},
  \bibinfo{organization}{IEEE}. pp. \bibinfo{pages}{3153--3160}.
%Type = Article
\bibitem[{Palmer(1975)}]{Palmer}
\bibinfo{author}{Palmer, S.}, \bibinfo{year}{1975}.
\newblock \bibinfo{title}{The effects of contextual scenes on the
  identification of objects.}
\newblock \bibinfo{journal}{Memory \& Cognition} \bibinfo{volume}{3},
  \bibinfo{pages}{519--526}.
%Type = Inproceedings
\bibitem[{Pathak et~al.(2016)Pathak, Krahenbuhl, Donahue, Darrell and
  Efros}]{pathak2016context}
\bibinfo{author}{Pathak, D.}, \bibinfo{author}{Krahenbuhl, P.},
  \bibinfo{author}{Donahue, J.}, \bibinfo{author}{Darrell, T.},
  \bibinfo{author}{Efros, A.A.}, \bibinfo{year}{2016}.
\newblock \bibinfo{title}{Context encoders: Feature learning by inpainting},
  in: \bibinfo{booktitle}{Proceedings of the IEEE conference on computer vision
  and pattern recognition}, pp. \bibinfo{pages}{2536--2544}.
%Type = Article
\bibitem[{Perko and Leonardis(2010)}]{perko2010framework}
\bibinfo{author}{Perko, R.}, \bibinfo{author}{Leonardis, A.},
  \bibinfo{year}{2010}.
\newblock \bibinfo{title}{A framework for visual-context-aware object detection
  in still images}.
\newblock \bibinfo{journal}{Computer Vision and Image Understanding}
  \bibinfo{volume}{114}, \bibinfo{pages}{700--711}.
%Type = Inproceedings
\bibitem[{Purushwalkam et~al.(2021)Purushwalkam, Gari, Ithapu, Schissler,
  Robinson, Gupta and Grauman}]{purushwalkam2021audio}
\bibinfo{author}{Purushwalkam, S.}, \bibinfo{author}{Gari, S.V.A.},
  \bibinfo{author}{Ithapu, V.K.}, \bibinfo{author}{Schissler, C.},
  \bibinfo{author}{Robinson, P.}, \bibinfo{author}{Gupta, A.},
  \bibinfo{author}{Grauman, K.}, \bibinfo{year}{2021}.
\newblock \bibinfo{title}{Audio-visual floorplan reconstruction}, in:
  \bibinfo{booktitle}{Proceedings of the IEEE International Conference on
  Computer Vision}, pp. \bibinfo{pages}{1183--1192}.
%Type = Inproceedings
\bibitem[{Rabinovich and Belongie(2009)}]{rabinovich2009scenes}
\bibinfo{author}{Rabinovich, A.}, \bibinfo{author}{Belongie, S.},
  \bibinfo{year}{2009}.
\newblock \bibinfo{title}{Scenes vs. objects: a comparative study of two
  approaches to context based recognition}, in: \bibinfo{booktitle}{2009 IEEE
  Computer Society Conference on Computer Vision and Pattern Recognition
  Workshops}, \bibinfo{organization}{IEEE}. pp. \bibinfo{pages}{92--99}.
%Type = Inproceedings
\bibitem[{Rabinovich et~al.(2007)Rabinovich, Vedaldi, Galleguillos, Wiewiora
  and Belongie}]{rabinovich2007objects}
\bibinfo{author}{Rabinovich, A.}, \bibinfo{author}{Vedaldi, A.},
  \bibinfo{author}{Galleguillos, C.}, \bibinfo{author}{Wiewiora, E.},
  \bibinfo{author}{Belongie, S.}, \bibinfo{year}{2007}.
\newblock \bibinfo{title}{Objects in context}, in: \bibinfo{booktitle}{2007
  IEEE 11th International Conference on Computer Vision},
  \bibinfo{organization}{IEEE}. pp. \bibinfo{pages}{1--8}.
%Type = Article
\bibitem[{Ren et~al.(2015)Ren, He, Girshick and Sun}]{ren2015faster}
\bibinfo{author}{Ren, S.}, \bibinfo{author}{He, K.}, \bibinfo{author}{Girshick,
  R.}, \bibinfo{author}{Sun, J.}, \bibinfo{year}{2015}.
\newblock \bibinfo{title}{Faster r-cnn: Towards real-time object detection with
  region proposal networks}.
\newblock \bibinfo{journal}{Advances in neural information processing systems}
  \bibinfo{volume}{28}, \bibinfo{pages}{91--99}.
%Type = Article
\bibitem[{Russell et~al.(2008)Russell, Torralba, Murphy and
  Freeman}]{russell2008labelme}
\bibinfo{author}{Russell, B.C.}, \bibinfo{author}{Torralba, A.},
  \bibinfo{author}{Murphy, K.P.}, \bibinfo{author}{Freeman, W.T.},
  \bibinfo{year}{2008}.
\newblock \bibinfo{title}{Labelme: a database and web-based tool for image
  annotation}.
\newblock \bibinfo{journal}{International journal of computer vision}
  \bibinfo{volume}{77}, \bibinfo{pages}{157--173}.
%Type = Misc
\bibitem[{Ryoo and Aggarwal(2010)}]{UT-Interaction-Data}
\bibinfo{author}{Ryoo, M.S.}, \bibinfo{author}{Aggarwal, J.K.},
  \bibinfo{year}{2010}.
\newblock \bibinfo{title}{{UT}-{I}nteraction {D}ataset, {ICPR} contest on
  {S}emantic {D}escription of {H}uman {A}ctivities ({SDHA})}.
\newblock
  \bibinfo{howpublished}{http://cvrc.ece.utexas.edu/SDHA2010/Human\_Interaction.html}.
%Type = Article
\bibitem[{Sabir et~al.(2018)Sabir, Moreno-Noguer and
  Padr{\'o}}]{sabir2018enhancing}
\bibinfo{author}{Sabir, A.}, \bibinfo{author}{Moreno-Noguer, F.},
  \bibinfo{author}{Padr{\'o}, L.}, \bibinfo{year}{2018}.
\newblock \bibinfo{title}{Enhancing text spotting with a language model and
  visual context information} .
%Type = Article
\bibitem[{Seymour et~al.(2017)Seymour, Dale, Hammill, Halpin and
  Johnston}]{seymour2017automated}
\bibinfo{author}{Seymour, A.}, \bibinfo{author}{Dale, J.},
  \bibinfo{author}{Hammill, M.}, \bibinfo{author}{Halpin, P.},
  \bibinfo{author}{Johnston, D.}, \bibinfo{year}{2017}.
\newblock \bibinfo{title}{Automated detection and enumeration of marine
  wildlife using unmanned aircraft systems (uas) and thermal imagery}.
\newblock \bibinfo{journal}{Scientific reports} \bibinfo{volume}{7},
  \bibinfo{pages}{1--10}.
%Type = Article
\bibitem[{Shotton et~al.(2009)Shotton, Winn, Rother and
  Criminisi}]{shotton2009textonboost}
\bibinfo{author}{Shotton, J.}, \bibinfo{author}{Winn, J.},
  \bibinfo{author}{Rother, C.}, \bibinfo{author}{Criminisi, A.},
  \bibinfo{year}{2009}.
\newblock \bibinfo{title}{Textonboost for image understanding: Multi-class
  object recognition and segmentation by jointly modeling texture, layout, and
  context}.
\newblock \bibinfo{journal}{International journal of computer vision}
  \bibinfo{volume}{81}, \bibinfo{pages}{2--23}.
%Type = Article
\bibitem[{Simonyan and Zisserman(2014)}]{simonyan2014very}
\bibinfo{author}{Simonyan, K.}, \bibinfo{author}{Zisserman, A.},
  \bibinfo{year}{2014}.
\newblock \bibinfo{title}{Very deep convolutional networks for large-scale
  image recognition}.
\newblock \bibinfo{journal}{arXiv preprint arXiv:1409.1556} .
%Type = Inproceedings
\bibitem[{Singhal et~al.(2003)Singhal, Luo and Zhu}]{singhal2003probabilistic}
\bibinfo{author}{Singhal, A.}, \bibinfo{author}{Luo, J.}, \bibinfo{author}{Zhu,
  W.}, \bibinfo{year}{2003}.
\newblock \bibinfo{title}{Probabilistic spatial context models for scene
  content understanding}, in: \bibinfo{booktitle}{2003 IEEE Computer Society
  Conference on Computer Vision and Pattern Recognition, 2003. Proceedings.},
  \bibinfo{organization}{IEEE}. pp. \bibinfo{pages}{I--I}.
%Type = Article
\bibitem[{Soomro et~al.(2012)Soomro, Zamir and Shah}]{soomro2012ucf101}
\bibinfo{author}{Soomro, K.}, \bibinfo{author}{Zamir, A.R.},
  \bibinfo{author}{Shah, M.}, \bibinfo{year}{2012}.
\newblock \bibinfo{title}{Ucf101: A dataset of 101 human actions classes from
  videos in the wild}.
\newblock \bibinfo{journal}{arXiv preprint arXiv:1212.0402} .
%Type = Article
\bibitem[{Strat and Fischler(1991)}]{strat1991context}
\bibinfo{author}{Strat, T.M.}, \bibinfo{author}{Fischler, M.A.},
  \bibinfo{year}{1991}.
\newblock \bibinfo{title}{Context-based vision: recognizing objects using
  information from both 2 d and 3 d imagery}.
\newblock \bibinfo{journal}{IEEE Transactions on Pattern Analysis and Machine
  Intelligence} \bibinfo{volume}{13}, \bibinfo{pages}{1050--1065}.
%Type = Inproceedings
\bibitem[{Sun and Jacobs(2017)}]{sun2017seeing}
\bibinfo{author}{Sun, J.}, \bibinfo{author}{Jacobs, D.W.},
  \bibinfo{year}{2017}.
\newblock \bibinfo{title}{Seeing what is not there: Learning context to
  determine where objects are missing}, in: \bibinfo{booktitle}{Proceedings of
  the IEEE Conference on Computer Vision and Pattern Recognition}, pp.
  \bibinfo{pages}{5716--5724}.
%Type = Article
\bibitem[{Swanson et~al.(2015)Swanson, Kosmala, Lintott, Simpson, Smith and
  Packer}]{swanson2015snapshot}
\bibinfo{author}{Swanson, A.}, \bibinfo{author}{Kosmala, M.},
  \bibinfo{author}{Lintott, C.}, \bibinfo{author}{Simpson, R.},
  \bibinfo{author}{Smith, A.}, \bibinfo{author}{Packer, C.},
  \bibinfo{year}{2015}.
\newblock \bibinfo{title}{Snapshot serengeti, high-frequency annotated camera
  trap images of 40 mammalian species in an african savanna}.
\newblock \bibinfo{journal}{Scientific data} \bibinfo{volume}{2},
  \bibinfo{pages}{1--14}.
%Type = Inproceedings
\bibitem[{Szegedy et~al.(2015)Szegedy, Liu, Jia, Sermanet, Reed, Anguelov,
  Erhan, Vanhoucke and Rabinovich}]{szegedy2015going}
\bibinfo{author}{Szegedy, C.}, \bibinfo{author}{Liu, W.}, \bibinfo{author}{Jia,
  Y.}, \bibinfo{author}{Sermanet, P.}, \bibinfo{author}{Reed, S.},
  \bibinfo{author}{Anguelov, D.}, \bibinfo{author}{Erhan, D.},
  \bibinfo{author}{Vanhoucke, V.}, \bibinfo{author}{Rabinovich, A.},
  \bibinfo{year}{2015}.
\newblock \bibinfo{title}{Going deeper with convolutions}, in:
  \bibinfo{booktitle}{Proceedings of the IEEE conference on computer vision and
  pattern recognition}, pp. \bibinfo{pages}{1--9}.
%Type = Inproceedings
\bibitem[{Tang et~al.(2019)Tang, Zhang, Wu, Luo and Liu}]{tang2019learning}
\bibinfo{author}{Tang, K.}, \bibinfo{author}{Zhang, H.}, \bibinfo{author}{Wu,
  B.}, \bibinfo{author}{Luo, W.}, \bibinfo{author}{Liu, W.},
  \bibinfo{year}{2019}.
\newblock \bibinfo{title}{Learning to compose dynamic tree structures for
  visual contexts}, in: \bibinfo{booktitle}{Proceedings of the IEEE/CVF
  Conference on Computer Vision and Pattern Recognition}, pp.
  \bibinfo{pages}{6619--6628}.
%Type = Inproceedings
\bibitem[{Tian et~al.(2018)Tian, Shi, Li, Duan and Xu}]{tian2018audio}
\bibinfo{author}{Tian, Y.}, \bibinfo{author}{Shi, J.}, \bibinfo{author}{Li,
  B.}, \bibinfo{author}{Duan, Z.}, \bibinfo{author}{Xu, C.},
  \bibinfo{year}{2018}.
\newblock \bibinfo{title}{Audio-visual event localization in unconstrained
  videos}, in: \bibinfo{booktitle}{Proceedings of the European Conference on
  Computer Vision}, pp. \bibinfo{pages}{247--263}.
%Type = Article
\bibitem[{Torralba(2003)}]{torralba2003contextual}
\bibinfo{author}{Torralba, A.}, \bibinfo{year}{2003}.
\newblock \bibinfo{title}{Contextual priming for object detection}.
\newblock \bibinfo{journal}{International journal of computer vision}
  \bibinfo{volume}{53}, \bibinfo{pages}{169--191}.
%Type = Article
\bibitem[{Torralba et~al.(2010)Torralba, Murphy and
  Freeman}]{torralba2010using}
\bibinfo{author}{Torralba, A.}, \bibinfo{author}{Murphy, K.P.},
  \bibinfo{author}{Freeman, W.T.}, \bibinfo{year}{2010}.
\newblock \bibinfo{title}{Using the forest to see the trees: exploiting context
  for visual object detection and localization}.
\newblock \bibinfo{journal}{Communications of the ACM} \bibinfo{volume}{53},
  \bibinfo{pages}{107--114}.
%Type = Inproceedings
\bibitem[{Van~Horn et~al.(2018)Van~Horn, Mac~Aodha, Song, Cui, Sun, Shepard,
  Adam, Perona and Belongie}]{van2018inaturalist}
\bibinfo{author}{Van~Horn, G.}, \bibinfo{author}{Mac~Aodha, O.},
  \bibinfo{author}{Song, Y.}, \bibinfo{author}{Cui, Y.}, \bibinfo{author}{Sun,
  C.}, \bibinfo{author}{Shepard, A.}, \bibinfo{author}{Adam, H.},
  \bibinfo{author}{Perona, P.}, \bibinfo{author}{Belongie, S.},
  \bibinfo{year}{2018}.
\newblock \bibinfo{title}{The inaturalist species classification and detection
  dataset}, in: \bibinfo{booktitle}{Proceedings of the IEEE conference on
  computer vision and pattern recognition}, pp. \bibinfo{pages}{8769--8778}.
%Type = Article
\bibitem[{V{\~o}(2021)}]{vo2021meaning}
\bibinfo{author}{V{\~o}, M.L.H.}, \bibinfo{year}{2021}.
\newblock \bibinfo{title}{The meaning and structure of scenes}.
\newblock \bibinfo{journal}{Vision Research} \bibinfo{volume}{181},
  \bibinfo{pages}{10--20}.
%Type = Inproceedings
\bibitem[{Wang et~al.(2020)Wang, Ma and Jiang}]{wang2020temporally}
\bibinfo{author}{Wang, J.}, \bibinfo{author}{Ma, L.}, \bibinfo{author}{Jiang,
  W.}, \bibinfo{year}{2020}.
\newblock \bibinfo{title}{Temporally grounding language queries in videos by
  contextual boundary-aware prediction}, in: \bibinfo{booktitle}{Proceedings of
  the AAAI Conference on Artificial Intelligence}, pp.
  \bibinfo{pages}{12168--12175}.
%Type = Inproceedings
\bibitem[{Wang et~al.(2011)Wang, Babenko and Belongie}]{wang2011end}
\bibinfo{author}{Wang, K.}, \bibinfo{author}{Babenko, B.},
  \bibinfo{author}{Belongie, S.}, \bibinfo{year}{2011}.
\newblock \bibinfo{title}{End-to-end scene text recognition}, in:
  \bibinfo{booktitle}{2011 International Conference on Computer Vision},
  \bibinfo{organization}{IEEE}. pp. \bibinfo{pages}{1457--1464}.
%Type = Inproceedings
\bibitem[{Wang et~al.(2007)Wang, Doretto, Sebastian, Rittscher and
  Tu}]{wang2007shape}
\bibinfo{author}{Wang, X.}, \bibinfo{author}{Doretto, G.},
  \bibinfo{author}{Sebastian, T.}, \bibinfo{author}{Rittscher, J.},
  \bibinfo{author}{Tu, P.}, \bibinfo{year}{2007}.
\newblock \bibinfo{title}{Shape and appearance context modeling}, in:
  \bibinfo{booktitle}{2007 ieee 11th international conference on computer
  vision}, \bibinfo{organization}{Ieee}. pp. \bibinfo{pages}{1--8}.
%Type = Inproceedings
\bibitem[{Wang and Ji(2012)}]{wang2012incorporating}
\bibinfo{author}{Wang, X.}, \bibinfo{author}{Ji, Q.}, \bibinfo{year}{2012}.
\newblock \bibinfo{title}{Incorporating contextual knowledge to dynamic
  bayesian networks for event recognition}, in: \bibinfo{booktitle}{Proceedings
  of the 21st International Conference on Pattern Recognition (ICPR2012)},
  \bibinfo{organization}{IEEE}. pp. \bibinfo{pages}{3378--3381}.
%Type = Inproceedings
\bibitem[{Wang and Ji(2015)}]{wang2015video}
\bibinfo{author}{Wang, X.}, \bibinfo{author}{Ji, Q.}, \bibinfo{year}{2015}.
\newblock \bibinfo{title}{Video event recognition with deep hierarchical
  context model}, in: \bibinfo{booktitle}{Proceedings of the IEEE Conference on
  Computer Vision and Pattern Recognition}, pp. \bibinfo{pages}{4418--4427}.
%Type = Article
\bibitem[{Wang and Ji(2016)}]{wang2016hierarchical}
\bibinfo{author}{Wang, X.}, \bibinfo{author}{Ji, Q.}, \bibinfo{year}{2016}.
\newblock \bibinfo{title}{Hierarchical context modeling for video event
  recognition}.
\newblock \bibinfo{journal}{IEEE transactions on pattern analysis and machine
  intelligence} \bibinfo{volume}{39}, \bibinfo{pages}{1770--1782}.
%Type = Article
\bibitem[{Wolf and Bileschi(2006)}]{wolf2006critical}
\bibinfo{author}{Wolf, L.}, \bibinfo{author}{Bileschi, S.},
  \bibinfo{year}{2006}.
\newblock \bibinfo{title}{A critical view of context}.
\newblock \bibinfo{journal}{International Journal of Computer Vision}
  \bibinfo{volume}{69}, \bibinfo{pages}{251--261}.
%Type = Inproceedings
\bibitem[{Wu et~al.(2020)Wu, Zhou, Yang, Zhang, Li and Yuan}]{Wu_2020_CVPR}
\bibinfo{author}{Wu, J.}, \bibinfo{author}{Zhou, C.}, \bibinfo{author}{Yang,
  M.}, \bibinfo{author}{Zhang, Q.}, \bibinfo{author}{Li, Y.},
  \bibinfo{author}{Yuan, J.}, \bibinfo{year}{2020}.
\newblock \bibinfo{title}{Temporal-context enhanced detection of heavily
  occluded pedestrians}, in: \bibinfo{booktitle}{Proceedings of the IEEE/CVF
  Conference on Computer Vision and Pattern Recognition (CVPR)}.
%Type = Inproceedings
\bibitem[{Xiao et~al.(2017)Xiao, Li, Wang, Lin and Wang}]{xiao2017joint}
\bibinfo{author}{Xiao, T.}, \bibinfo{author}{Li, S.}, \bibinfo{author}{Wang,
  B.}, \bibinfo{author}{Lin, L.}, \bibinfo{author}{Wang, X.},
  \bibinfo{year}{2017}.
\newblock \bibinfo{title}{Joint detection and identification feature learning
  for person search}, in: \bibinfo{booktitle}{Proceedings of the IEEE
  Conference on Computer Vision and Pattern Recognition}, pp.
  \bibinfo{pages}{3415--3424}.
%Type = Inproceedings
\bibitem[{Xiong et~al.(2015)Xiong, Zhu, Lin and Tang}]{xiong2015recognize}
\bibinfo{author}{Xiong, Y.}, \bibinfo{author}{Zhu, K.}, \bibinfo{author}{Lin,
  D.}, \bibinfo{author}{Tang, X.}, \bibinfo{year}{2015}.
\newblock \bibinfo{title}{Recognize complex events from static images by fusing
  deep channels}, in: \bibinfo{booktitle}{Proceedings of the IEEE Conference on
  Computer Vision and Pattern Recognition}, pp. \bibinfo{pages}{1600--1609}.
%Type = Article
\bibitem[{Xu et~al.(2019)Xu, Li, Wong, Zhao and Kankanhalli}]{xu2019interact}
\bibinfo{author}{Xu, B.}, \bibinfo{author}{Li, J.}, \bibinfo{author}{Wong, Y.},
  \bibinfo{author}{Zhao, Q.}, \bibinfo{author}{Kankanhalli, M.S.},
  \bibinfo{year}{2019}.
\newblock \bibinfo{title}{Interact as you intend: Intention-driven human-object
  interaction detection}.
\newblock \bibinfo{journal}{IEEE Transactions on Multimedia}
  \bibinfo{volume}{22}, \bibinfo{pages}{1423--1432}.
%Type = Inproceedings
\bibitem[{Xu et~al.(2017)Xu, Zhu, Choy and Fei-Fei}]{xu2017scene}
\bibinfo{author}{Xu, D.}, \bibinfo{author}{Zhu, Y.}, \bibinfo{author}{Choy,
  C.B.}, \bibinfo{author}{Fei-Fei, L.}, \bibinfo{year}{2017}.
\newblock \bibinfo{title}{Scene graph generation by iterative message passing},
  in: \bibinfo{booktitle}{Proceedings of the IEEE conference on computer vision
  and pattern recognition}, pp. \bibinfo{pages}{5410--5419}.
%Type = Inproceedings
\bibitem[{Yan et~al.(2019)Yan, Zhang, Ni, Zhang, Xu and Yang}]{yan2019learning}
\bibinfo{author}{Yan, Y.}, \bibinfo{author}{Zhang, Q.}, \bibinfo{author}{Ni,
  B.}, \bibinfo{author}{Zhang, W.}, \bibinfo{author}{Xu, M.},
  \bibinfo{author}{Yang, X.}, \bibinfo{year}{2019}.
\newblock \bibinfo{title}{Learning context graph for person search}, in:
  \bibinfo{booktitle}{Proceedings of the IEEE/CVF Conference on Computer Vision
  and Pattern Recognition}, pp. \bibinfo{pages}{2158--2167}.
%Type = Inproceedings
\bibitem[{Yang et~al.(2018)Yang, Lu, Lee, Batra and Parikh}]{yang2018graph}
\bibinfo{author}{Yang, J.}, \bibinfo{author}{Lu, J.}, \bibinfo{author}{Lee,
  S.}, \bibinfo{author}{Batra, D.}, \bibinfo{author}{Parikh, D.},
  \bibinfo{year}{2018}.
\newblock \bibinfo{title}{Graph r-cnn for scene graph generation}, in:
  \bibinfo{booktitle}{Proceedings of the European conference on computer vision
  (ECCV)}, pp. \bibinfo{pages}{670--685}.
%Type = Inproceedings
\bibitem[{Yang et~al.(2015)Yang, Luo, Loy and Tang}]{yang2015facial}
\bibinfo{author}{Yang, S.}, \bibinfo{author}{Luo, P.}, \bibinfo{author}{Loy,
  C.C.}, \bibinfo{author}{Tang, X.}, \bibinfo{year}{2015}.
\newblock \bibinfo{title}{From facial parts responses to face detection: A deep
  learning approach}, in: \bibinfo{booktitle}{Proceedings of the IEEE
  international conference on computer vision}, pp.
  \bibinfo{pages}{3676--3684}.
%Type = Inproceedings
\bibitem[{Yang et~al.(2016)Yang, Luo, Loy and Tang}]{yang2016wider}
\bibinfo{author}{Yang, S.}, \bibinfo{author}{Luo, P.}, \bibinfo{author}{Loy,
  C.C.}, \bibinfo{author}{Tang, X.}, \bibinfo{year}{2016}.
\newblock \bibinfo{title}{Wider face: A face detection benchmark}, in:
  \bibinfo{booktitle}{Proceedings of the IEEE conference on computer vision and
  pattern recognition}, pp. \bibinfo{pages}{5525--5533}.
%Type = Inproceedings
\bibitem[{Yang et~al.(2019)Yang, Yang, Liu, Xiao, Davis and
  Kautz}]{yang2019step}
\bibinfo{author}{Yang, X.}, \bibinfo{author}{Yang, X.}, \bibinfo{author}{Liu,
  M.Y.}, \bibinfo{author}{Xiao, F.}, \bibinfo{author}{Davis, L.S.},
  \bibinfo{author}{Kautz, J.}, \bibinfo{year}{2019}.
\newblock \bibinfo{title}{Step: Spatio-temporal progressive learning for video
  action detection}, in: \bibinfo{booktitle}{Proceedings of the IEEE/CVF
  Conference on Computer Vision and Pattern Recognition}, pp.
  \bibinfo{pages}{264--272}.
%Type = Inproceedings
\bibitem[{Yao and Fei-Fei(2010)}]{yao2010modeling}
\bibinfo{author}{Yao, B.}, \bibinfo{author}{Fei-Fei, L.}, \bibinfo{year}{2010}.
\newblock \bibinfo{title}{Modeling mutual context of object and human pose in
  human-object interaction activities}, in: \bibinfo{booktitle}{2010 IEEE
  Computer Society Conference on Computer Vision and Pattern Recognition},
  \bibinfo{organization}{IEEE}. pp. \bibinfo{pages}{17--24}.
%Type = Article
\bibitem[{Yuan et~al.(2019)Yuan, Ma, Wang, Liu and Zhu}]{yuan2019semantic}
\bibinfo{author}{Yuan, Y.}, \bibinfo{author}{Ma, L.}, \bibinfo{author}{Wang,
  J.}, \bibinfo{author}{Liu, W.}, \bibinfo{author}{Zhu, W.},
  \bibinfo{year}{2019}.
\newblock \bibinfo{title}{Semantic conditioned dynamic modulation for temporal
  sentence grounding in videos}.
\newblock \bibinfo{journal}{Advances in Neural Information Processing Systems}
  \bibinfo{volume}{32}.
%Type = Inproceedings
\bibitem[{Zellers et~al.(2018)Zellers, Yatskar, Thomson and
  Choi}]{zellers2018neural}
\bibinfo{author}{Zellers, R.}, \bibinfo{author}{Yatskar, M.},
  \bibinfo{author}{Thomson, S.}, \bibinfo{author}{Choi, Y.},
  \bibinfo{year}{2018}.
\newblock \bibinfo{title}{Neural motifs: Scene graph parsing with global
  context}, in: \bibinfo{booktitle}{Proceedings of the IEEE Conference on
  Computer Vision and Pattern Recognition}, pp. \bibinfo{pages}{5831--5840}.
%Type = Inproceedings
\bibitem[{Zhang et~al.(2020)Zhang, Tseng and Kreiman}]{zhang2020putting}
\bibinfo{author}{Zhang, M.}, \bibinfo{author}{Tseng, C.},
  \bibinfo{author}{Kreiman, G.}, \bibinfo{year}{2020}.
\newblock \bibinfo{title}{Putting visual object recognition in context}, in:
  \bibinfo{booktitle}{Proceedings of the IEEE/CVF Conference on Computer Vision
  and Pattern Recognition}, pp. \bibinfo{pages}{12985--12994}.
%Type = Inproceedings
\bibitem[{Zhang et~al.(2017)Zhang, Wu, Costeira and
  Moura}]{zhang2017understanding}
\bibinfo{author}{Zhang, S.}, \bibinfo{author}{Wu, G.},
  \bibinfo{author}{Costeira, J.P.}, \bibinfo{author}{Moura, J.M.},
  \bibinfo{year}{2017}.
\newblock \bibinfo{title}{Understanding traffic density from large-scale web
  camera data}, in: \bibinfo{booktitle}{Proceedings of the IEEE Conference on
  Computer Vision and Pattern Recognition}, pp. \bibinfo{pages}{5898--5907}.
%Type = Article
\bibitem[{Zheng et~al.(2016)Zheng, Zhang, Sun, Chandraker and
  Tian}]{zheng2016person}
\bibinfo{author}{Zheng, L.}, \bibinfo{author}{Zhang, H.}, \bibinfo{author}{Sun,
  S.}, \bibinfo{author}{Chandraker, M.}, \bibinfo{author}{Tian, Q.},
  \bibinfo{year}{2016}.
\newblock \bibinfo{title}{Person re-identification in the wild}.
\newblock \bibinfo{journal}{arXiv preprint arXiv:1604.02531} .
%Type = Article
\bibitem[{Zhu et~al.(2016)Zhu, Gao and Uchida}]{zhu2016could}
\bibinfo{author}{Zhu, A.}, \bibinfo{author}{Gao, R.}, \bibinfo{author}{Uchida,
  S.}, \bibinfo{year}{2016}.
\newblock \bibinfo{title}{Could scene context be beneficial for scene text
  detection?}
\newblock \bibinfo{journal}{Pattern Recognition} \bibinfo{volume}{58},
  \bibinfo{pages}{204--215}.
%Type = Inproceedings
\bibitem[{Zhu et~al.(2021)Zhu, Chen, Ahmed, Shen and
  Savvides}]{zhu2021semantic}
\bibinfo{author}{Zhu, C.}, \bibinfo{author}{Chen, F.}, \bibinfo{author}{Ahmed,
  U.}, \bibinfo{author}{Shen, Z.}, \bibinfo{author}{Savvides, M.},
  \bibinfo{year}{2021}.
\newblock \bibinfo{title}{Semantic relation reasoning for shot-stable few-shot
  object detection}, in: \bibinfo{booktitle}{Proceedings of the IEEE/CVF
  Conference on Computer Vision and Pattern Recognition}, pp.
  \bibinfo{pages}{8782--8791}.
%Type = Inproceedings
\bibitem[{Zhu et~al.(2013)Zhu, Nayak and Roy-Chowdhury}]{zhu2013context}
\bibinfo{author}{Zhu, Y.}, \bibinfo{author}{Nayak, N.M.},
  \bibinfo{author}{Roy-Chowdhury, A.K.}, \bibinfo{year}{2013}.
\newblock \bibinfo{title}{Context-aware modeling and recognition of activities
  in video}, in: \bibinfo{booktitle}{Proceedings of the IEEE Conference on
  Computer Vision and Pattern Recognition}, pp. \bibinfo{pages}{2491--2498}.

\end{thebibliography}

% \section*{Supplementary Material}

% Supplementary material that may be helpful in the review process should
% be prepared and provided as a separate electronic file. That file can
% then be transformed into PDF format and submitted along with the
% manuscript and graphic files to the appropriate editorial office.

\end{document}